\documentclass[twoside]{article}

\newif\ifanonymize
\anonymizefalse

\newif\ifarxiv
\arxivtrue

\ifanonymize
	\usepackage{aistats2019}
\else
	\usepackage[accepted]{aistats2019}
\fi

\usepackage{hyperref}
\usepackage{url}
\usepackage{booktabs}
\usepackage{amsmath, amsfonts}
\usepackage{nicefrac}
\usepackage{microtype}
\usepackage{bm}
\usepackage{xr}
\usepackage[algoruled, noend]{algorithm2e}
\usepackage{color}
\usepackage{graphicx}
\usepackage{subfig}
\usepackage{enumitem}
\usepackage[numbers, sort]{natbib}
\usepackage{siunitx}

\newcommand{\vect}[1]{\mathbf{#1}}
\newcommand{\mat}[1]{\mathbf{#1}}
\newcommand{\br}[1]{\mathopen{}\left(#1\right)\mathclose{}}
\newcommand{\set}[1]{\left\{#1\right\}}
\newcommand{\pair}[2]{\br{#1,#2}}
\newcommand{\abs}[1]{\left|#1\right|}
\newcommand{\norm}[1]{\left\|#1\right\|}

\newcommand{\prob}[1]{p\br{#1}}
\newcommand{\prop}[1]{\tilde{p}\br{#1}}
\newcommand{\g}{\,|\,}
\newcommand{\kl}[2]{D_{\mathrm{KL}}\br{#1\,\|\,#2}}
\newcommand{\avg}[2]{{\mathop{\mathbb{E}}}_{#2}\br{#1}}

\newcommand{\diff}[1]{\operatorname{d}\!{#1}}
\newcommand{\deriv}[2]{\frac{\partial{#1}}{\partial{#2}}}
\newcommand{\gaussian}[2]{\mathcal{N}\br{#1,#2}}
\newcommand{\gaussianx}[3]{\mathcal{N}\br{#1\g #2,#3}}
\newcommand{\uniform}[2]{\mathcal{U}\br{#1,#2}}

\newcommand{\appref}[1]{\ifarxiv\ref{#1}\else\ref*{app-#1}\fi}
\newcommand{\x}{\vect{x}}
\newcommand{\y}{\vect{u}}
\newcommand{\z}{\vect{v}}
\newcommand{\p}{\bm{\theta}}
\newcommand{\f}{\bm{\phi}}
\newcommand{\q}{q_{\bm{\f}}}

\makeatletter
\newcommand{\removelatexerror}{\let\@latex@error\@gobble}
\makeatother

\newsavebox{\tempbox}
\newlength{\ttempht}
\newcommand{\labelfig}[2]{\sbox{\tempbox}{#2}%
  \setlength{\ttempht}{\ht\tempbox}%
  \addtolength{\ttempht}{-1.5ex}%
  \raisebox{\ttempht}[0pt]{\textbf{\textsf{#1}}}\usebox{\tempbox}}
  
\newcommand{\algref}[1]{\hyperlink{#1_anchor}{\ref*{#1}}}

\newcommand{\ourtitle}{Sequential Neural Likelihood}
\newcommand{\oursubtitle}{Fast Likelihood-free Inference with Autoregressive Flows}

\begin{document}

\twocolumn[

\runningtitle{\ourtitle{}: \oursubtitle}
\aistatstitle{\ourtitle{}:\\\oursubtitle}

\ifanonymize
	\aistatsauthor{Anonymous author(s)}
	\aistatsaddress{Institution(s)}
\else
	\aistatsauthor{George Papamakarios \And David C.~Sterratt \And  Iain Murray}
	\aistatsaddress{University of Edinburgh \And  University of Edinburgh \And University of Edinburgh}
\fi
]

\begin{abstract}

We present Sequential Neural Likelihood (SNL), a new method for Bayesian inference in simulator models, where the likelihood is intractable but simulating data from the model is possible. SNL trains an autoregressive flow on simulated data in order to learn a model of the likelihood in the region of high posterior density. A sequential training procedure guides simulations and reduces simulation cost by orders of magnitude. We show that SNL is more robust, more accurate and requires less tuning than related neural-based methods, and we discuss diagnostics for assessing calibration, convergence and goodness-of-fit.

\end{abstract}

\section{Introduction}
\label{sec:introduction}

In many areas of science and engineering, a natural way to model a stochastic system of interest is as a \emph{simulator}, which may be controlled by potentially interpretable parameters and can be run forward to generate data. Such simulator models lend themselves naturally to modelling mechanistic processes such as particle collisions in high energy physics \citep{Sjostrand:2008, Agostinelli:2003}, universe evolution in cosmology \citep{Schafer:2012, Alsing:2018}, stochastic dynamical systems in ecology and evolutionary biology \citep{Pritchard:1999, Ratmann:2007, Wood:2010, Wilkinson:2011}, macroeconomic models in econometrics \citep{Gourieroux:1993, Bansal:2004}, and biological neural networks in neuroscience \citep{Pospischil:2008, Markram:2015, Sterratt:2011}. Simulators are a powerful and flexible modelling tool; they can take the form of a computer graphics engine \citep{Mansinghka:2013} or a physics engine \citep{Wu:2015}, and they can become part of probabilistic programs \citep{Kulkarni:2015, Cusumano:2017, Casado:2017}.

Due to the wide applicability of simulator models, several tasks of interest can be framed as inference of the parameters of a simulator from observed data. For instance, inferring the parameters of physical theories given particle collisions can be the means for discovering new physics \citep{Brehmer:2018b}, and inferring the scene parameters of a graphics engine given a natural image can be the basis for computer vision \citep{Yuille:2006, Romaszko:2017}. Hence, the development of accurate and efficient general-purpose inference methods for simulator models can have a profound impact in scientific discovery and artificial intelligence.

The fundamental difficulty in inferring the parameters of a simulator given data is the unavailability of the likelihood function. In Bayesian inference, posterior beliefs about parameters $\p$ given data $\x$ can be obtained by multiplying the likelihood $\prob{\x\g\p}$ with prior beliefs $\prob{\p}$ and normalizing. However, calculating the likelihood $\prob{\x\g\p}$ of a simulator model for given parameters $\p$ and data $\x$ is computationally infeasible in general, and hence traditional likelihood-based Bayesian methods, such as variational inference \citep{Wainwright:2008} or Markov Chain Monte Carlo \citep{Neal:1993}, are not directly applicable.

To overcome this difficulty, several methods for \emph{likelihood-free inference} have been developed, such as Approximate Bayesian Computation \citep{Marin:2012} and Synthetic Likelihood \citep{Wood:2010}, that require only the ability to generate data from the simulator. Such methods simulate the model repeatedly, and use the simulated data to build estimates of the parameter posterior. In general, the accuracy of likelihood-free inference improves as the number of simulations increases, but so does the computational cost, especially if the simulator is expensive to run. The challenge in developing new likelihood-free inference methods is to achieve a good trade-off between estimation accuracy and simulation cost.

In this paper we present \emph{Sequential Neural Likelihood (SNL)}, a new likelihood-free method for Bayesian inference of simulator models, based on state-of-the-art conditional neural density estimation with autoregressive flows. The main idea of SNL is to train a Masked Autoregressive Flow \citep{Papamakarios:2017} on simulated data in order to estimate the conditional probability density of data given parameters, which then serves as an accurate model of the likelihood function. During training, a Markov Chain Monte Carlo sampler selects the next batch of simulations to run using the most up-to-date estimate of the likelihood function, leading to a reduction in the number of simulations of several orders of magnitude. To aid the practitioner in deploying SNL, we discuss practical issues such as implementation and tuning, diagnosing convergence, and assessing goodness-of-fit. Experimental results show that SNL is robust and well-calibrated, and that it outperforms existing state-of-the-art methods based on neural density estimation both in terms of posterior estimation accuracy and simulation cost.

\section{Likelihood-free inference with neural density estimation}
\label{sec:background}

A simulator model is a computer program, which takes a vector of parameters $\p$, makes internal calls to a random number generator, and outputs a data vector~$\x$. Implicitly, this procedure defines a conditional probability distribution $\prob{\x\g\p}$ which in general we cannot evaluate, but we can easily sample from by running the program. Given an observed data vector $\x_o$ and a prior distribution $\prob{\p}$, we are interested in estimating the parameter posterior $\prob{\p\g\x_o} \propto \prob{\x_o\g\p}\,\prob{\p}$.

This paper focuses on an approach to likelihood-free inference that is based on neural density estimation. A \emph{conditional neural density estimator} is a parametric model $\q$ (such as a neural network) controlled by a set of parameters $\f$, which takes a pair of datapoints $\pair{\y}{\z}$ and 
outputs a conditional probability density $\q\br{\y\g\z}$. Given a set of training data $\set{\y_n, \z_n}_{1:N}$ that are independent and identically distributed according to a joint probability density $\prob{\y,\z}$, we can train $\q$ by maximizing the total log probability $\sum_n{\log{\q{\br{\y_n\g\z_n}}}}$ with respect to $\f$.
With enough training data, and with a sufficiently flexible model, $\q\br{\y\g\z}$ will learn to approximate the conditional $\prob{\y\g\z}$.

We can use a neural density estimator $\q\br{\p\g\x}$ that models the conditional of parameters given data to approximate the posterior $\prob{\p\g\x_o}$ as follows. First, we obtain a set of samples $\set{\p_n, \x_n}_{1:N}$ from the joint distribution $\prob{\p,\x}$, by $\p_n\sim\prob{\p}$ and $\x_n\sim\prob{\x\g\p_n}$ for $n=1,\ldots,N$.
Then, we train $\q$ using $\set{\p_n, \x_n}_{1:N}$ as training data in order to obtain a global approximation of $\prob{\p\g\x}$. Finally, $\prob{\p\g\x_o}$ can be simply estimated by $\q\br{\p\g\x_o}$.
In practice, a large number of simulations may be required for there to be enough training data in the vicinity of $\x_o$ in order to obtain an accurate posterior fit. However, running the simulator many times can be prohibitively expensive.

\emph{Sequential Neural Posterior Estimation} is a strategy for reducing the number of simulations needed by conditional neural density estimation, originally proposed by \citet{Papamakarios:2016} and further developed by \citet{Lueckmann:2017}. The name SNPE was first used for the method of \citet{Lueckmann:2017}, but in this paper we will use it to refer to both methods, due to their close relationship. The main idea of SNPE is to generate parameter samples $\p_n$ from a proposal $\prop{\p}$ instead of the prior $\prob{\p}$ that makes generated data $\x_n$ more likely to be close to the observed datapoint~$\x_o$. SNPE finds a good proposal $\prop{\p}$ by training the estimator $\q$ over a number of rounds, whereby in each round $\prop{\p}$ is taken to be the approximate posterior obtained in the round before. This sequential procedure converges rapidly and can be implemented with a relatively small number of simulations per round, which leads to massive savings in simulation cost. For its neural density estimator, SNPE uses a \emph{Mixture Density Network} \citep{Bishop:1994}, which is a feedforward neural network that takes $\x$ as input and outputs the parameters of a Gaussian mixture over $\p$. To avoid overfitting due to the small number of training data per round, the MDN is trained with variational dropout \citep{Kingma:2015:dropout}.

The main issue with SNPE is that the proposal biases the approximation of the posterior. Since the parameter samples follow $\prop{\p}$ instead of $\prob{\p}$, the MDN will approximate $\prob{\x_o\g\p}\,\prop{\p}$ instead of $\prob{\x_o\g\p}\,\prob{\p}$ (up to normalization). Hence, an adjustment of either the learned posterior or the proposed samples must be made to account for sampling from the `wrong' prior.
In the variant of \citet{Papamakarios:2016}, which we refer to as SNPE-A, the learned posterior $\q\br{\p\g\x_o}$ is adjusted, by dividing it by $\prop{\p}$ and multiplying it by $\prob{\p}$. SNPE-A restricts $\prop{\p}$ to be Gaussian; since $\q\br{\p\g\x_o}$ is a Gaussian mixture, the division by $\prop{\p}$ can be done analytically. The problem with \hbox{SNPE-A} is that if $\prop{\p}$ happens to have a smaller variance than any of the components of $\q\br{\p\g\x_o}$, the division yields a Gaussian with negative variance, from which the algorithm is unable to recover and thus is forced to terminate prematurely.
In the variant of \citet{Lueckmann:2017}, which we refer to as SNPE-B, the parameter samples $\p_n$ are adjusted, by assigning them weights $w_n=\prob{\p_n}/\prop{\p_n}$. During training, the weighted log likelihood $\,\sum_n{w_n\log{\q{\br{\p_n\g\x_n}}}}$ is used instead of the total log likelihood. Compared to SNPE-A, this method does not require the proposal $\prop{\p}$ to be Gaussian, and it does not suffer from negative variances. However, the weights can have high variance, which may result in high-variance gradients and instability during training.

\section{Sequential Neural Likelihood}
\label{sec:snl}

Our new method, \emph{Sequential Neural Likelihood (SNL)}, avoids the bias introduced by the proposal, by opting to learn a model of the likelihood instead of the posterior. Let $\prop{\p}$ be a proposal distribution over parameters (not necessarily the prior) and let $\prob{\x\g\p}$ be the intractable likelihood of a simulator model. Consider a set of samples $\set{\p_n,\x_n}_{1:N}$ obtained by $\p_n\!\sim\!\prop{\p}$ and $\x_n\!\sim\!\prob{\x\g\p_n}$, and define $\prop{\p,\x} = \prob{\x\g\p}\,\prop{\p}$ to be the joint distribution of each pair $\pair{\p_n}{\x_n}$. Suppose we train a conditional neural density estimator $\q\br{\x\g\p}$, which models the conditional of data given parameters, on the set $\set{\p_n,\x_n}_{1:N}$. For large $N$, maximizing the total log likelihood $\sum_n{\log\q\br{\x_n\g\p_n}}$ is approximately equivalent to maximizing:
\begin{align}
&\avg{\log\q\br{\x\g\p}}{\prop{\p,\x}} =\notag\\
&-\avg{\kl{\prob{\x\g\p}}{\q\br{\x\g\p}}}{\prop{\p}} + \text{const},
\end{align}
where $\kl{\cdot}{\cdot}$ is the Kullback--Leibler divergence. The above quantity attains its maximum when the KL is zero in the support of $\prop{\p}$, i.e.~when $\q\br{\x\g\p} = \prob{\x\g\p}$ for all $\p$ such that $\prop{\p}>0$.
Therefore, given enough simulations, a sufficiently flexible conditional neural density estimator will eventually approximate the likelihood in the support of the proposal, regardless of the shape of the proposal. In other words, as long as we do not exclude parts of the parameter space, the way we propose parameters does not bias learning the likelihood asymptotically. Unlike when learning the posterior, no adjustment is necessary to account for our proposing strategy.

In practice, for a moderate number of simulations, the proposal $\prop{\p}$ controls where $\q\br{\x\g\p}$ will be most accurate. In a parameter region where $\prop{\p}$ is high, there will be a high concentration of training data, hence $\prob{\x\g\p}$ will be approximated better. Since we are ultimately interested in estimating the posterior $\prob{\p\g\x_o}$ for a specific datapoint $\x_o$, it makes sense to use a proposal that is high in regions of high posterior density, and low otherwise, to avoid expending simulations in regions that are not relevant to the inference task.
Inspired by SNPE, we train $\q$ over multiple rounds, indexed by $r\ge 1$. Let $\hat{p}_{r-1}\br{\p\g\x_o}$ be the approximate posterior obtained in round $r\!-\!1$, and take $\hat{p}_0\br{\p\g\x_o} = \prob{\p}$. In round $r$, we generate a new batch of $N$ parameters $\p_n$ from $\hat{p}_{r-1}\br{\p\g\x_o}$, and data $\x_n$ from $\prob{\x\g\p_n}$. We then (re-)train $\q$ on all data generated in rounds $1$ up to $r$, and set $\hat{p}_r\br{\p\g\x_o} \propto \q\br{\x_o\g\p}\,\prob{\p}$. This method, which we call \emph{Sequential Neural Likelihood}, is detailed in Algorithm \algref{alg:snl}.

\begin{figure}[t]
\hypertarget{alg:snl_anchor}{}
\removelatexerror
\begin{algorithm}[H]
\SetKwInOut{Input}{Input}
\SetKwInOut{Output}{Output}
\Input{~observed data $\x_o$, estimator $\q\br{\x\g\p}$, number of rounds $R$, simulations per round $N$}
\Output{~approximate posterior $\hat{p}\br{\p\g\x_o}$}
\BlankLine
set $\hat{p}_0\br{\p\g\x_o} = \prob{\p}$ and $\mathcal{D} = \set{}$ \\
\For{$r=1:R$}{
\For{$n=1:N$}{
sample $\p_n \sim \hat{p}_{r-1}\br{\p\g\x_o}$ with MCMC \\
simulate $\x_n \sim \prob{\x\g \p_n}$ \\
add $\pair{\p_n}{\x_n}$ into $\mathcal{D}$ 
}
{(re-)train} $\q\br{\x\g\p}$ on $\mathcal{D}$ and set $\hat{p}_r\br{\p\g\x_o} \propto \q\br{\x_o\g\p}\,\prob{\p}$
}
\Return $\hat{p}_R\br{\p\g\x_o}$
\caption{Sequential Neural Likelihood (SNL)}
\label{alg:snl}
\end{algorithm}
\end{figure}

In round $r$, SNL effectively uses $rN$ parameter samples from $\tilde{p}_r\br{\p} = \frac{1}{r}\sum_{i=0}^{r-1}{\hat{p}_i\br{\p\g\x_o}}$.
As the amount of training data grows in each round, $\q\br{\x\g\p}$ becomes a more accurate model of the likelihood in the region of high proposal density, and hence the approximate posterior $\hat{p}_r\br{\p\g\x_o}$ gets closer to the exact posterior. In turn, the proposal $\tilde{p}_r\br{\p}$ also tends to the exact posterior, and therefore most of the simulations in later rounds come from parameter regions of high posterior density. In Section~\ref{sec:experiments} we see that focusing on high posterior parameters massively reduces the number of simulations. Finally, unlike SNPE which trains only on simulations from the latest round, SNL trains on all simulations obtained up to each round.

In general, we are free to choose any neural density estimator $\q\br{\x\g\p}$ that is suitable for the task at hand. Because we are interested in a general-purpose solution, we propose taking $\q\br{\x\g\p}$ to be a conditional \emph{Masked Autoregressive Flow} \citep{Papamakarios:2017}, which has been shown to perform well in a variety of general-purpose density estimation tasks. MAF represents $\q\br{\x\g\p}$ as a transformation of a standard Gaussian density $\gaussian{\vect{0}}{\mat{I}}$ through a series of $K$ autoregressive functions $f_1, \ldots, f_K$ each of which depends on $\p$, that is:
\begin{equation}
\x = \vect{z_K}
\quad\text{where}\quad
\begin{array}{l}
\vect{z}_0 \sim \gaussian{\vect{0}}{\mat{I}}\\
\vect{z}_k = f_k\br{\vect{z}_{k-1}, \p}.\\
\end{array}
\label{eq:maf_gen}
\end{equation}
Each $f_k$ is a bijection with a lower-triangular Jacobian matrix, and is implemented by a \emph{Masked Autoencoder for Distribution Estimation} \citep{Germain:2015} conditioned on $\p$. By change of variables, the conditional density is given by
$\q\br{\x\g\p} =
\gaussianx{\vect{z}_0}{\vect{0}}{\mat{I}}
\prod_k{\abs{\det\br{\deriv{f_k}{\vect{z}_{k-1}}}}^{-1}}$.

In order to sample from the approximate posterior $\hat{p}\br{\p\g\x_o}\propto \q\br{\x_o\g\p}\,\prob{\p}$, we use Markov Chain Monte Carlo (MCMC) in the form of Slice Sampling with axis-aligned updates \citep{Neal:2003}. The Markov chain is initialized with a sample from the prior and it persists across rounds; that is, in every round other than the first, the initial state of the chain is the same as its last state in the round before. At the beginning of each round, the chain is burned-in for $200$ iterations, in order to adapt to the new approximate posterior. This scheme requires no tuning, and we found that it performed robustly across our experiments, so we recommend it as a default for general use, at least for tens of parameters that do not have pathologically strong correlations. For parameter spaces of larger dimensionality, other MCMC schemes could be considered, such as Hamiltonian Monte Carlo~\citep{Neal:2011}.

\section{Related work}
\label{sec:related}

\textbf{Approximate Bayesian Computation} \citep{Beaumont:2002, Beaumont:2010, Marin:2012, Blum:2010a, Lintusaari:2017}. ABC is a family of mainly non-parametric methods that repeatedly simulate the model, and reject simulations that do not reproduce the observed data. In practice, to reduce rejection rates to manageable levels, (a) lower-dimensional features (or summary statistics) are used instead of raw data, and (b) simulations are accepted whenever simulated features are within a distance $\epsilon$ from the observed features.
A basic implementation of ABC would simulate parameters from the prior; more sophisticated variants such as \emph{Markov Chain Monte Carlo ABC} \citep{Marjoram:2003, Sisson:2011, Meeds:2015:hamiltonian_abc} and \emph{Sequential Monte Carlo ABC} \citep{Sisson:2007, Beaumont:2009, Toni:2009, Bonassi:2015} guide future simulations based on previously accepted parameters. An issue with ABC in general is that in practice the required number of simulations increases dramatically as $\epsilon$ becomes small, which, as we shall see in Section~\ref{sec:experiments}, can lead to an unfavourable trade-off between estimation accuracy and simulation cost. Advanced ABC algorithms that work for $\epsilon=0$ exist \citep{Graham:2017}, but require the simulator to be differentiable.

\textbf{Learning the posterior}.
Another approach to likelihood-free inference is learning a parametric model of the posterior from simulations. \emph{Regression Adjustment} \citep{Beaumont:2002, Blum:2010b} employs a parametric regressor (such as a linear model or a neural network) to learn the dependence of parameters $\p$ given data $\x$ in order to correct posterior parameter samples (obtained by e.g.~ABC)\@. \emph{Gaussian Copula ABC} \citep{Li:2017} estimates the posterior with a parametric Gaussian copula model. \emph{Variational Likelihood-Free Inference} \citep{Moreno:2016, MinhTran:2017, DustinTran:2017} trains a parametric posterior model by maximizing the (here intractable) variational lower bound; either the bound or its gradients are stochastically estimated from simulations. Parametrically learning the posterior is the target of \emph{Sequential Neural Posterior Estimation} \citep{Lueckmann:2017, Papamakarios:2016}, which SNL directly builds on, and which was discussed in detail in Section~\ref{sec:background}. Beyond SNPE, conditional neural density estimators have been used to learn the posterior from simulations in graphical models \citep{Morris:2001, Paige:2016}, and universal probabilistic programs \citep{Le:2017}.

\textbf{Learning the likelihood}.
Similarly to SNL, a body of work has focused on parametrically approximating the intractable likelihood function. \emph{Synthetic Likelihood} \citep{Wood:2010, Price:2018, Everitt:2018, Ong:2018} estimates the mean $\vect{m}_{\p}$ and covariance matrix $\mat{S}_{\p}$ of a batch of data $\x_n\!\sim\!\prob{\x\g\p}$ sampled at a given $\p$, and then approximates $\prob{\x_o\g\p} \approx \gaussianx{\x_o}{\vect{m}_{\p}}{\mat{S}_{\p}}$. Non-Gaussian likelihood approximations are also possible, e.g.~saddlepoint approximations \citep{Fasiolo:2018}. Typically, SL would be run as an inner loop in the context of an MCMC sampler. \emph{Gaussian Process Surrogate ABC} \citep{Meeds:2014} employs a Gaussian process to model the dependence of $\pair{\vect{m}_{\p}}{\mat{S}_{\p}}$ on $\p$, and uses the uncertainty in the GP to decide whether to run more simulations to estimate $\pair{\vect{m}_{\p}}{\mat{S}_{\p}}$. Other predecessors to SNL include approximating the likelihood as a linear-Gaussian model \citep{Leuenberger:2010}, or as a mixture of Gaussian copulas with marginals modelled by mixtures of experts \citep{Fan:2013}.

\textbf{Learning the likelihood ratio}.
Rather than learning the likelihood $\prob{\x\g\p}$, an alternative approach is learning the likelihood ratio $\prob{\x\g\p_1}/\prob{\x\g\p_2}$ \citep{Cranmer:2016, Pham:2014, Gutmann:2018, Brehmer:2018b, Brehmer:2018c, Stoye:2018}, or the likelihood-to-marginal ratio $\prob{\x\g\p}/\prob{\x}$ \citep{Dutta:2016, Izbicki:2014, DustinTran:2017}. In practice, this can be done by training a classifier to discriminate between data simulated at $\p_1$ vs $\p_2$ for the likelihood ratio, or between data simulated at $\p$ vs random parameter samples from the prior for the likelihood-to-marginal ratio. The likelihood ratio can be used directly for Bayesian inference (e.g.~by multiplying it with the prior and then sampling with MCMC), for hypothesis testing \citep{Cranmer:2016, Brehmer:2018b}, or for estimating the variational lower bound in the context of variational inference \citep{DustinTran:2017}. We note that the strategy for guiding simulations proposed in this paper can (at least in principle) be used with likelihood-ratio estimation too; it might be possible that large computational savings can be achieved as in SNL\@.

\textbf{Guiding simulations}.
Various approaches have been proposed for guiding simulations in order to reduce simulation cost, typically by making use of the observed data $\x_o$ in some way. \emph{Optimization Monte Carlo} \citep{Meeds:2015:omc} directly optimizes parameter samples so as to generate data similar to $\x_o$. \emph{Bayesian Optimization Likelihood-free Inference} \citep{Gutmann:2016} uses Bayesian optimization to find parameter samples that minimize the distance $\norm{\x-\x_o}$. An active-learning-style approach is to use the Bayesian uncertainty in the posterior estimate to choose what simulation to run next \citep{Jarvenpaa:2018, Lueckmann:2018}; in this scheme, the next simulation is chosen to maximally reduce the Bayesian uncertainty. Similarly to \emph{Sequential Neural Posterior Estimation} \citep{Lueckmann:2017, Papamakarios:2016}, SNL selects future simulations by proposing parameters from preliminary approximations to the posterior, which in Section~\ref{sec:experiments} is shown to reduce simulation cost significantly.

\section{Experiments}
\label{sec:experiments}

\subsection{Setup}

In all experiments, we used a Masked Autoregressive Flow (MAF) with $5$ autoregressive layers, each of which has two hidden layers of $50$ units each and tanh nonlinearities. Our design and training choices follow closely the reference implementation \citep{Papamakarios:2017}. We used batch normalization \citep{Ioffe:2015} between autoregressive layers, and trained MAF by stochastically maximizing the total log likelihood using Adam \citep{Kingma:2015:adam} with a minibatch size of $100$ and a learning rate of $10^{-4}$. We used $1000$ simulations in each round of SNL, $5\%$ of which were randomly selected to be used as a validation set; we stopped training if validation log likelihood did not improve after $20$ epochs. No other form of regularization was used other than early stopping. These settings were held constant and performed robustly across all experiments, which is evidence for the robustness of SNL to parameter tuning. We recommend these settings as defaults for general use.

We compare SNL to the following algorithms:

\textbf{Neural Likelihood (NL)}\@. By this we refer to training a MAF (of the same architecture as above) on $N$ simulations from the prior. This is essentially SNL without simulation guiding; we use it as a control to assess the benefit of SNL's guiding strategy. We vary $N$ from $10^3$ to $10^6$ (or more if the simulation cost permits it), and plot the performance for each $N$.

\textbf{SNPE-A} \citep{Papamakarios:2016}. We use $1000$ simulations in each of the proposal-estimating rounds, and $2000$ simulations in the posterior-estimating round (the final round). In all experiments, SNPE-A uses a Mixture Density Network \citep{Bishop:1994} with two hidden layers of $50$ units each, $8$ mixture components, and tanh nonlinearities. We chose the MDN to have roughly as many parameters as MAF in SNL\@. The MDN is trained for $1000$ epochs per round, except for the final round which uses $5000$ epochs, and is regularized with variational dropout \citep{Kingma:2015:dropout}.

\textbf{SNPE-B} \citep{Lueckmann:2017}. We use $1000$ simulations in each round. The same MDN as in SNPE-A is used, and it is trained for $1000$ epochs per round using variational dropout.

\textbf{Synthetic Likelihood (SL)} \citep{Wood:2010}. Our implementation uses axis-aligned Slice Sampling \citep{Neal:2003}, where the intractable likelihood is approximated by a Gaussian fitted to a batch of $N$ simulations run on the fly at each visited parameter $\p$. We vary $N$ from $10$ to $100$--$1000$ (depending on what the simulation cost of each experiment permits) and plot the performance of SL for each $N$. We run Slice Sampling until we obtain $1000$ posterior samples.

\textbf{Sequential Monte Carlo ABC (SMC-ABC)}\@. We use the version of \citet{Beaumont:2009}. We use $1000$ particles, and resample the population if the effective sample size falls below $50\%$. We set the initial $\epsilon$ such that the acceptance probability in the first round is at least $20\%$, and decay $\epsilon$ by a factor of $0.9$ per round.

We chose SNPE-A/B because they are the most related methods, and because they achieve state-of-the-art results in the literature; the comparison with them is intended to establish the competitiveness of SNL\@. The versions of SL and SMC-ABC we use here are not state-of-the-art, but they are robust and widely-used, and are intended as baselines. The comparison with them is meant to demonstrate the gap between state-of-the-art neural-based methods and off-the-shelf, commonly-used alternatives.

\subsection{Results}

We demonstrate SNL in four cases: (a) a toy model with complex posterior, (b) an M/G/1 queue model, (c) a Lotka--Volterra model from ecology, and (d) a Hodgkin--Huxley model of neural activity from neuroscience. The first two models are fast to simulate, and are intended as toy demonstrations. The last two models are relatively slow to simulate (as they involve numerically solving differential equations), and are illustrative of real-world problems of interest. In the last two models, the computational cost of training the neural networks for SNPE-A/B and SNL is negligible compared to the cost of simulating training data.

A detailed description of the models and the full set of results are in Appendices~\appref{sec:simulators} and \appref{sec:results}. Code that reproduces the experiments with detailed user instructions can be found \ifanonymize in the supplementary material. \else at \url{https://github.com/gpapamak/snl}. \fi

\begin{figure*}[tb]
\centering

\begin{tabular}{@{}cc@{}}

\begin{minipage}{0.44\textwidth}
\labelfig{a}{\includegraphics[width=\textwidth]{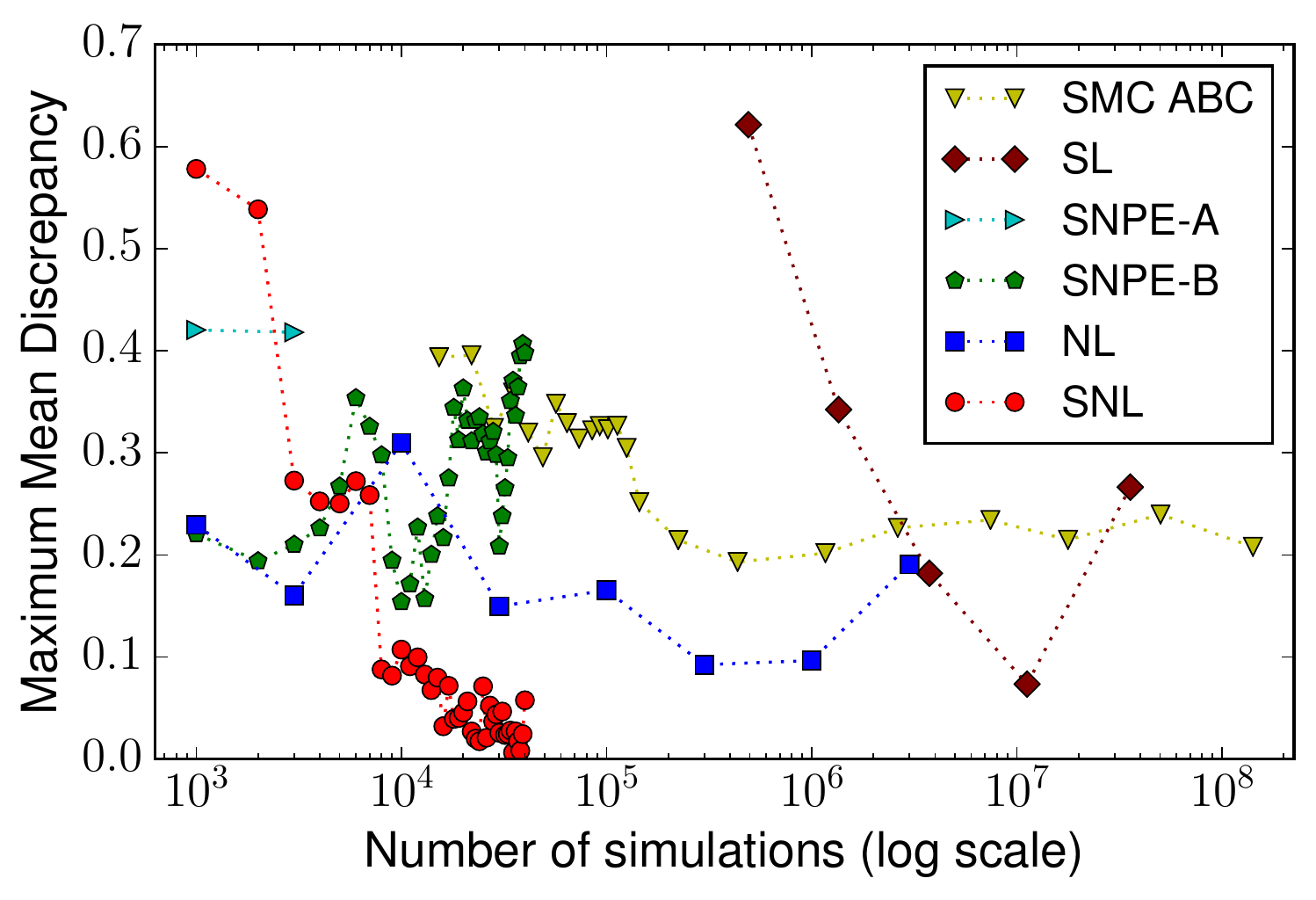}}
\end{minipage}

&

\begin{minipage}{0.44\textwidth}
\begin{tabular}{c@{}c@{}c}
\labelfig{b}{\includegraphics[width=0.309\textwidth]{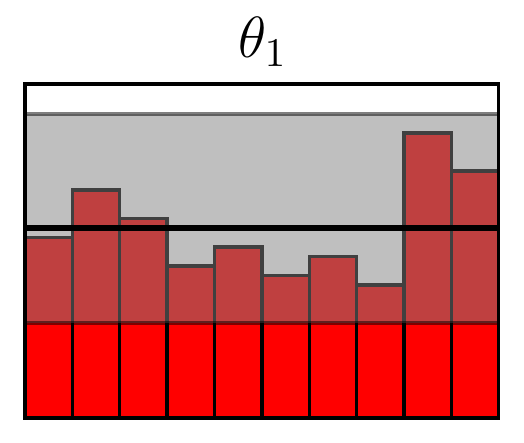}} &
\includegraphics[width=0.309\textwidth]{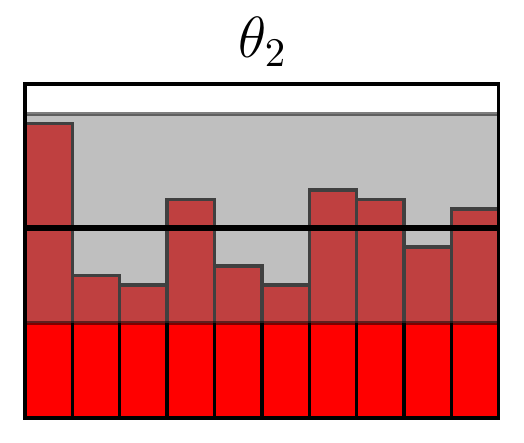} &
\includegraphics[width=0.309\textwidth]{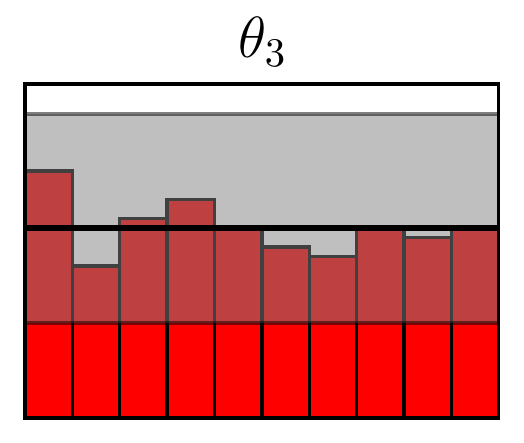} \\
\multicolumn{3}{c}{\labelfig{c}{\includegraphics[width=0.925\textwidth]{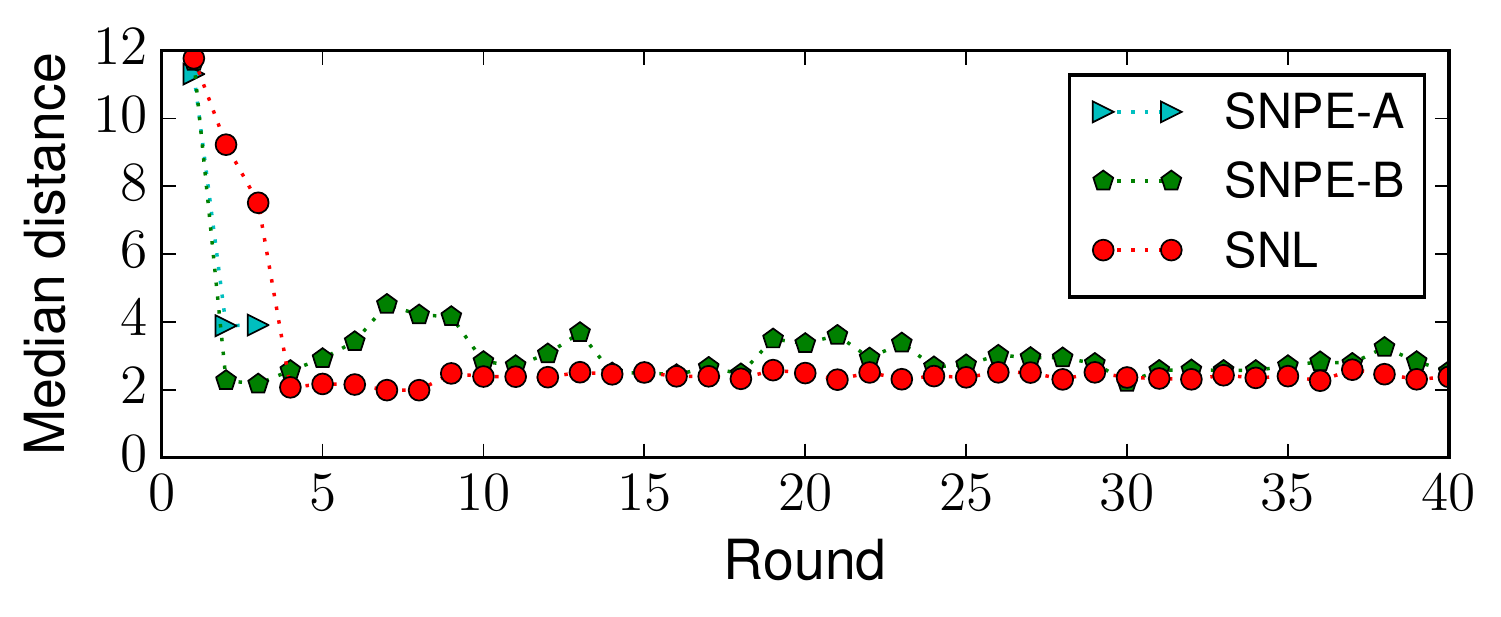}}} 
\end{tabular}
\end{minipage}

\end{tabular}

\caption{A toy model with complex posterior. \textbf{a}:~Accuracy vs simulation cost: bottom left is best. \textbf{b}:~Calibration test for SNL, histogram outside gray band indicates poor calibration. \textbf{c}:~Median distance from simulated to observed data.}
\label{fig:gauss:main}

\end{figure*}

\textbf{A toy model with complex posterior}.
We consider the following model, where $\p$ is $5$-dimensional, and $\x$ is a set of four $2$-dimensional points (or an $8$-dimensional vector) sampled from a Gaussian whose mean $\vect{m}_{\p}$ and covariance matrix $\mat{S}_{\p}$ are functions of $\p$:
\begin{align}
&\theta_i\sim\uniform{-3}{3}\quad\text{for}\quad i=1,\ldots,5 \\
&\vect{m}_{\p} = \pair{\theta_1}{\theta_2} \\
&s_1 = \theta_3^2, \quad s_2 = \theta_4^2, \quad \rho=\tanh\br{\theta_5}\\
&\mat{S}_{\p} =\br{\begin{matrix}
s_1^2 & \rho s_1s_2 \\
\rho s_1s_2 & s_2^2 \\
\end{matrix}} \\
&\vect{x} = \br{\vect{x}_1, \ldots, \vect{x}_4}\quad\text{where}\quad \vect{x}_j \sim \gaussian{\vect{m}_{\p}}{\mat{S}_{\p}}.
\end{align}
The likelihood is $\prob{\x\g\p}=\prod_j{\gaussianx{\vect{x}_j}{\vect{m}_{\p}}{\mat{S}_{\p}}}$. Despite the model's simplicity, the posterior is complex and non-trivial: it has four symmetric modes (due to squaring), and vertical cut-offs (due to the uniform prior). This example illustrates that the posterior can be complicated even if the prior and the likelihood are composed entirely of simple operations (which is a common situation e.g.~in probabilistic programming). In such situations, approximating the likelihood can be simpler than approximating the posterior.

Figure~\hyperref[fig:gauss:main]{\ref*{fig:gauss:main}a} shows the Maximum Mean Discrepancy (MMD) \citep{Gretton:2012} between the approximate posterior of each method and the true posterior vs the total number of simulations used. SNL achieves the best trade-off between accuracy and simulation cost, and achieves the most accurate approximation to the posterior overall. SNPE-A fails in the second round due to the variance of the proposal becoming negative (hence there are only two points in the graph) and SNPE-B experiences high variability; SNL is significantly more robust in comparison. SMC-ABC and SL require orders of magnitude more simulations than the sequential neural methods.

To assess whether SNL is well-calibrated, we performed a simulation-based calibration test \citep{Talts:2018}: we generated $200$ pairs $\pair{\p_n}{\x_n}$ 
from the joint $\prob{\p,\x}$, used SNL to approximate the posterior for each $\x_n$,
obtained $9$ close-to-independent samples from each posterior, and calculated the rank statistic of each parameter $\theta_{ni}$ for $i=1,\ldots,5$ in the corresponding set of posterior samples. If SNL is well-calibrated, the distribution of each rank statistic must be uniform. The histograms of the first three rank statistics are shown in Figure~\hyperref[fig:gauss:main]{\ref*{fig:gauss:main}b}, with a gray band showing the expected variability of a uniform histogram. The test does not find evidence of any gross mis-calibration, and suggests that SNL performs consistently across multiple runs (here $200$).

Another diagnostic is shown in Figure~\hyperref[fig:gauss:main]{\ref*{fig:gauss:main}c}, where we plot the median distance between simulated and observed data for each round. From this plot we can assess convergence, and determine the minimum number of rounds to run for. SNL has lower median distance compared to SNPE-B, which is evidence that SNPE-B has not estimated the posterior accurately enough (as also shown in the left plot).

\begin{figure*}[tb]
\centering

\begin{tabular}{@{}cc@{}}

\begin{minipage}{0.44\textwidth}
\labelfig{a}{\includegraphics[width=\textwidth]{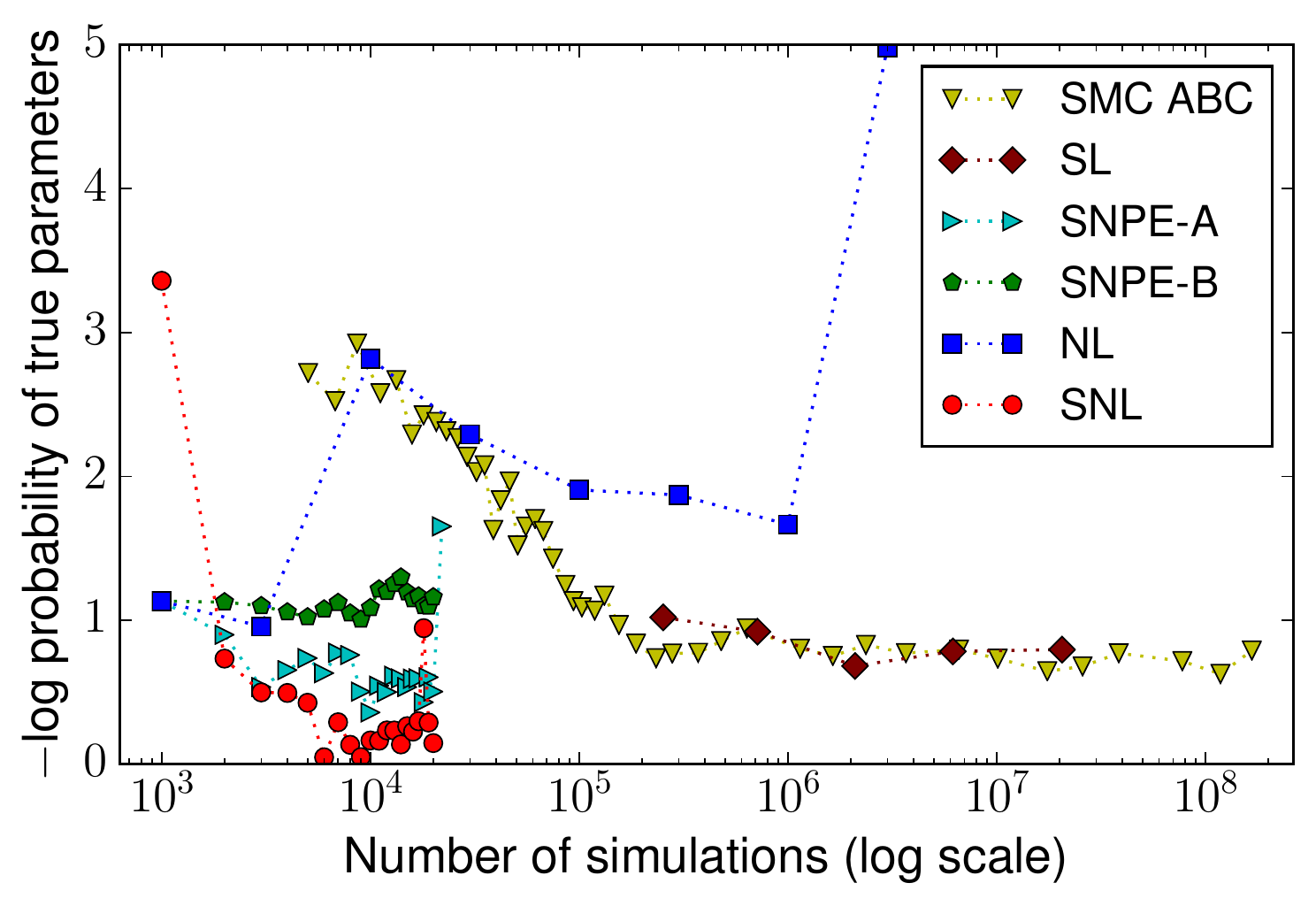}}
\end{minipage}

&

\begin{minipage}{0.44\textwidth}
\begin{tabular}{c@{}c@{}c}
\labelfig{b}{\includegraphics[width=0.309\textwidth]{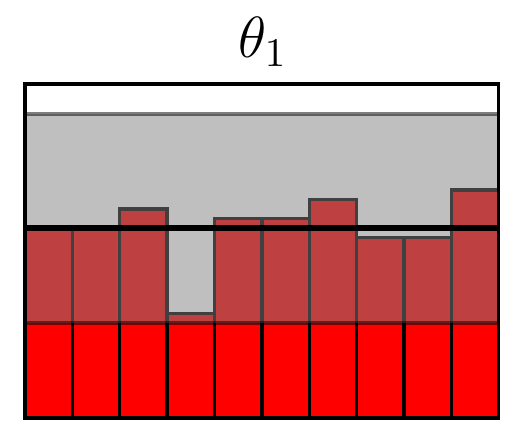}} &
\includegraphics[width=0.309\textwidth]{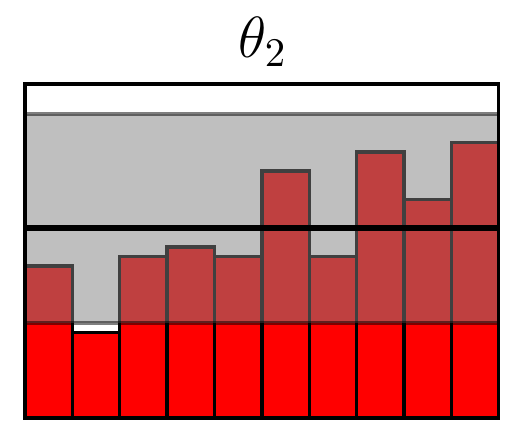} &
\includegraphics[width=0.309\textwidth]{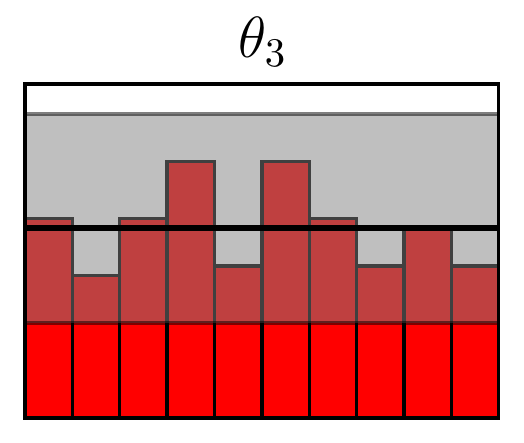} \\
\multicolumn{3}{c}{\labelfig{c}{\includegraphics[width=0.925\textwidth]{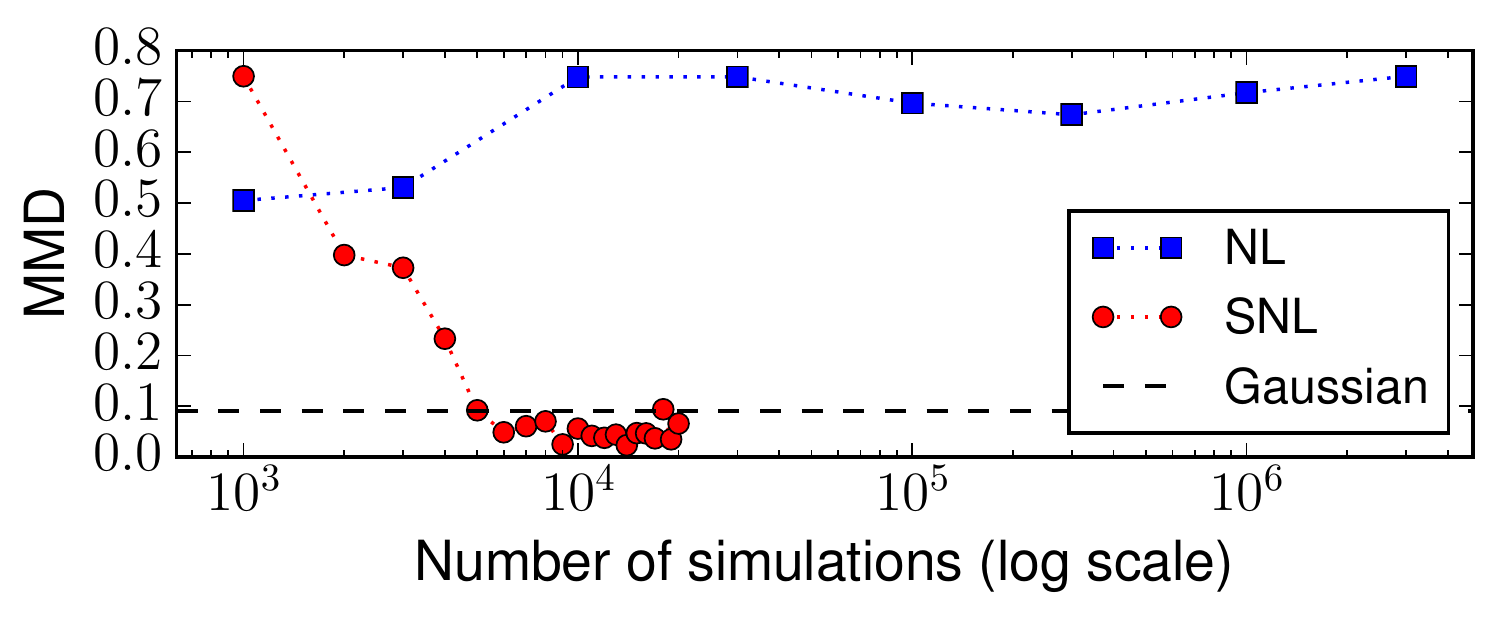}}} 
\end{tabular}
\end{minipage}

\end{tabular}

\caption{M/G/1 queue model. \textbf{a}:~Accuracy vs simulation cost: bottom left is best. \textbf{b}:~Calibration test for SNL, histogram outside gray band indicates poor calibration. \textbf{c}:~Likelihood goodness-of-fit vs simulation cost, calculated at true parameters.}
\label{fig:mg1:main}
\end{figure*}

\textbf{M/G/1 queue model} \citep{Shestopaloff:2014}.
The model describes a server processing customers waiting in a queue, and has parameters $\p=\br{\theta_1, \theta_2, \theta_3}$. The time it takes to serve a customer is uniformly distributed in $\left[\theta_1, \theta_2\right]$, and the time between customer arrivals is exponentially distributed with rate $\theta_3$. The data $\x$ is $5$ equally spaced quantiles of the distribution of inter-departure times. The model can be easily simulated, but the likelihood $\prob{\x\g\p}$ is intractable. Our experimental setup follows \citet{Papamakarios:2016}.

The trade-off between accuracy and simulation cost is shown in Figure~\hyperref[fig:mg1:main]{\ref*{fig:mg1:main}a}. Here we do not have access to the true posterior; instead we plot the negative log probability of the true parameters, obtained by kernel density estimation on posterior samples, vs number of simulations. SNL recovers the true parameters faster and more accurately than all other methods. Although high posterior probability of true parameters is not enough evidence that the posterior is correctly calibrated (e.g.~it could be that the approximate posterior happens to be centred at the right value, but is overconfident), the calibration test (Figure~\hyperref[fig:mg1:main]{\ref*{fig:mg1:main}b}) suggests that SNL is indeed well-calibrated.

Another diagnostic possible with SNL is to check how well $\q\br{\x\g\p}$ fits the data distribution $\prob{\x\g\p}$ for a certain value of $\p$. To do this we (a)~generate $N$ independent samples from $\prob{\x\g\p}$ by running the simulator, (b)~generate $N$ independent samples from $\q\br{\x\g\p}$ using Equation~\eqref{eq:maf_gen}, and (c) calculate the Maximum Mean Discrepancy \citep{Gretton:2012} between the two sets of samples. This is shown in Figure~\hyperref[fig:mg1:main]{\ref*{fig:mg1:main}c} using the true parameters, where we compare with NL and a baseline Gaussian directly fitted to the samples from $\prob{\x\g\p}$. This plot can be used to assess the performance of SNL, and also determine how many rounds to run SNL for. We note that this kind of diagnostic is not possible with methods that approximate the posterior or the likelihood ratio instead of the likelihood.

\textbf{Lotka--Volterra population model} \cite{Wilkinson:2011}.
This is a Markov jump process that models the interaction of a population of predators with a population of prey. It is a classic model of oscillating populations. It has four parameters $\p$, which control the rate of (a) predator births, (b) predator deaths, (c) prey births, and (d) predator-prey interactions. The process can be simulated exactly (by using e.g.~the Gillespie algorithm \citep{Gillespie:1977}), but its likelihood is intractable. In our experiments we follow the setup of \citet{Papamakarios:2016}, where $\x$ is $9$ features of the population timeseries.

\begin{figure*}[tb]
\centering

\begin{tabular}{@{}cc@{}}

\begin{minipage}{0.44\textwidth}
\labelfig{a}{\includegraphics[width=\textwidth]{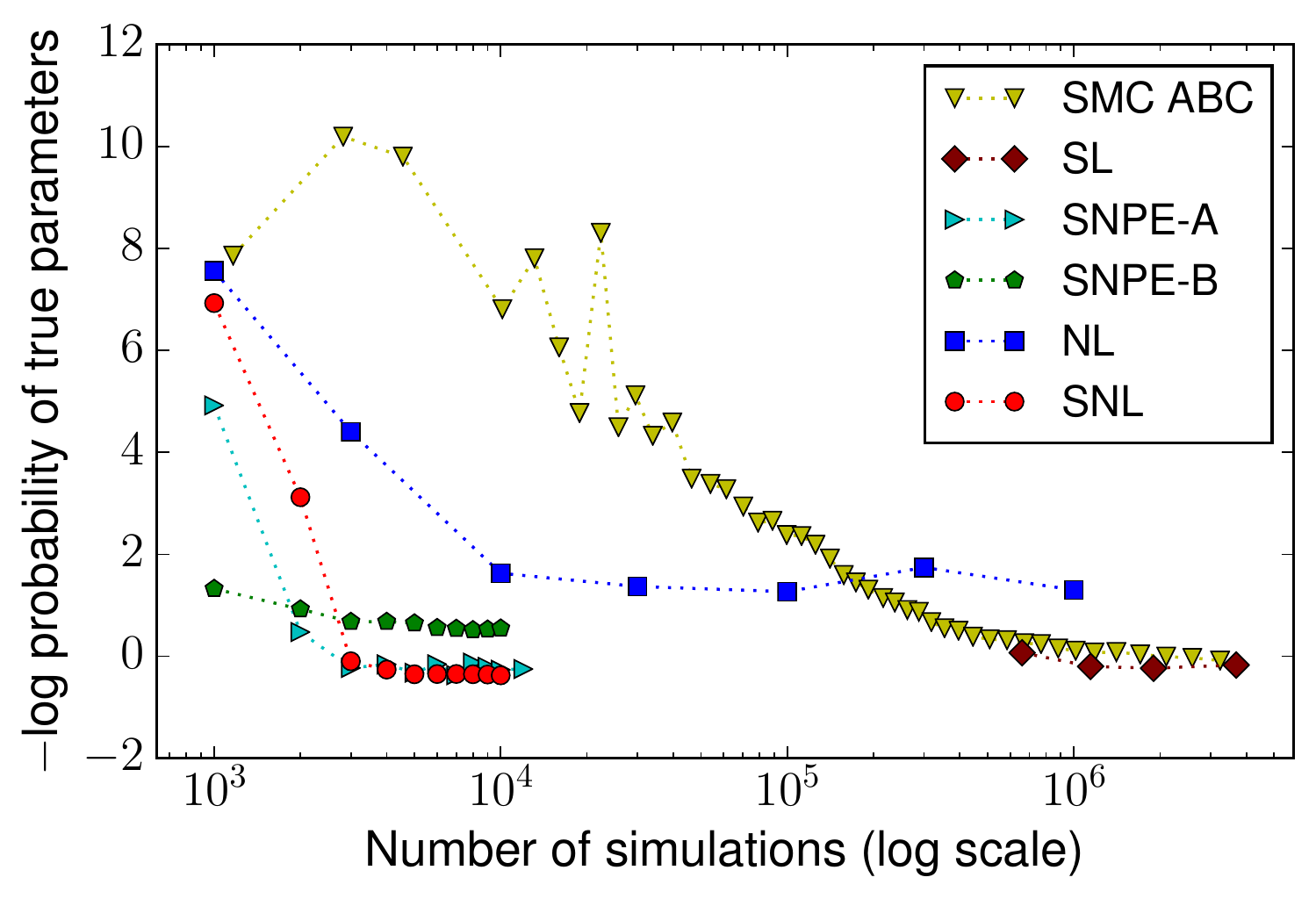}}
\end{minipage}

&

\begin{minipage}{0.44\textwidth}
\begin{tabular}{cc}
\labelfig{b}{\hspace{1.2em}\includegraphics[width=0.4\textwidth]{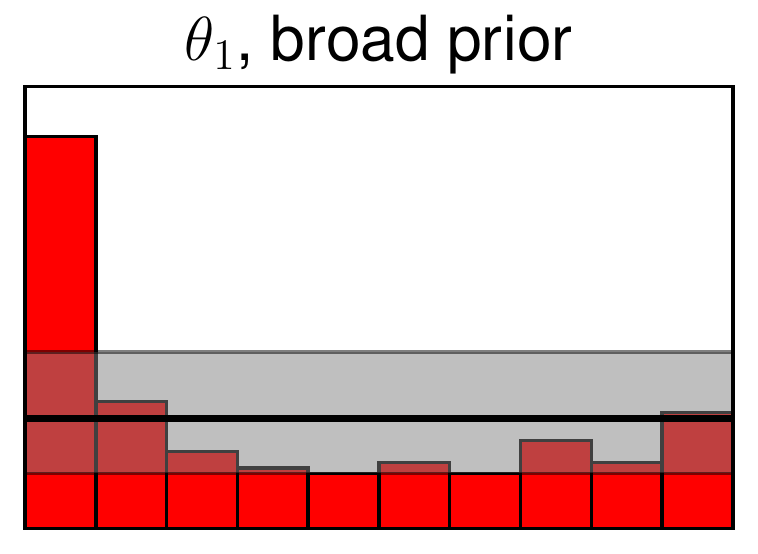}} &
\hspace{-1em}\includegraphics[width=0.4\textwidth]{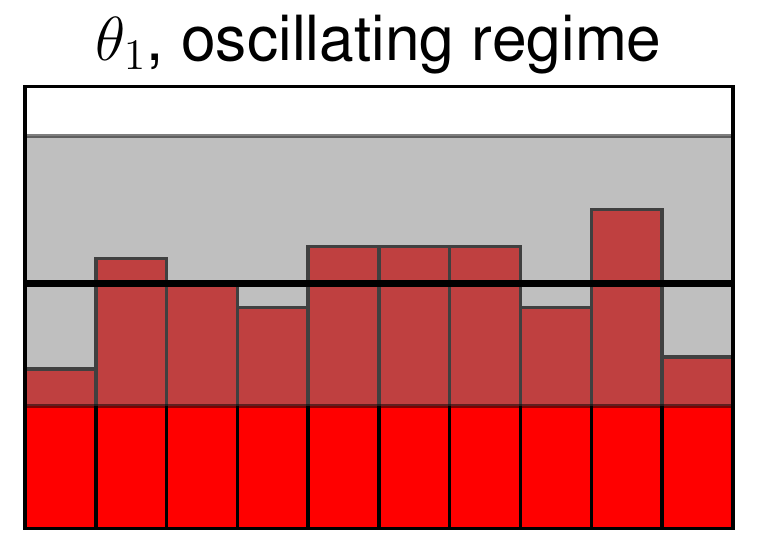} \\
\multicolumn{2}{c}{\labelfig{c}{\includegraphics[width=0.925\textwidth]{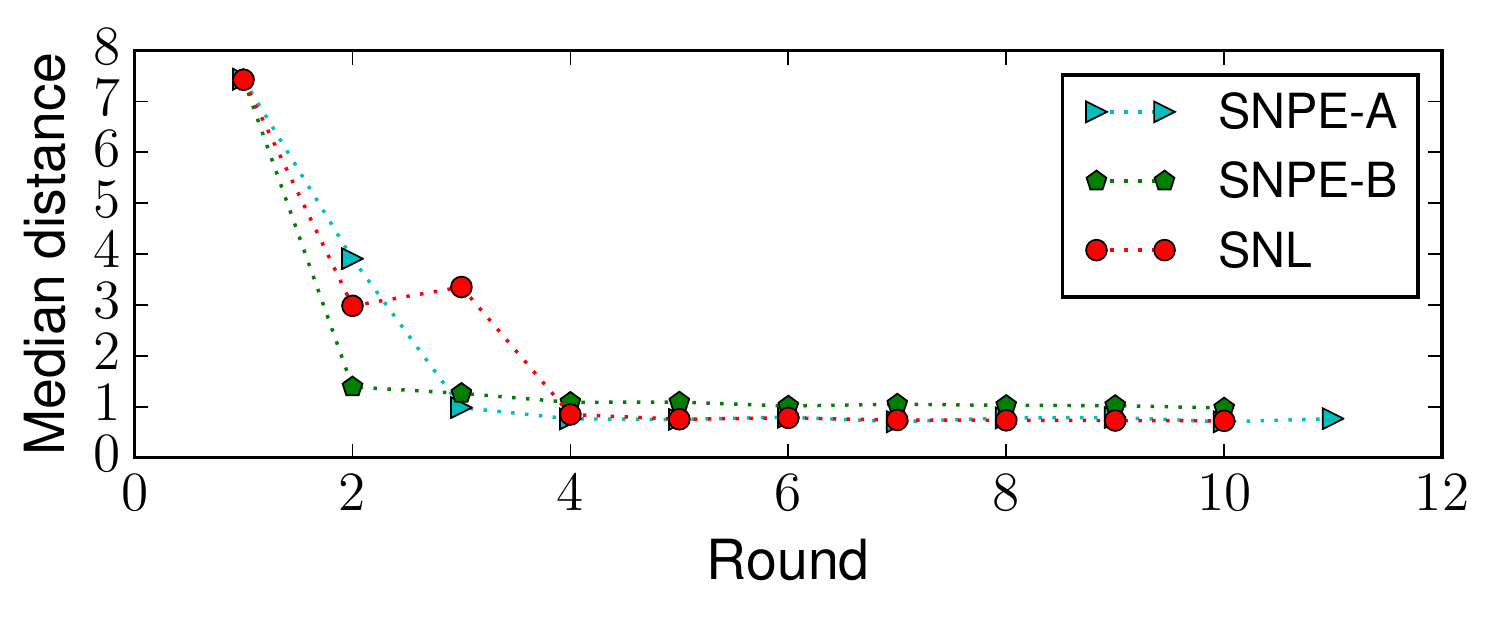}}}
\end{tabular}
\end{minipage}

\end{tabular}

\caption{Lotka--Volterra model. \textbf{a}:~Accuracy vs simulation cost: bottom left is best. \textbf{b}:~Calibration tests for SNL, histogram outside gray band indicates poor calibration.
Calibration depends on data regime (see main text).
\textbf{c}:~Median distance from simulated to observed data.}
\label{fig:lv:main}
\end{figure*}

Figure~\hyperref[fig:lv:main]{\ref*{fig:lv:main}a} shows negative log probability of true parameters vs simulation cost. SNL and SNPE-A perform the best, whereas SNPE-B is less accurate.
Running the calibration test for SNL, using parameters drawn from a broad prior, shows the procedure is sometimes over-confident (Figure~\hyperref[fig:lv:main]{\ref*{fig:lv:main}b}, left plot). In this test, many of the `true' parameters considered corresponded to uninteresting models, where both populations died out quickly, or the prey population diverged. In an application we want to know if the procedure is well-behaved for the sorts of parameters we seem to have. We ran a calibration test for an alternative prior, constrained to parameters that give the oscillating behaviour observed in interesting data. This test suggests that the calibration is reasonable when modelling oscillating data (Figure~\hyperref[fig:lv:main]{\ref*{fig:lv:main}b}, right plot). If we had still observed calibration problems we would have investigated using larger neural networks and/or longer MCMC runs in SNL\@. 
Figure~\hyperref[fig:lv:main]{\ref*{fig:lv:main}c} shows the median distance between the simulated data and the observed data for each round of SNPE-A, SNPE-B and SNL\@. We see that SNPE-A and SNL have a lower median distance compared to SNPE-B, which suggests that SNPE-B has not estimated the posterior accurately enough.

\textbf{Hodgkin--Huxley neuron model} \citep{Hodgkin:1952}.
This model describes how the electrical potential measured across the cell membrane of a neuron varies over time as a function of current injected through an electrode. We used a model of a regular-spiking cortical
pyramidal cell \cite{Pospischil:2008}; the model is described by a set of five coupled ordinary differential equations, which we solved numerically using NEURON \citep{Carnevale:2006}\@. This is a challenging problem of practical interest in neuroscience \citep{Daly:2015, Lueckmann:2017}, whose task is to infer $12$ parameters describing the function of the neuron from features of the membrane potential timeseries. We followed the setup of \citet{Lueckmann:2017}, and we used  $18$ features extracted from the timeseries as data $\x$.

\begin{figure*}[tb]
\centering

\begin{tabular}{@{}cc@{}}

\begin{minipage}{0.44\textwidth}
\labelfig{a}{\includegraphics[width=\textwidth]{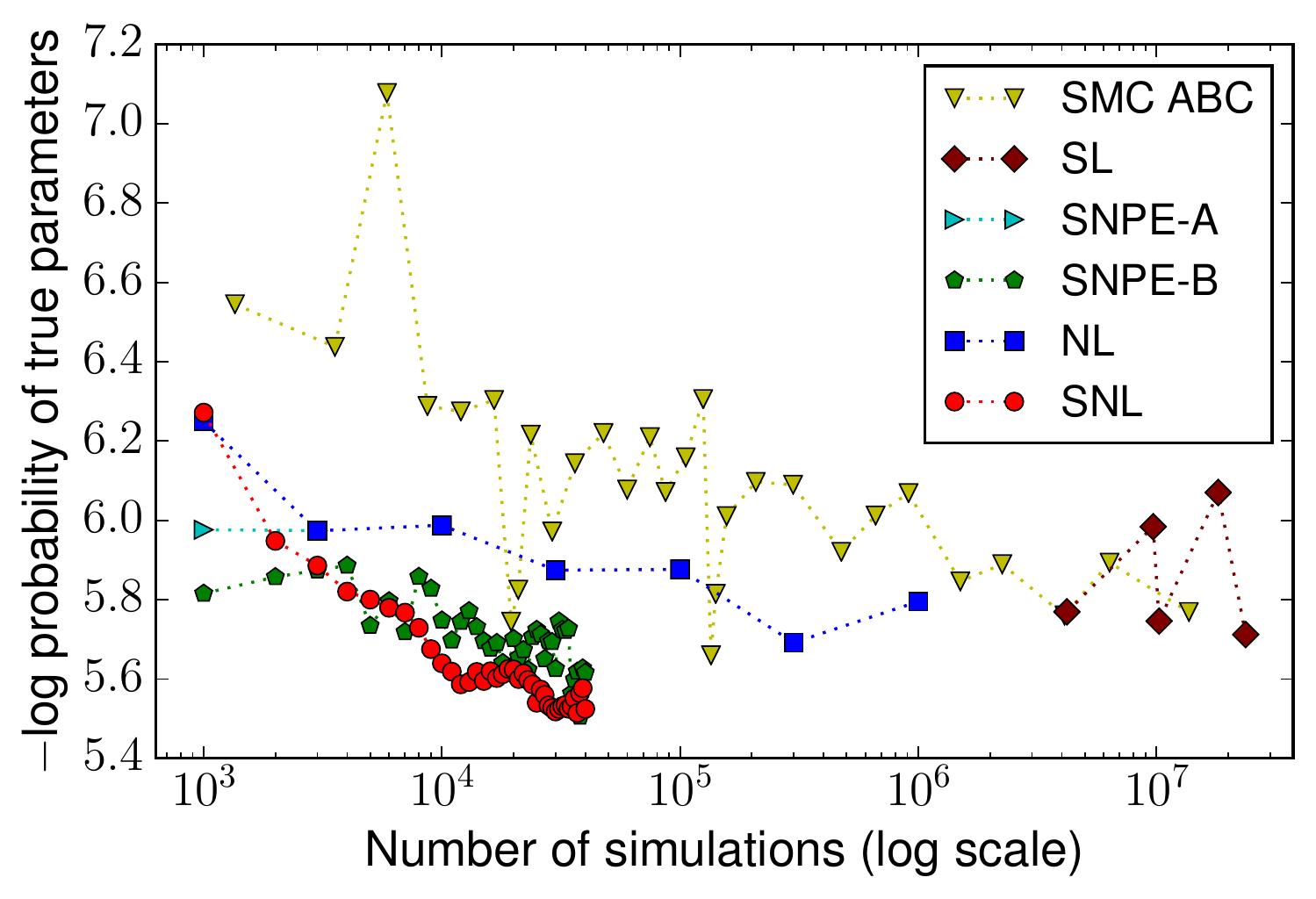}}
\end{minipage}

&

\begin{minipage}{0.44\textwidth}
\begin{tabular}{c@{}c@{}c}
\labelfig{b}{\includegraphics[width=0.309\textwidth]{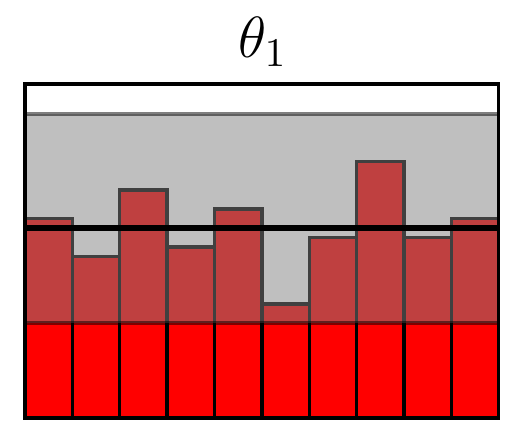}} &
\includegraphics[width=0.309\textwidth]{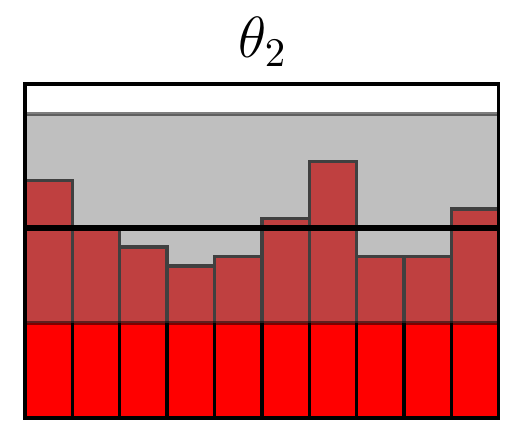} &
\includegraphics[width=0.309\textwidth]{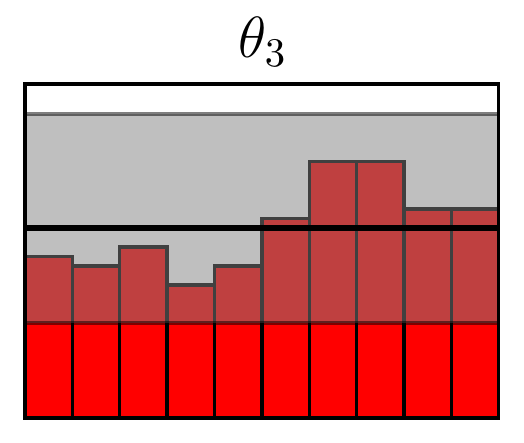} \\
\multicolumn{3}{c}{\labelfig{c}{\includegraphics[width=0.925\textwidth]{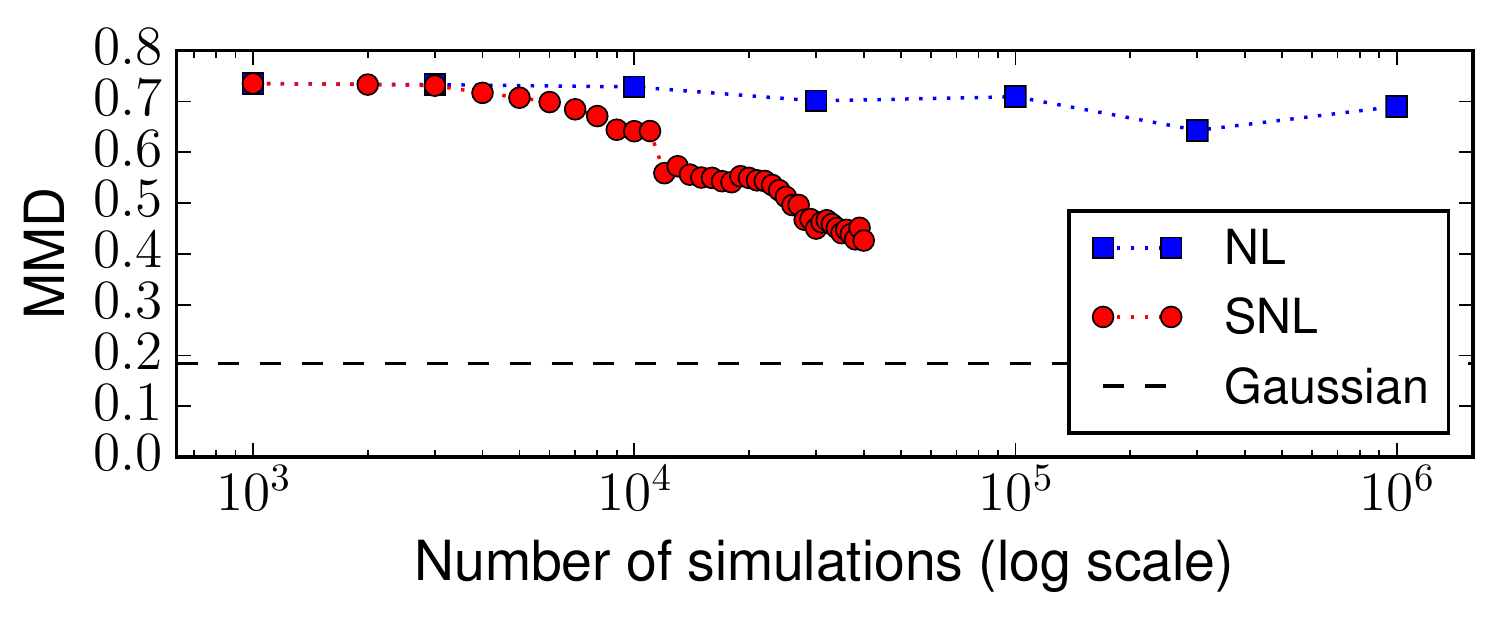}}} 
\end{tabular}
\end{minipage}

\end{tabular}

\caption{Hodgkin--Huxley model. \textbf{a}: Accuracy vs simulation cost: bottom left is best. \textbf{b}: Calibration test for SNL, histogram outside gray band indicates poor calibration. \textbf{c}: Likelihood goodness-of-fit vs simulation cost, calculated at true parameters.}
\label{fig:hh:main}
\end{figure*}

Figure~\hyperref[fig:hh:main]{\ref*{fig:hh:main}a} shows negative log probability of true parameters vs simulation cost; SNL outperforms all other methods. SNPE-A fails in the second round due to the variance of the proposal becoming negative. The calibration test in Figure~\hyperref[fig:hh:main]{\ref*{fig:hh:main}b} suggests that SNL is well-calibrated. Figure~\hyperref[fig:hh:main]{\ref*{fig:hh:main}c} shows goodness-of-fit between MAF and the true likelihood as measured by MMD; SNL converges faster than NL to the true likelihood. The comparison with a baseline Gaussian fit is evidence that SNL has not fully converged, which indicates that running SNL for longer may improve results further.

\section{Discussion}
\label{sec:discussion}

\textbf{Performance and robustness of SNL}.
Our experiments show that SNL is accurate, efficient and robust. The comparison between SNL and NL demonstrates that guiding the simulations is crucial in achieving a good likelihood fit in the region of interest with a reasonable number of simulations. The comparison between SNL and SMC-ABC shows that flexible parametric models can be more accurate and cost-effective compared to non-parametric alternatives. The comparison with SNPE shows that SNL is more robust; SNPE-A failed in two out of four cases, and SNPE-B exhibited high variability. We used the same MAF architecture and training hyperparameters in all our experiments, which shows that a flexible neural architecture can be broadly applicable without extensive problem-specific tuning. We also discussed a number of diagnostics that can be used with SNL to determine the number of rounds to run it for, assess calibration, diagnose convergence and check goodness-of-fit.

\textbf{Scaling to high-dimensional data}.
Likelihood-free inference becomes challenging when the dimensionality of the data $\vect{x}$ is large, which in practice necessitates the use of low-dimensional features (or summary statistics) in place of the raw data. SNL relies on estimating the density of the data, which is a hard problem in high dimensions. A potential strategy for scaling SNL up to high dimensions is exploiting the structure of the data. For instance, if $\x=\br{\x_1, \ldots, \x_N}$ is a dataset of a large number of i.i.d.~datapoints, we could decompose the likelihood as $\prob{\x\g\p}=\prod_n{\prob{\x_n\g\p}}$ and only learn a model ${\q\br{\x_n\g\p}}$ of the density in the lower-dimensional space of individual points. By exploiting data structure, neural density estimators have been shown to accurately model high-dimensional data such as images \citep{VanDenOord:2016:PixelRNN, VanDenOord:2016:PixelCNN, Dinh:2017} and raw audio \citep{VanDenOord:2016:WaveNet, VanDenOord:2018}; SNL could easily incorporate such or further advances in neural density estimation for high-dimensional structured objects.

\textbf{Learning the likelihood vs the posterior}.
A general question is whether it is preferable to learn the posterior or the likelihood; SNPE learns the posterior, whereas SNL targets the likelihood. As we saw, learning the likelihood can often be easier than learning the posterior, and it does not depend on the choice of proposal, which makes learning easier and more robust. Moreover, a model of the likelihood can be reused with different priors, and is in itself an object of interest that can be used for identifiability analysis \citep{Lintusaari:2016} or hypothesis testing \citep{Cranmer:2016, Brehmer:2018b}. On the other hand, methods such as SNPE return a parametric model of the posterior directly, whereas a further inference step (e.g.~variational inference or MCMC) is needed on top of SNL to obtain a posterior estimate, which introduces further computational cost and approximation error. Ultimately, the best approach depends on the problem and application at hand.

\ifanonymize \else
\subsubsection*{Acknowledgments}

We thank Kyle Cranmer and Michael Gutmann for useful discussions, and James Ritchie for pointing us to simulation-based calibration \citep{Talts:2018}.
George Papamakarios was supported by the Centre for Doctoral Training in Data Science, funded by EPSRC (grant EP/L016427/1) and the University of Edinburgh, and by Microsoft Research through its PhD Scholarship Programme. David C.~Sterratt received funding from the European Union’s
Horizon 2020 research and innovation programme under grant agreement No 785907.
\fi

\renewcommand{\theHsection}{\Alph{section}}
\appendix

\section{Detailed description of simulator models used in experiments}
\label{sec:simulators}

\subsection{A toy problem with complex posterior}

This toy model illustrates that even simple models can have non-trivial posteriors. The model has $5$ parameters $\p=\br{\theta_1, \ldots, \theta_5}$ sampled from a uniform prior as follows:
\begin{equation}
\theta_i\sim\uniform{-3}{\,3}\quad\text{for}\quad i=1,\ldots,5.
\end{equation}
Given parameters $\p$, the data $\x$ is generated as follows:
\begin{align}
&\vect{m}_{\p} = \pair{\theta_1}{\theta_2}\displaybreak[0]\\
\label{eq:gauss:sq1}&s_1 = \theta_3^2\displaybreak[0]\\
\label{eq:gauss:sq2}&s_2 = \theta_4^2\displaybreak[0]\\
&\rho=\tanh\br{\theta_5}\displaybreak[0]\\
&\mat{S}_{\p} =\br{\begin{matrix}
s_1^2 & \rho s_1s_2 \\
\rho s_1s_2 & s_2^2 \\
\end{matrix}} \displaybreak[0]\\
&\vect{x}_j \sim \gaussian{\vect{m}_{\p}}{\mat{S}_{\p}}\quad\text{for}\quad j=1,\ldots,4 \displaybreak[0]\\
&\vect{x} = \br{\vect{x}_1, \ldots, \vect{x}_4}.
\end{align}
The data $\x$ is $8$-dimensional. The likelihood is:
\begin{equation}
\prob{\x\g\p} = \prod_{j=1}^4{\gaussianx{\x_j}{\vect{m}_{\p}}{\mat{S}_{\p}}}.
\end{equation}
In our experiments, we took the ground truth parameters to be:
\begin{equation}
\p^* = \br{0.7,\;\; -2.9,\;\; -1,\;\; -0.9,\;\; 0.6},
\end{equation}
and simulated the model with parameters $\p^*$ to get observed data $\x_o$.

\subsection{M/G/1 queue model}

The M/G/1 queue model \citep{Shestopaloff:2014} describes how a server processes a queue of arriving customers. Our experimental setup follows \citet{Papamakarios:2016}. There are $3$ parameters $\p=\br{\theta_1, \theta_2, \theta_3}$ sampled from a uniform prior as follows:
\begin{align}
\theta_1 &\sim \mathcal{U}\br{0,10} \label{eq:mg1:s}\\
\theta_2-\theta_1 &\sim \mathcal{U}\br{0,10} \label{eq:mg1:a}\\
\theta_3 &\sim \mathcal{U}\br{0,\nicefrac{1}{3}}.\label{eq:mg1:d}
\end{align}
Let $I$ be the total number of customers, $s_i$ be the time the server takes to serve customer $i$, $a_i$ be the time customer $i$ arrived, and $d_i$ be the time customer $i$ departed. Take $a_0=d_0=0$. The M/G/1 queue model is described by:
\begin{align}
s_i &\sim \mathcal{U}\br{\theta_1,\theta_2} \\
a_i - a_{i-1} &\sim \mathrm{Exp}\br{\theta_3} \\
d_i - d_{i-1} &= s_i + \max{\br{0, a_i-d_{i-1}}}.
\end{align}
In our experiments we used $I=50$. The data $\x$ is $5$-dimensional, and is obtained by (a) calculating the $0$th, $25$th, $50$th, $75$th and $100$th quantiles of the set of inter-departure times $\set{d_i - d_{i-1}}_{1:I}$, and (b) linearly transforming the quantiles to have approximately zero mean and unit covariance matrix. The parameters of the linear transformation were determined by a pilot run. We took the ground truth parameters to be:
\begin{equation}
\p^* = \br{1,\;\; 5,\;\; 0.2},
\end{equation}
and simulated the model with parameters $\p^*$ to get observed data $\x_o$.

\subsection{Lotka--Volterra population model}

The Lotka--Volterra model \citep{Wilkinson:2011} is a Markov jump process describing the evolution of a population of predators interacting with a population of prey, and has four parameters $\p=\br{\theta_1, \ldots, \theta_4}$. 
Let $X$ be the number of predators, and $Y$ be the number of prey. According to the model, the following can take place:
\begin{itemize}
\item With rate $\exp\br{\theta_1}XY$ a predator may be born, increasing $X$ by one.
\item With rate $\exp\br{\theta_2}X$ a predator may die, decreasing $X$ by one.
\item With rate $\exp\br{\theta_3}Y$ a prey may be born, increasing $Y$ by one.
\item With rate $\exp\br{\theta_4}XY$ a prey may be eaten by a predator, decreasing $Y$ by one.
\end{itemize}
Our experimental setup follows that of \citet{Papamakarios:2016}. We used initial populations $X=50$ and $Y=100$. We simulated the model using the Gillespie algorithm \citep{Gillespie:1977} for a total of $30$ time units. We recorded the two populations every $0.2$ time units, which gives two timeseries of $151$ values each. The data $\x$ is $9$-dimensional, and corresponds to the following timeseries features:
\begin{itemize}
\item The mean of each timeseries.
\item The log variance of each timeseries.
\item The autocorrelation coefficient of each timeseries at lags $0.2$ and $0.4$ time units.
\item The cross-correlation coefficient between the two timeseries.
\end{itemize}
Each feature was normalized to have approximately zero mean and unit variance based on a pilot run. The ground truth parameters were taken to be:
\begin{equation}
\p^* = \br{\log{0.01},\;\; \log{0.5},\;\; \log{1},\;\; \log{0.01}},
\end{equation}
and the observed data $\x_o$ were generated from a simulation of the model at $\p^*$. In our experiments we used two priors: (a) a broad prior defined by:
\begin{equation}
p_\mathrm{broad}\br{\p} \propto \prod_{i=1}^4I\br{-5\le \theta_i\le 2},
\end{equation}
and (b) a prior corresponding to the oscillating regime, defined by:
\begin{equation}
p_\mathrm{osc}\br{\p} \propto \gaussianx{\p}{\p^*}{0.5^2} \,p_\mathrm{broad}\br{\p} .
\end{equation}

\subsection{Hodgkin--Huxley cortical pyramidal neuron model}

In neuroscience, the formalism developed by Hodgkin--Huxley in their
classic model of the squid giant axon \citep{Hodgkin:1952} is used to
model many different types of neuron. In our experiments, we used a
slightly modified version of a regular-spiking cortical pyramidal cell
\citep{Pospischil:2008}, for which NEURON \citep{Carnevale:2006}
simulation code is available in
ModelDB\@.\footnote{\url{https://senselab.med.yale.edu/ModelDB/ShowModel.cshtml?model=123623}}
The model is formulated as a set of five coupled ordinary differential
equations (ODEs) and describes how the electrical potential $V\br{t}$
measured across the neuronal cell membrane varies over time as a
function of current $I_\mathrm{e}\br{t}$ injected through an
electrode. In essence, the membrane is a capacitor punctuated by
conductances formed by multiple types of ion channel through which
currents flow. The currents charge and discharge the membrane
capacitance, causing the membrane potential to change, as described by
the first ODE, the membrane equation:
\begin{equation}
C_\mathrm{m}\frac{\diff{V}}{\diff{t}} = 
- I_\ell
- I_\mathrm{Na} - I_\mathrm{K} - I_\mathrm{M}
- I_\mathrm{e}.
\end{equation}
Here, $C_\mathrm{m}=\SI{1}{\micro\farad\per\square\centi\meter}$ is the
specific membrane capacitance, and $I_\ell$, $I_\mathrm{Na}$,
$I_\mathrm{K}$ and $I_\mathrm{M}$ are the ionic currents flowing
through `leak' channels, sodium channels, potassium
delayed-rectifier channels and M-type potassium channels respectively.
Each ionic current depends on a conductance that corresponds to how
many channels are open, and on the difference between the membrane
potential and an equilibrium potential. For example, for the leak
current:
\begin{equation}
I_\ell = g_\ell\,\br{V - E_\ell} , 
\end{equation}
where $g_\ell$ is the leak conductance, and $E_\ell$
is the leak equilibrium potential. Here the conductance
$g_\ell$ is constant through time, but for the sodium,
potassium and M-type channels, the conductances vary over time, as
described by the product of a fixed conductance and time-varying state
variables:
\begin{align}
I_\mathrm{Na} &= \overline{g}_\mathrm{Na}\, m^3h\, \br{V - E_\mathrm{Na}}\\
I_\mathrm{K} &= \overline{g}_\mathrm{K} \,n^4\, \br{V - E_\mathrm{K}}\label{eq:hh:IK}\\
I_\mathrm{M} &= \overline{g}_\mathrm{M} \,p \,\br{V - E_\mathrm{K}}.
\end{align}
Here, $\overline{g}_\mathrm{Na}$, $\overline{g}_\mathrm{K}$ and
$\overline{g}_\mathrm{M}$ are the per-channel maximum conductances,
$m$, $h$, $n$ and $p$ are state variables that range between 0 and 1,
and $E_\mathrm{Na}$ and $E_\mathrm{K}$ are the sodium and potassium
reversal potentials. The state variables evolve according to
differential equations of the form first introduced by Hodgkin and
Huxley \citep{Hodgkin:1952}:
\begin{align}
\frac{\diff{x}}{\diff{t}} &= \alpha_x \br{V} \br{1-x}
- \beta_x\br{V} \,x  \quad \text{for }x \in \left\{m, h, n\right\} \\
\frac{\diff{p}}{\diff{t}} &= \frac{p - p_\infty\br{V}}{\tau_\mathrm{p}\br{V}},
\end{align}
where $\alpha_x\br{V}$, $\beta_x\br{V}$, $p_\infty\br{V}$ and
$\tau_\mathrm{p}\br{V}$ are nonlinear functions of the membrane
potential. We use the published equations \citep{Pospischil:2008} for
$\alpha_\mathrm{m}\br{V}$, $\beta_\mathrm{m}\br{V}$, $\alpha_\mathrm{h}\br{V}$,
$\beta_\mathrm{h}\br{V}$ and $\alpha_\mathrm{n}\br{V}$, which contain a
parameter $V_\mathrm{T}$, and the published equations for $p_\infty\br{V}$
and $\tau_\mathrm{p}\br{V}$, the latter of which contains a parameter
$\tau_\mathrm{max}$. We use a generalized version of
$\beta_\mathrm{n}\br{V}$:
\begin{equation}
\beta_\mathrm{n}\br{V}= k_\mathrm{\beta n1}
\exp\br{-\frac{V-V_\mathrm{T}-10}{k_\mathrm{\beta n2}}}
\end{equation}
in which $k_\mathrm{\beta n1}$ and $k_\mathrm{\beta n2}$ are
adjustable parameters (rather than \SI{0.5}{\per\milli\second} and
\SI{40}{\milli\volt}). In order to simulate the model, we use NEURON \citep{Carnevale:2006} to solve the ODEs numerically
from initial conditions of:
\begin{equation}
m=h=n=p=0
\quad\text{and}\quad
V=\SI{-70}{\milli\volt},
\end{equation}
using the ``CNexp'' method and a time-step of \SI{25}{\micro\second}.
At each time step the injected current $I_\mathrm{e}$ is drawn from a
normal distribution with mean \SI{0.5}{\nano\ampere} and variance $\sigma^2$. The duration of the simulation is \SI{100}{\milli\second} and the voltage is recorded, which generates a timeseries of $4001$ voltage recordings.

Our inference setup follows \citet{Lueckmann:2017}. There are $12$ parameters $\p=\br{\theta_1, \ldots, \theta_{12}}$ to infer, defined as:
\begin{align}
\renewcommand{\arraystretch}{1.2}
\begin{array}{@{}l@{\hspace{3.4em}}l@{}}
\theta_1=\log\br{g_\ell} & \theta_{7\hphantom{0}}=\log\br{-E_\mathrm{K}} \\
\theta_2=\log\br{\overline{g}_\mathrm{Na}} & \theta_{8\hphantom{0}}=\log\br{-V_\mathrm{T}} \\
\theta_3=\log\br{\overline{g}_\mathrm{K}} & \theta_{9\hphantom{0}}=\log\br{k_\mathrm{\beta n1}} \\
\theta_4=\log\br{\overline{g}_\mathrm{M}} & \theta_{10}=\log\br{k_\mathrm{\beta n2}} \\
\theta_5=\log\br{-E_\ell} & \theta_{11}=\log\br{\tau_\mathrm{max}} \\
\theta_6=\log\br{E_\mathrm{Na}} & \theta_{12}=\log\br{\sigma}. \\ 
\end{array}
\end{align}
The data $\x$ is taken to be $18$ features of the voltage timeseries $V\br{t}$, in particular:
\begin{itemize}
\item The mean and log standard deviation of $V\br{t}$.
\item The normalized $3$rd, $5$th and $7$th moments of $V\br{t}$.
\item The logs of the normalized $4$th, $6$th and $8$th moments of $V\br{t}$.
\item The autocorrelation coefficients of $V\br{t}$ at lags $k\times$\SI{2.5}{\milli\second} for $k=1, \ldots, 10$.
\end{itemize}
The features are linearly transformed to have approximately zero mean and unit covariance matrix; the parameters of the transformation are calculated based on a pilot run. The ground truth parameters $\p^*$ are taken to be:
\begin{align}
\renewcommand{\arraystretch}{1.2}
\begin{array}{@{}l@{\hspace{2em}}l@{}}
\theta_1^*=\log\br{10^{-4}} & \theta_{7\hphantom{0}}^*=\log\br{100} \\
\theta_2^*=\log\br{0.2} & \theta_{8\hphantom{0}}^*=\log\br{60} \\
\theta_3^*=\log\br{0.05} & \theta_{9\hphantom{0}}^*=\log\br{0.5} \\
\theta_4^*=\log\br{7\times 10^{-5}} & \theta_{10}^*=\log\br{40} \\
\theta_5^*=\log\br{70} & \theta_{11}^*=\log\br{1000} \\
\theta_6^*=\log\br{50} & \theta_{12}^*=\log\br{1}. \\
\end{array}
\end{align}
The prior over parameters is:
\begin{equation}
\theta_i \sim \uniform{\theta_i^*-\log2}{\,\,\theta_i^*+\log 1.5}\quad\text{for}\quad i=1,\ldots,12.
\end{equation}
Finally, the observed data $\x_o$ were generated by simulating the model at $\p^*$.

\section{Full experimental results}
\label{sec:results}

In this section, we include the full set of experimental results. For each simulator model, we report:
\begin{itemize}
\item The approximate posterior computed by SNL\@.
\item The trade-off between accuracy and simulation cost. This is reported for all methods.
\item The full results of the simulation-based calibration test, consisting of one histogram per parameter.
\item The distance-based convergence diagnostic, i.e.~the distance between simulated and observed data vs the number of rounds. This is reported for SNL, SNPE-A and SNPE-B\@.
\item The goodness-of-fit diagnostic, i.e.~the Maximum Mean Discrepancy between simulated data and data generated from the likelihood model, for a given parameter value (we use the true parameters $\p^*$). We report this for SNL, NL and a baseline Gaussian fit.
\end{itemize}

\subsection{A toy problem with complex posterior}

\begin{figure*}[p]
\centering
\captionsetup[subfigure]{justification=centering}

\def\imwidth{0.085\textwidth}
\renewcommand{\arraystretch}{0}
\subfloat[MCMC samples from exact posterior. True parameters indicated in red.\label{fig:gauss:all:true_post}]{
\begin{tabular}{@{}c@{}c@{}c@{}c@{}c@{}}
\includegraphics[width=\imwidth]{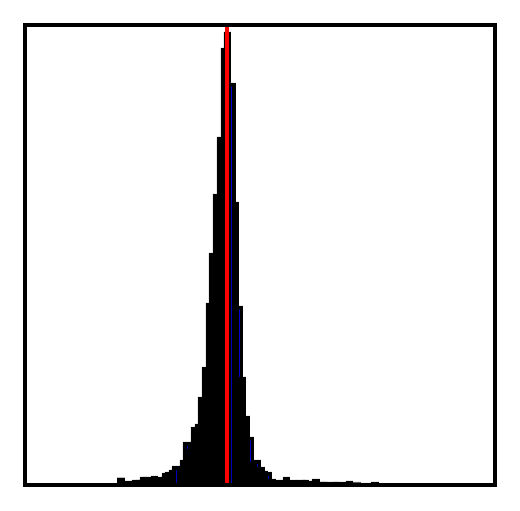} &
\includegraphics[width=\imwidth]{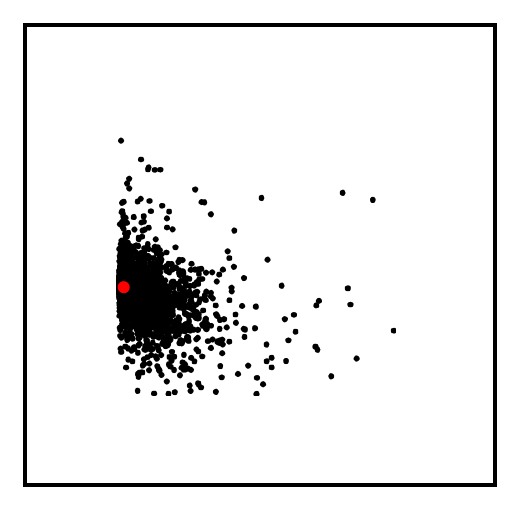} &
\includegraphics[width=\imwidth]{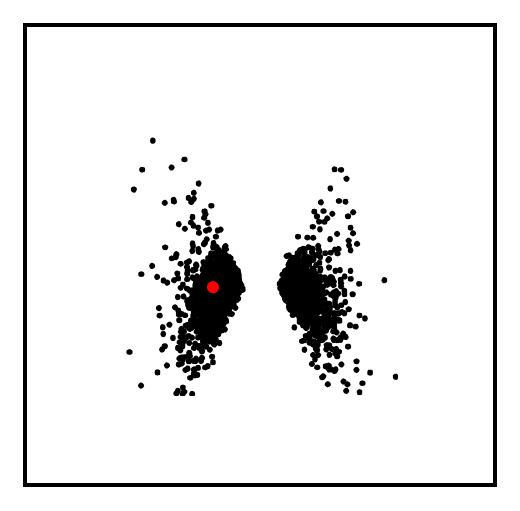} &
\includegraphics[width=\imwidth]{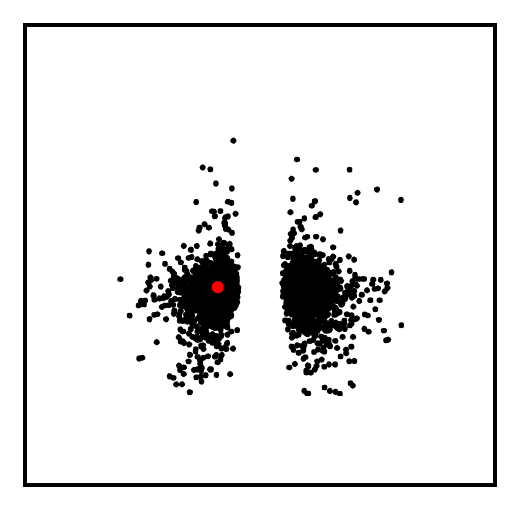} &
\includegraphics[width=\imwidth]{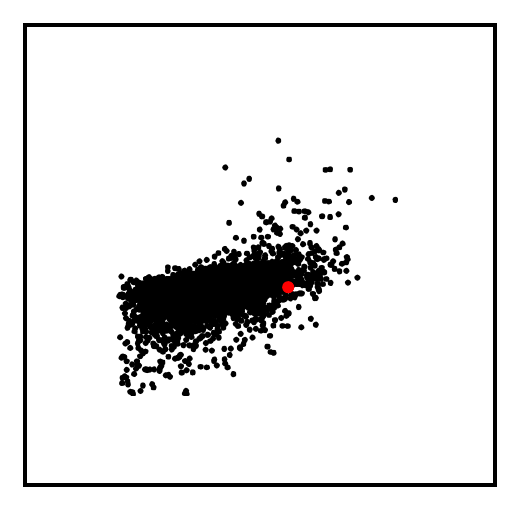} \\
\raisebox{3em}{$\theta_1$} &
\includegraphics[width=\imwidth]{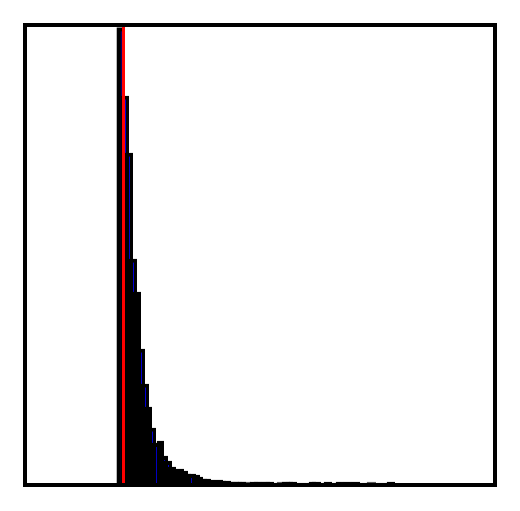} &
\includegraphics[width=\imwidth]{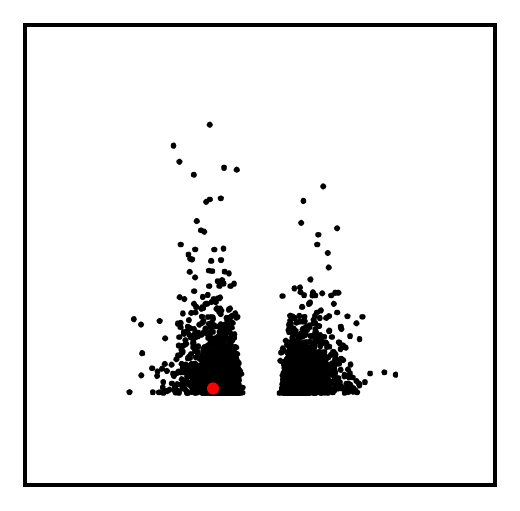} &
\includegraphics[width=\imwidth]{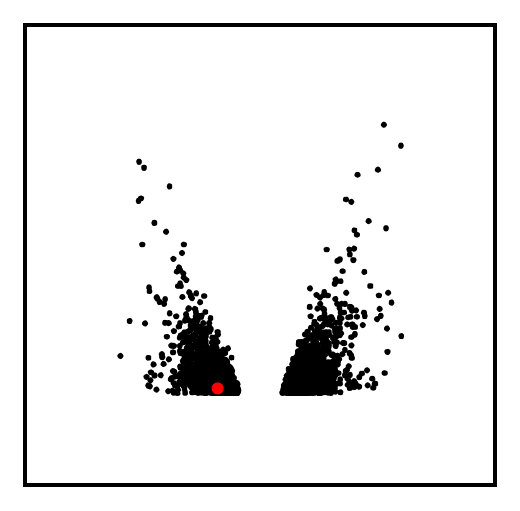} &
\includegraphics[width=\imwidth]{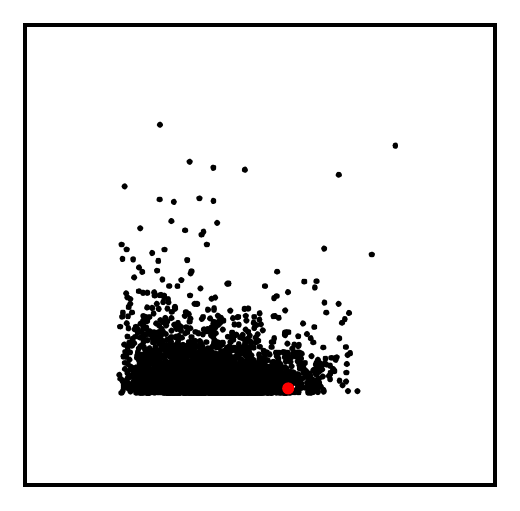} \\
& \raisebox{3em}{$\theta_2$} &
\includegraphics[width=\imwidth]{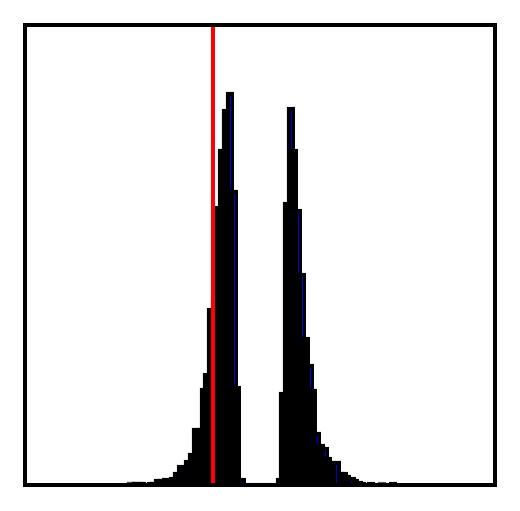} &
\includegraphics[width=\imwidth]{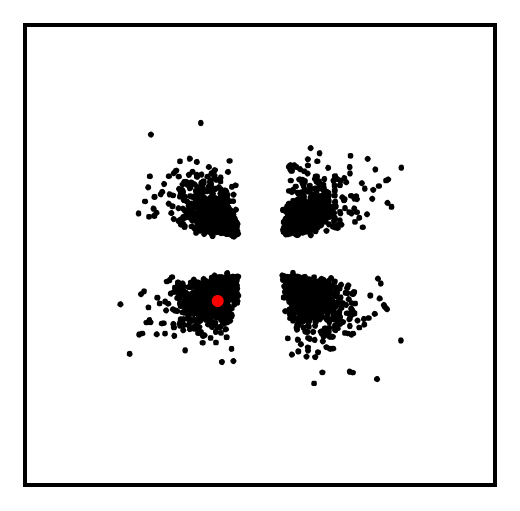} &
\includegraphics[width=\imwidth]{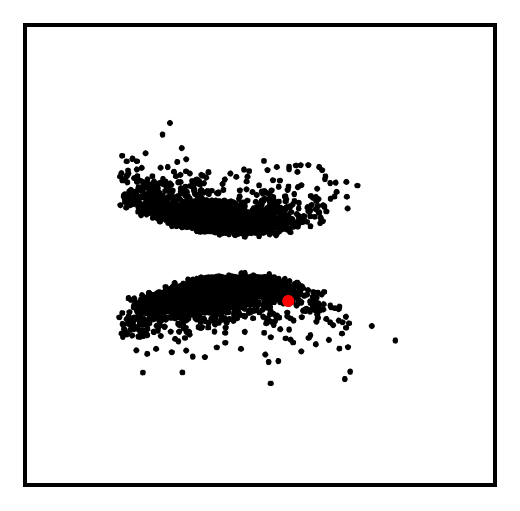} \\
& & \raisebox{3em}{$\theta_3$} &
\includegraphics[width=\imwidth]{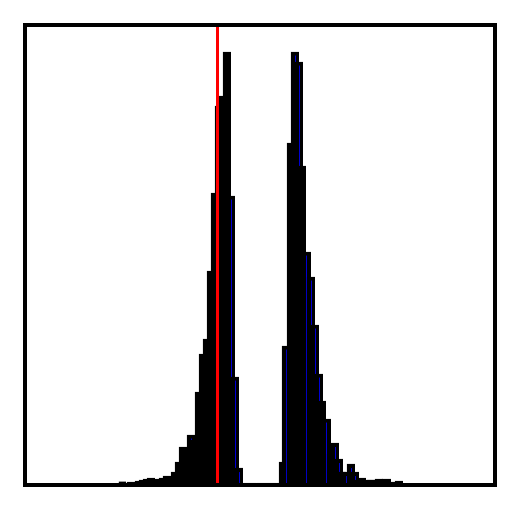} &
\includegraphics[width=\imwidth]{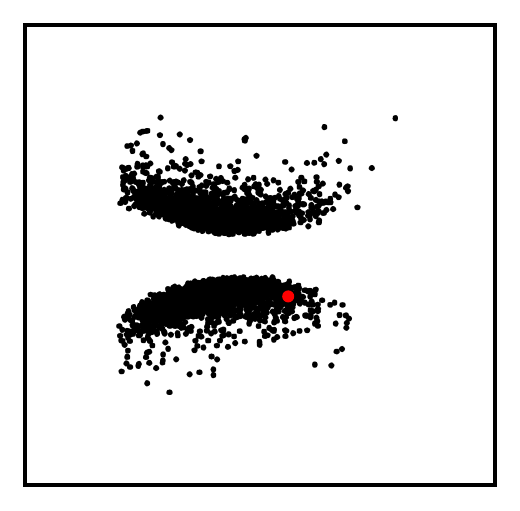} \\
& & & \raisebox{3em}{$\theta_4$} &
\includegraphics[width=\imwidth]{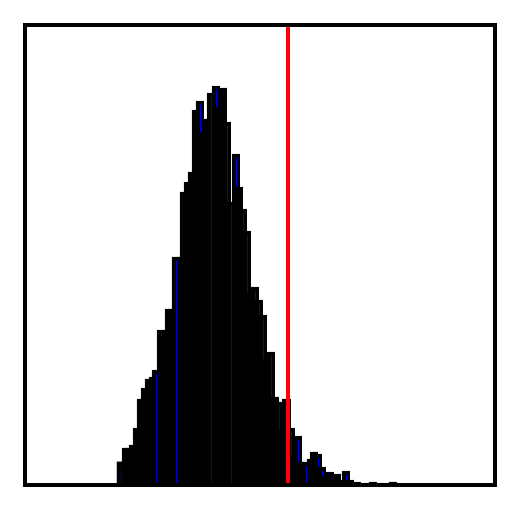} \\
& & & & $\theta_5$
\end{tabular}
}\hspace{0.4em}
\subfloat[MCMC samples from SNL posterior. True parameters indicated in red.\label{fig:gauss:all:snl_post}]{
\begin{tabular}{@{}c@{}c@{}c@{}c@{}c@{}}
\includegraphics[width=\imwidth]{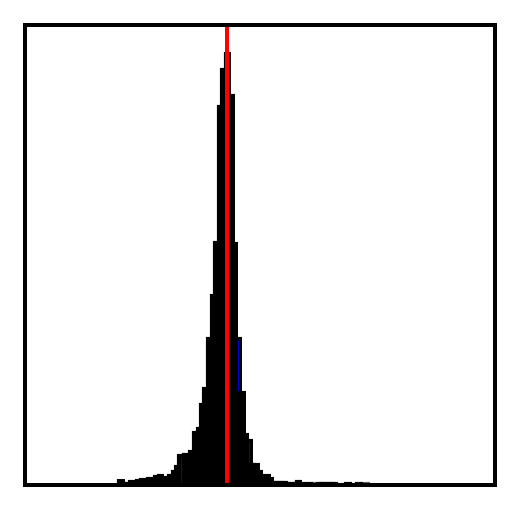} &
\includegraphics[width=\imwidth]{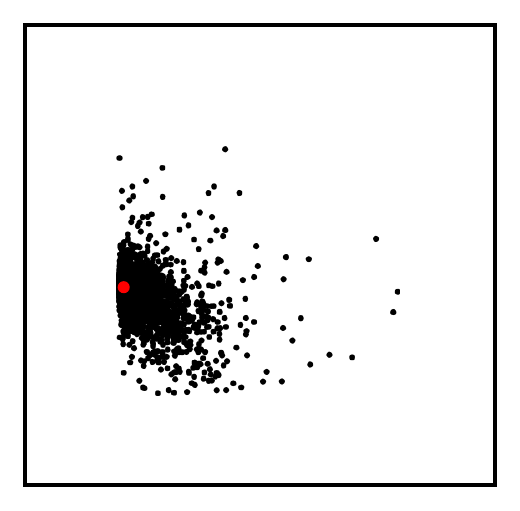} &
\includegraphics[width=\imwidth]{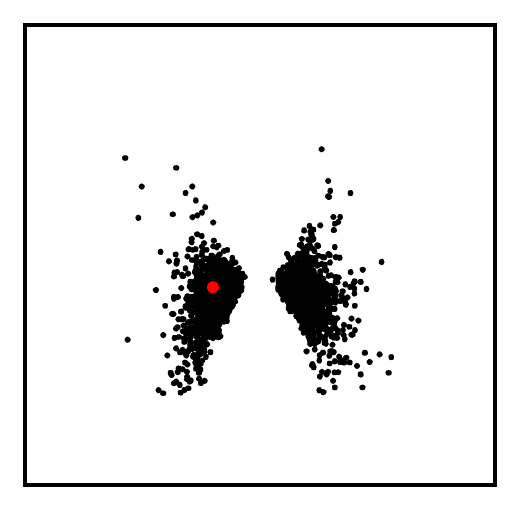} &
\includegraphics[width=\imwidth]{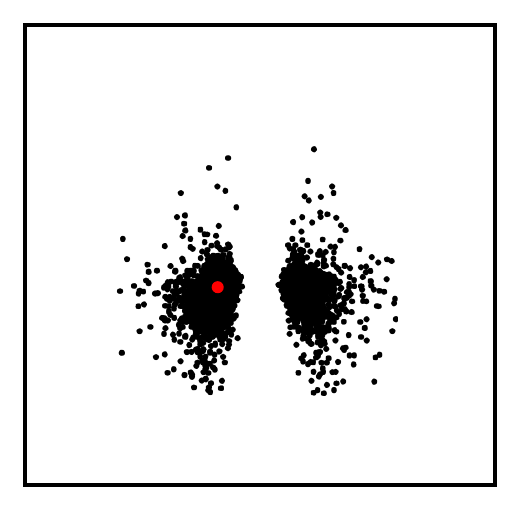} &
\includegraphics[width=\imwidth]{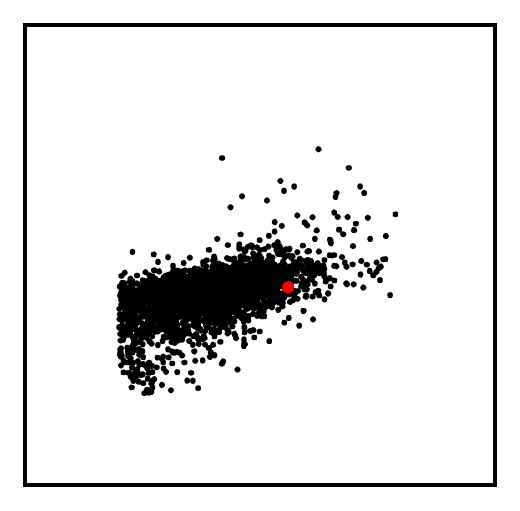} \\
\raisebox{3em}{$\theta_1$} &
\includegraphics[width=\imwidth]{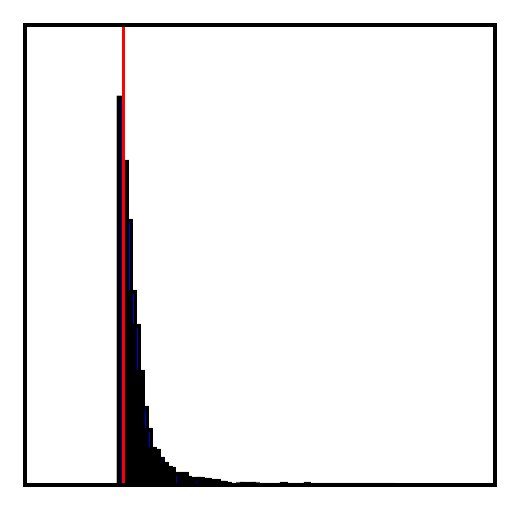} &
\includegraphics[width=\imwidth]{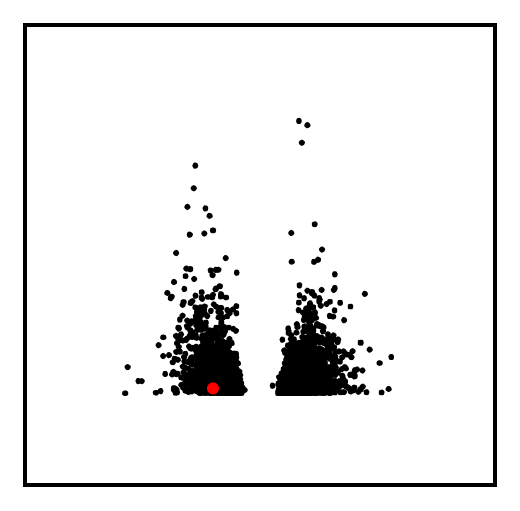} &
\includegraphics[width=\imwidth]{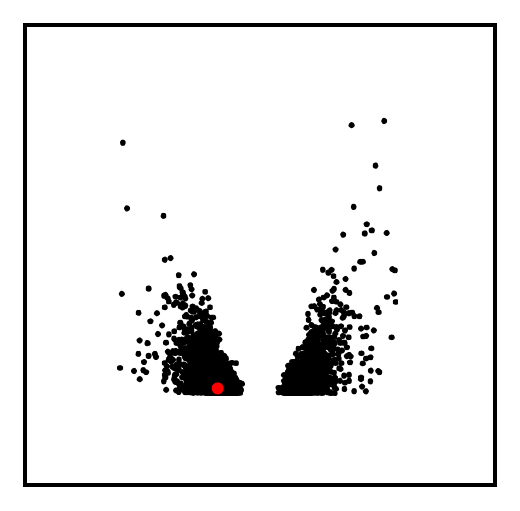} &
\includegraphics[width=\imwidth]{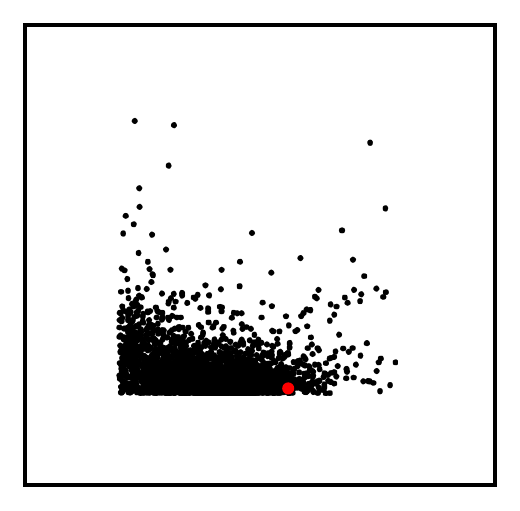} \\
& \raisebox{3em}{$\theta_2$} &
\includegraphics[width=\imwidth]{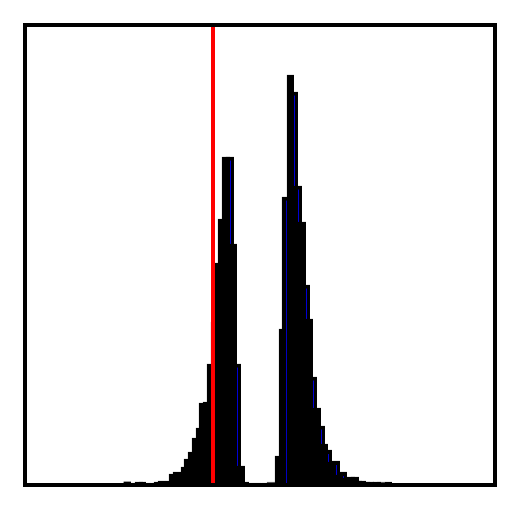} &
\includegraphics[width=\imwidth]{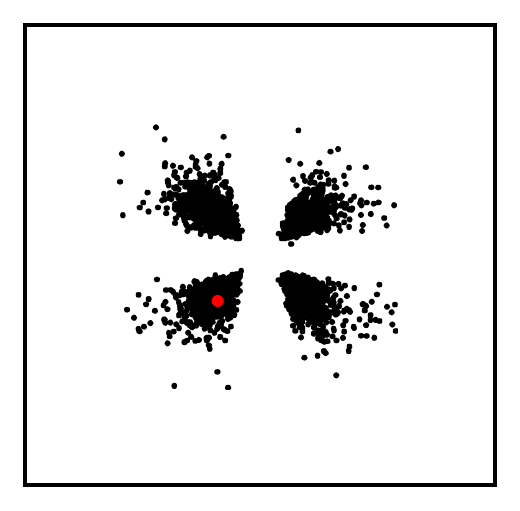} &
\includegraphics[width=\imwidth]{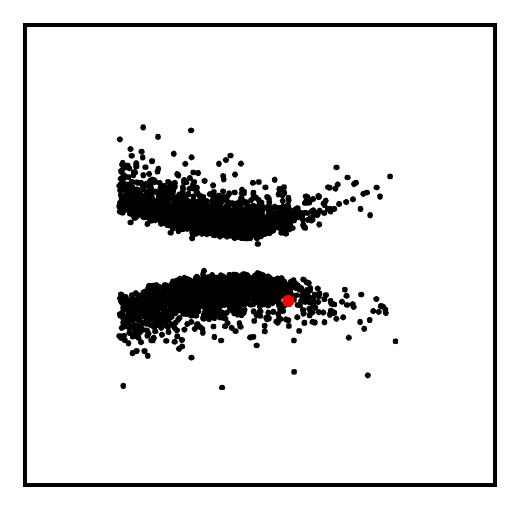} \\
& & \raisebox{3em}{$\theta_3$} &
\includegraphics[width=\imwidth]{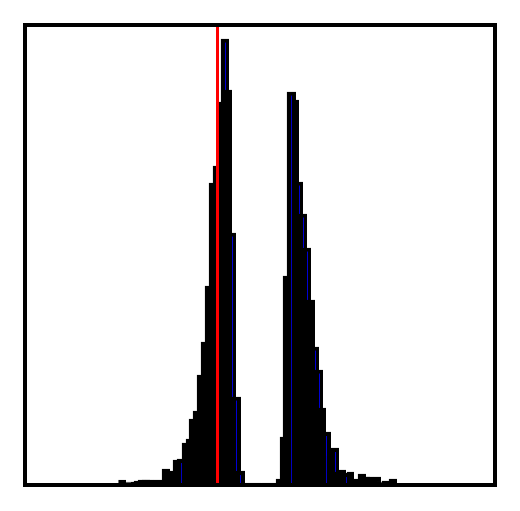} &
\includegraphics[width=\imwidth]{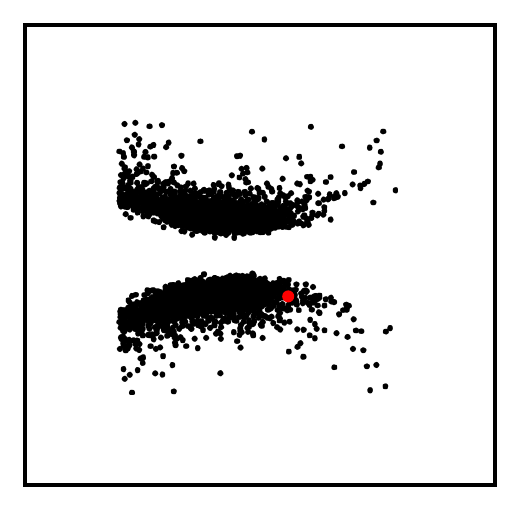} \\
& & & \raisebox{3em}{$\theta_4$} &
\includegraphics[width=\imwidth]{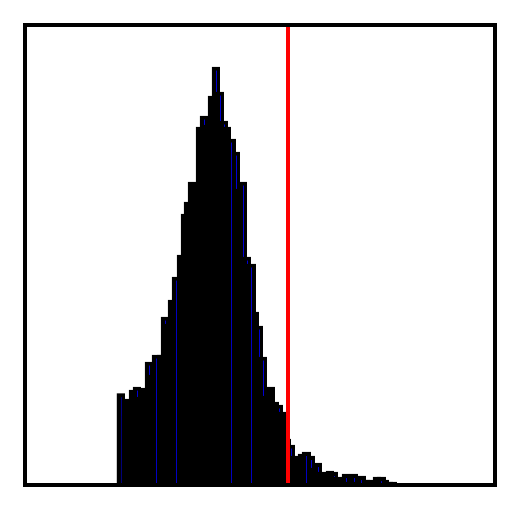} \\
& & & & $\theta_5$
\end{tabular}
}\\[0.6em]

\subfloat[Maximum Mean Discrepancy vs simulation cost.]{
\includegraphics[width=0.4\textwidth]{figs/gauss/mmd.pdf}
}\hspace{0.35em}
\subfloat[Minus log probability of true parameters vs simulation cost.]{
\includegraphics[width=0.4\textwidth]{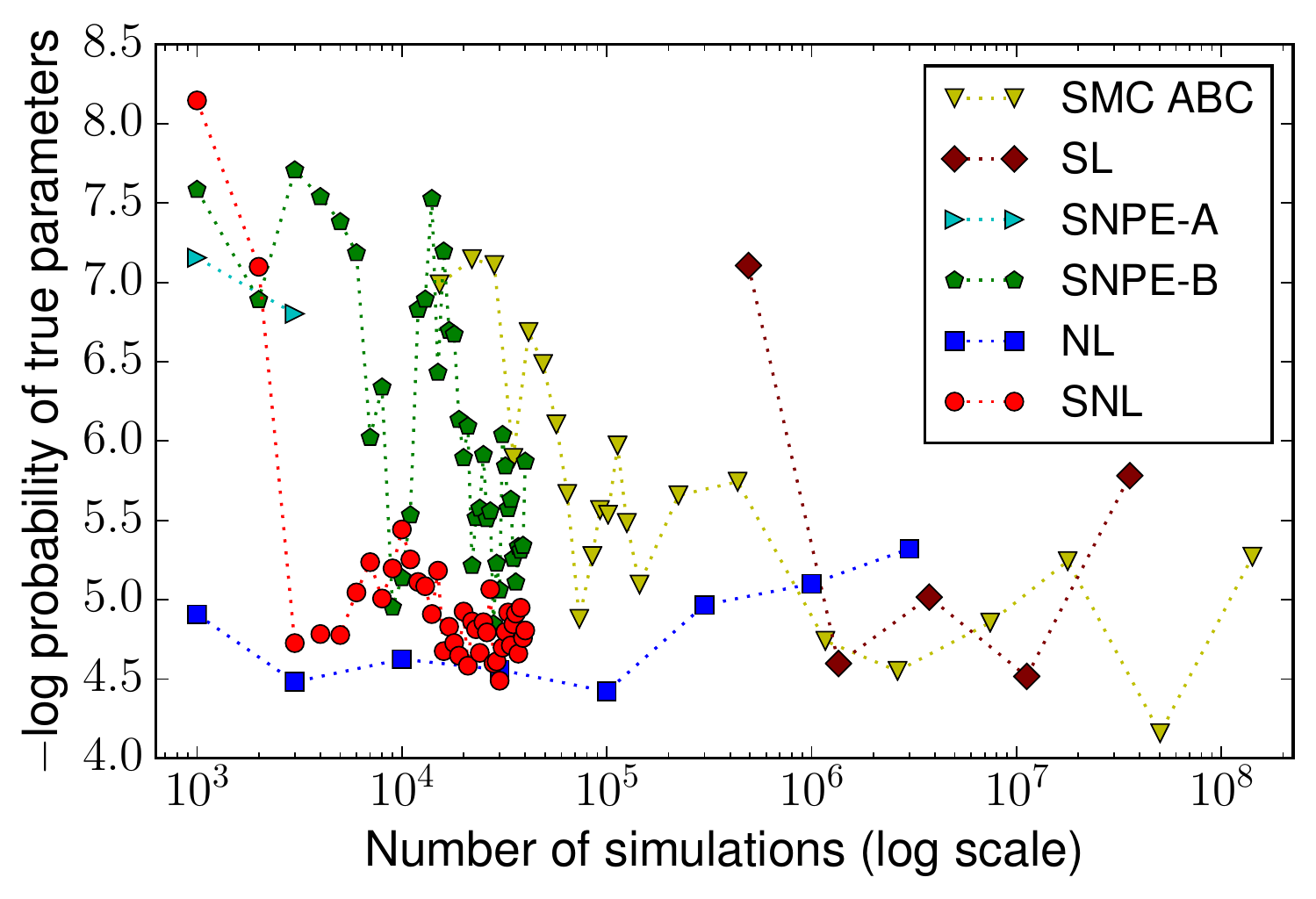}
}\\[1.5em]

\subfloat[Calibration test for SNL\@. Histogram outside gray band indicates poor calibration.]{
\begin{tabular}{@{}c@{}c@{}c@{}c@{}c@{}}
\includegraphics[width=0.14\textwidth]{figs/gauss/calib_likmcmc_original_1.pdf} &
\includegraphics[width=0.14\textwidth]{figs/gauss/calib_likmcmc_original_2.pdf} &
\includegraphics[width=0.14\textwidth]{figs/gauss/calib_likmcmc_original_3.pdf} &
\includegraphics[width=0.14\textwidth]{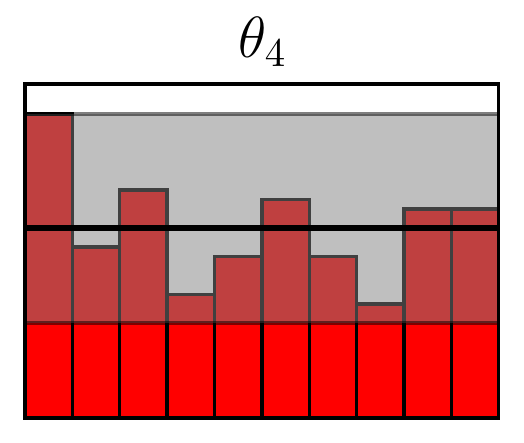} &
\includegraphics[width=0.14\textwidth]{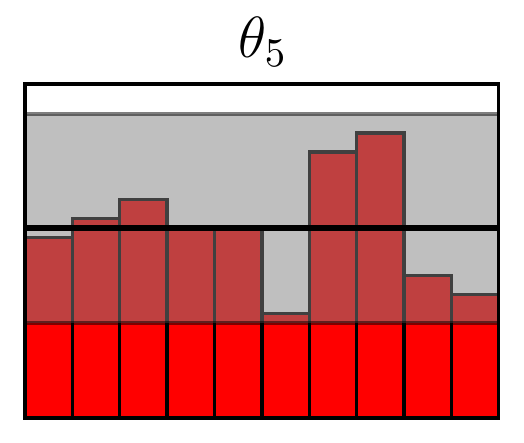} 
\end{tabular}
}\\[1.2em]

\subfloat[Median distance from simulated to observed data.]{
\includegraphics[width=0.4\textwidth]{figs/gauss/dist_median.pdf}
}
\subfloat[Likelihood goodness-of-fit vs simulation cost, calculated at true parameters.]{
\includegraphics[width=0.4\textwidth]{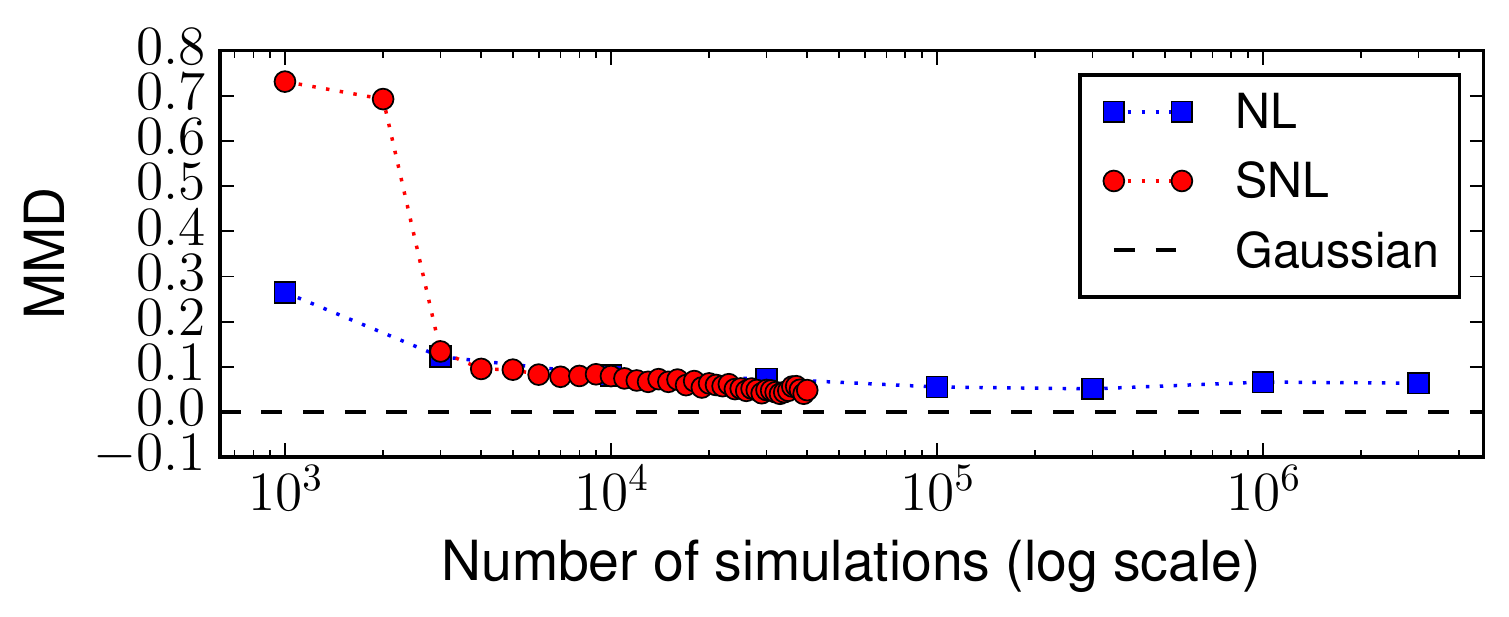}
}\\[1em]

\caption{A toy model with complex posterior.}
\label{fig:gauss:all}
\end{figure*}

Figure~\ref{fig:gauss:all} shows the results.
The exact posterior $\prob{\p\g\x_o}$ is plotted in Figure~\ref{fig:gauss:all:true_post}. Even though the prior is uniform and the likelihood is Gaussian, the posterior is complex and non-trivial: it has four symmetric modes due to the two squaring operations in Equations~\eqref{eq:gauss:sq1} and \eqref{eq:gauss:sq2}, and vertical cut-offs due to the hard constraints imposed by the prior. We can see that the SNL posterior (Figure~\ref{fig:gauss:all:snl_post}) approximates the exact posterior well. 

\subsection{M/G/1 queue model}

\begin{figure*}[p]
\centering
\captionsetup[subfigure]{justification=centering}

\subfloat[MCMC samples from SNL posterior. True parameters indicated in red.\label{fig:mg1:all:snl_post}]{
\def\imwidth{0.098\textwidth}
\renewcommand{\arraystretch}{0}
\begin{tabular}{@{}c@{}c@{}c@{}}
\includegraphics[width=\imwidth]{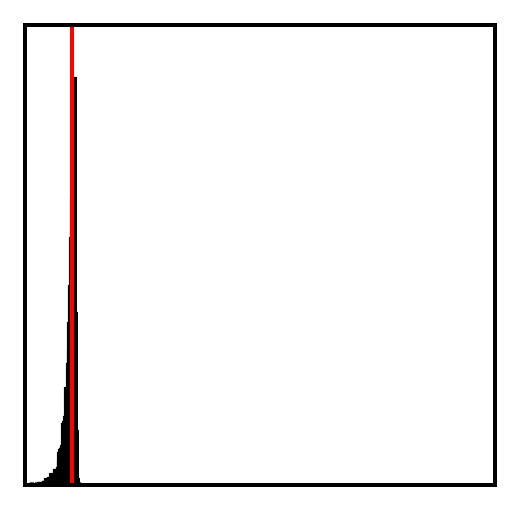} &
\includegraphics[width=\imwidth]{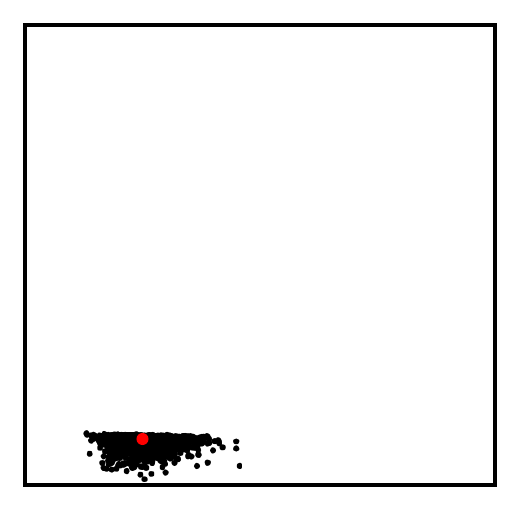} &
\includegraphics[width=\imwidth]{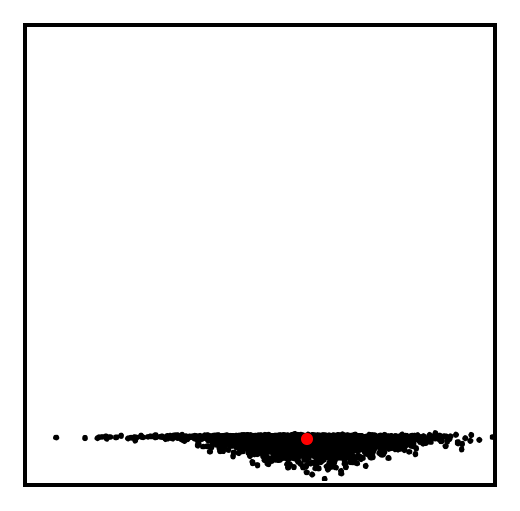} \\
\raisebox{4.38em}{$\theta_1$} &
\includegraphics[width=\imwidth]{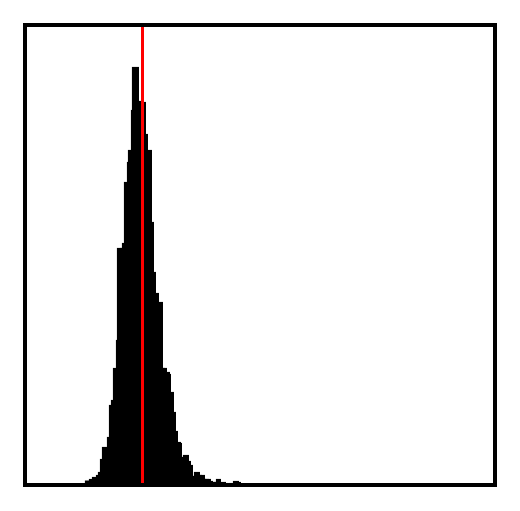} &
\includegraphics[width=\imwidth]{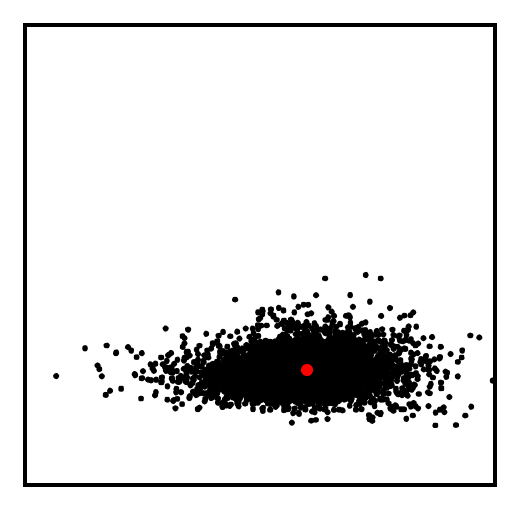} \\
& \raisebox{4.38em}{$\theta_2$} &
\includegraphics[width=\imwidth]{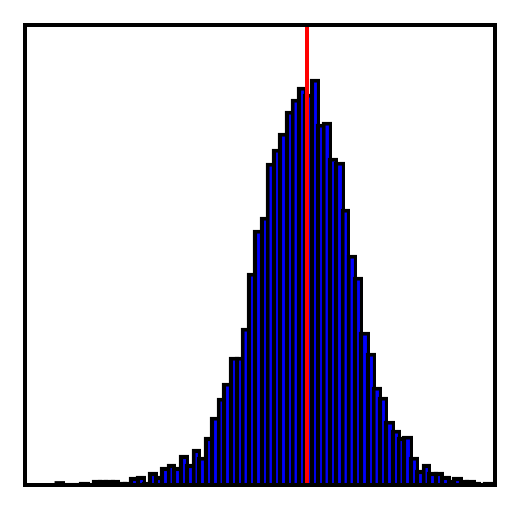} \\
& & $\theta_3$
\end{tabular}
}\\[1.8em]

\subfloat[Minus log probability of true parameters vs simulation cost.]{
\includegraphics[width=0.46\textwidth]{figs/mg1/loglik.pdf}
}\\[2.5em]

\subfloat[Calibration test for SNL\@. Histogram outside gray band indicates poor calibration.]{
\begin{tabular}{@{}c@{}c@{}c@{}}
\includegraphics[width=0.16\textwidth]{figs/mg1/calib_likmcmc_original_1.pdf} &
\includegraphics[width=0.16\textwidth]{figs/mg1/calib_likmcmc_original_2.pdf} &
\includegraphics[width=0.16\textwidth]{figs/mg1/calib_likmcmc_original_3.pdf}
\end{tabular}
}\\[1.4em]

\subfloat[Median distance from simulated to observed data.]{
\includegraphics[width=0.42\textwidth]{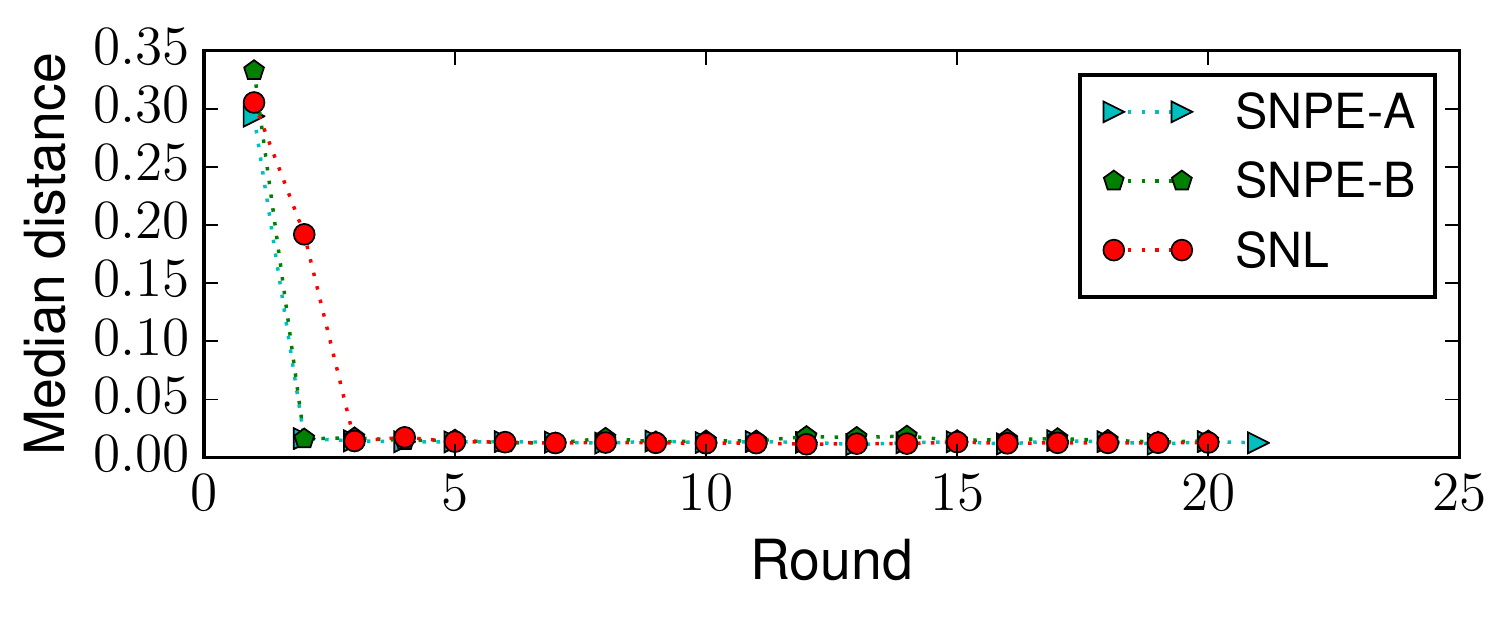}
}
\subfloat[Likelihood goodness-of-fit vs simulation cost, calculated at true parameters.]{
\includegraphics[width=0.42\textwidth]{figs/mg1/gof_mmd.pdf}
}\\[1.3em]

\caption{M/G/1 queue model.}
\label{fig:mg1:all}
\end{figure*}

Figure~\ref{fig:mg1:all} shows the results. The SNL posterior is shown in Figure~\ref{fig:mg1:all:snl_post}. We can see that the posterior is concentrated around the true parameters. Parameter $\theta_1$ is particularly well constrained. From the description of the model in Equations~\eqref{eq:mg1:s}--\eqref{eq:mg1:d}, it directly follows that:
\begin{equation}
\theta_1 \le {\min}_i\,\br{d_i - d_{i-1}}.
\end{equation}
The data $\vect{x}$ is an invertible linear transformation of the quantiles of $\set{d_i - d_{i-1}}_{1:I}$ including the $0th$ quantile, which is precisely equal to $\min_i\,\br{d_i - d_{i-1}}$. Hence, the data $\vect{x}$ imposes a hard constraint on the maximum possible value of $\theta_1$, as correctly captured by the SNL posterior. On the other hand, the data is less informative about $\theta_2$ and $\theta_3$, hence the parameters are less constrained.

\subsection{Lotka--Volterra population model}

\begin{figure*}[p]
\centering
\captionsetup[subfigure]{justification=centering}

\subfloat[MCMC samples from SNL posterior. True parameters indicated in red.\label{fig:lv:all:snl_post}]{
\def\imwidth{0.08\textwidth}
\renewcommand{\arraystretch}{0}
\begin{tabular}{@{}c@{}c@{}c@{}c@{}}
\includegraphics[width=\imwidth]{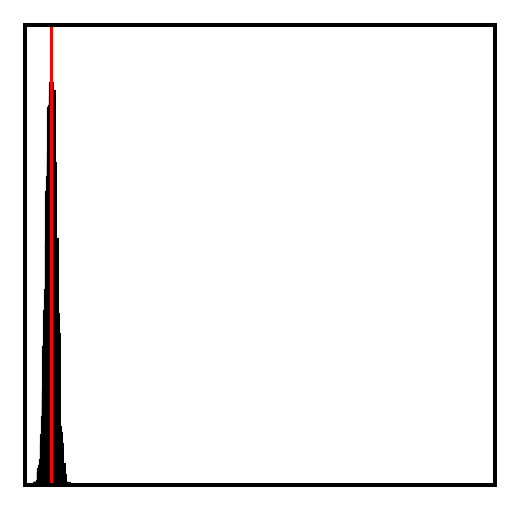} &
\includegraphics[width=\imwidth]{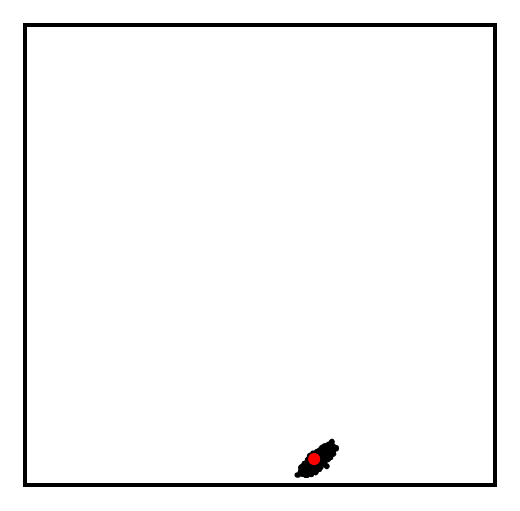} &
\includegraphics[width=\imwidth]{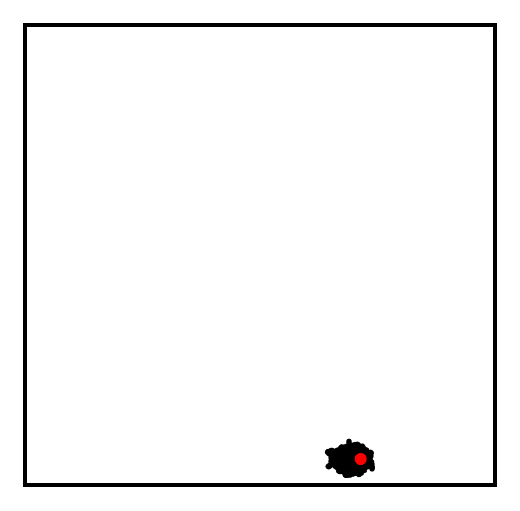} &
\includegraphics[width=\imwidth]{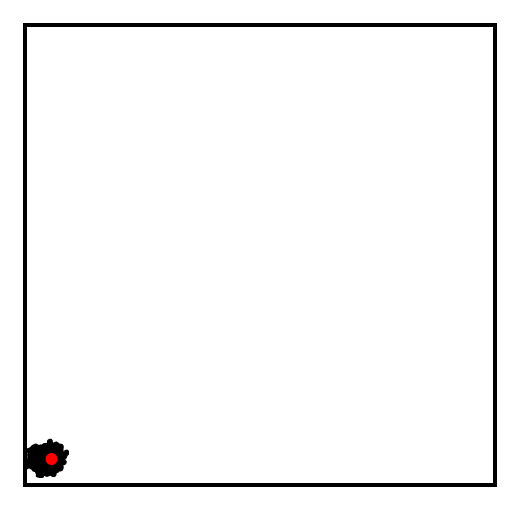} \\
\raisebox{3.35em}{$\theta_1$} &
\includegraphics[width=\imwidth]{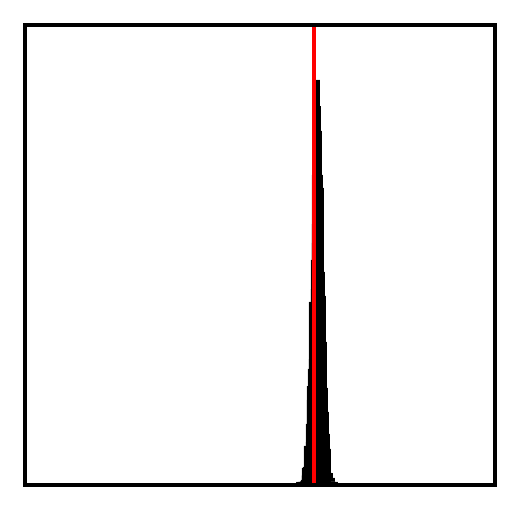} &
\includegraphics[width=\imwidth]{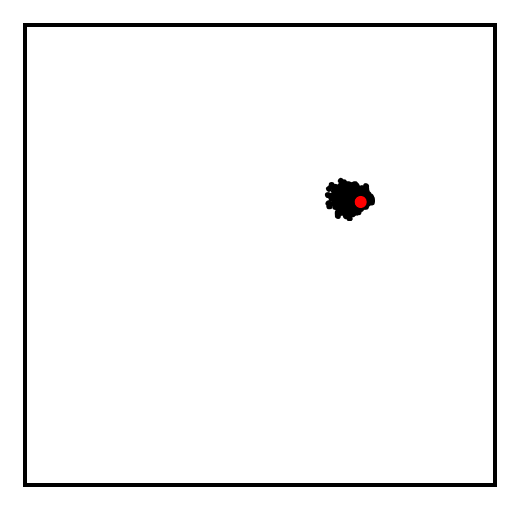} &
\includegraphics[width=\imwidth]{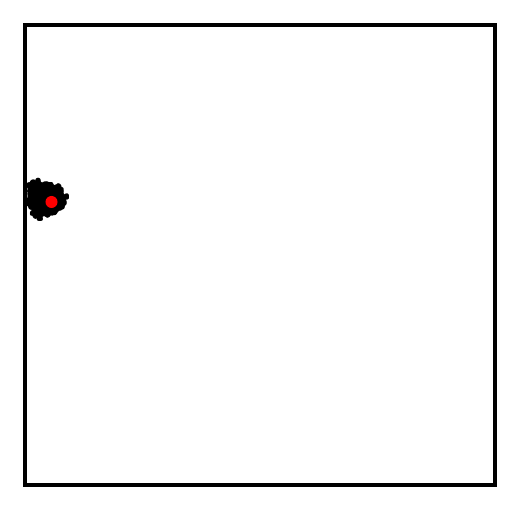} \\
& \raisebox{3.35em}{$\theta_2$} &
\includegraphics[width=\imwidth]{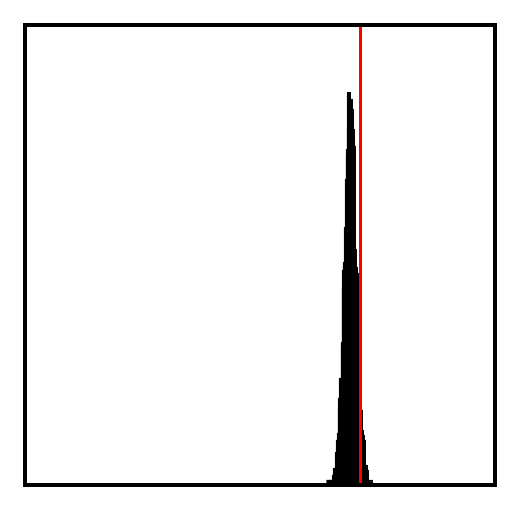} &
\includegraphics[width=\imwidth]{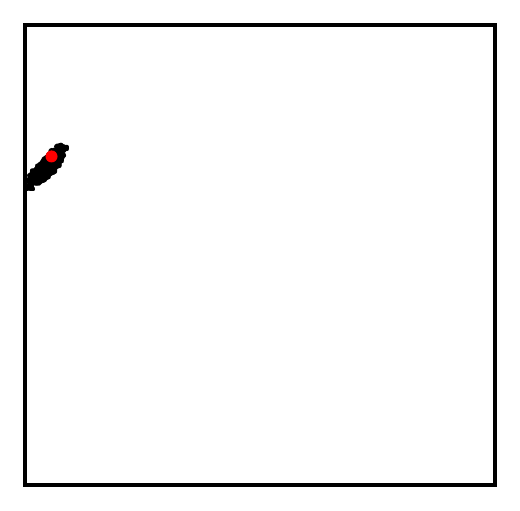} \\
& & \raisebox{3.35em}{$\theta_3$} &
\includegraphics[width=\imwidth]{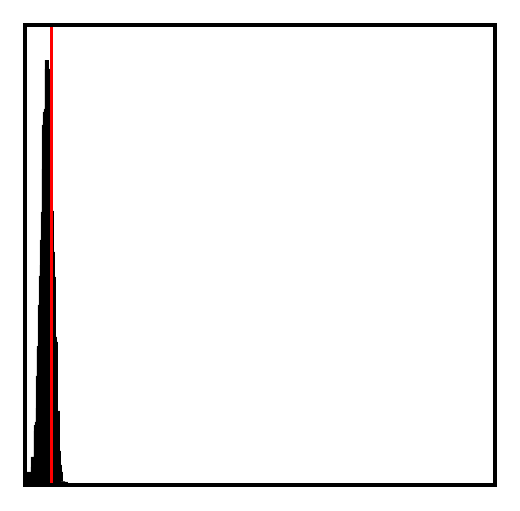} \\
& & & $\theta_4$
\end{tabular}
}

\vspace{-0.25em}
\subfloat[Minus log probability of true parameters vs simulation cost.]{
\includegraphics[width=0.44\textwidth]{figs/lv/loglik.pdf}
}\\[0.3em]

\subfloat[Calibration test for SNL\@, oscillating regime.]{
\begin{tabular}{@{}c@{}c@{}}
\includegraphics[width=0.15\textwidth]{figs/lv/calib_likmcmc_near_truth_1.pdf} &
\includegraphics[width=0.15\textwidth]{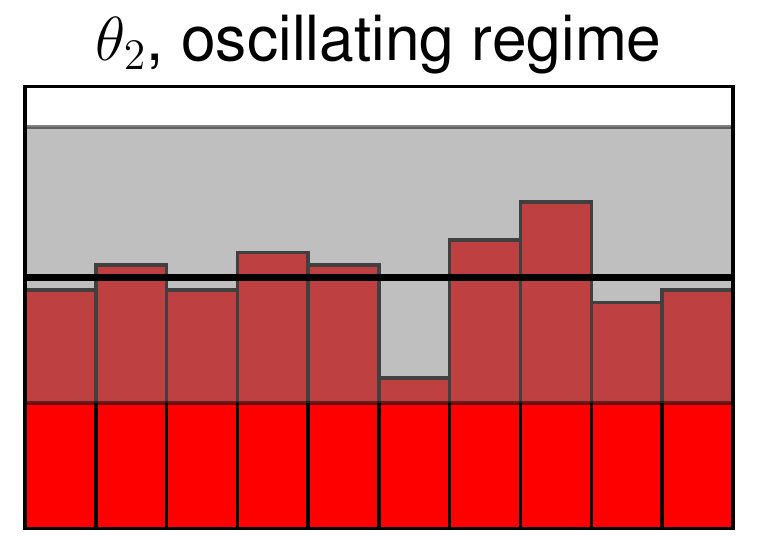} \\
\includegraphics[width=0.15\textwidth]{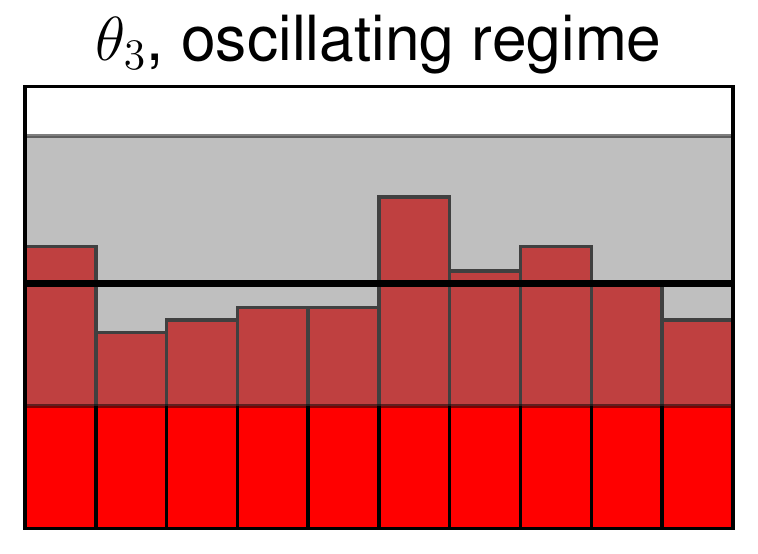} &
\includegraphics[width=0.15\textwidth]{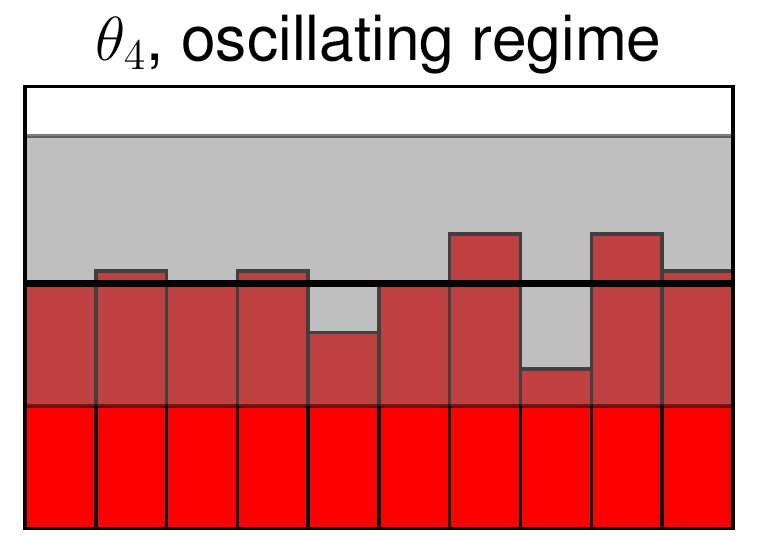}
\end{tabular}
}\hspace{4em}
\subfloat[Calibration test for SNL\@, broad prior.]{
\begin{tabular}{@{}c@{}c@{}}
\includegraphics[width=0.15\textwidth]{figs/lv/calib_likmcmc_original_1.pdf} &
\includegraphics[width=0.15\textwidth]{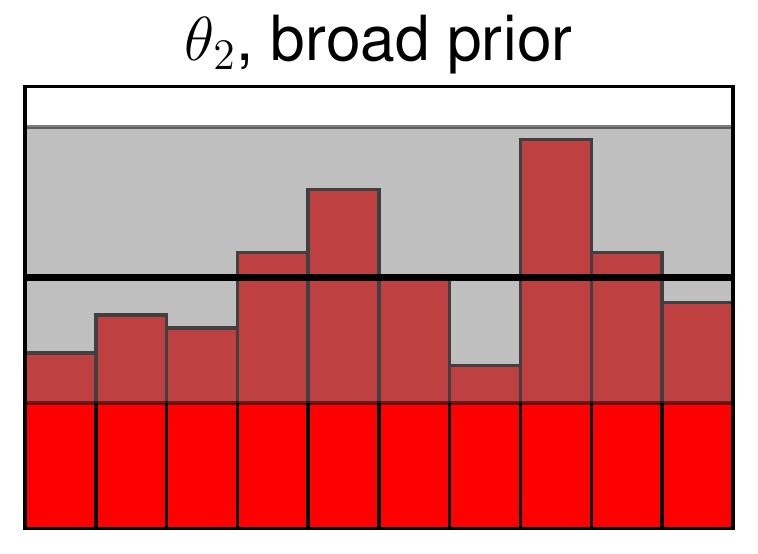} \\
\includegraphics[width=0.15\textwidth]{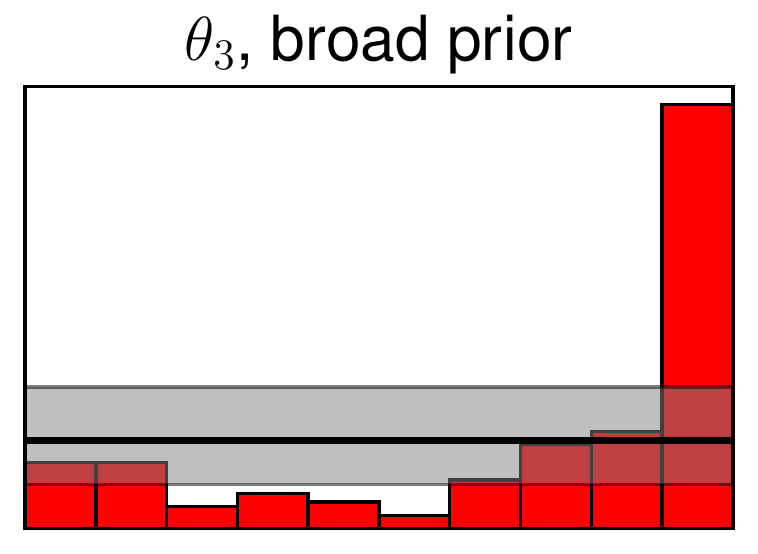} &
\includegraphics[width=0.15\textwidth]{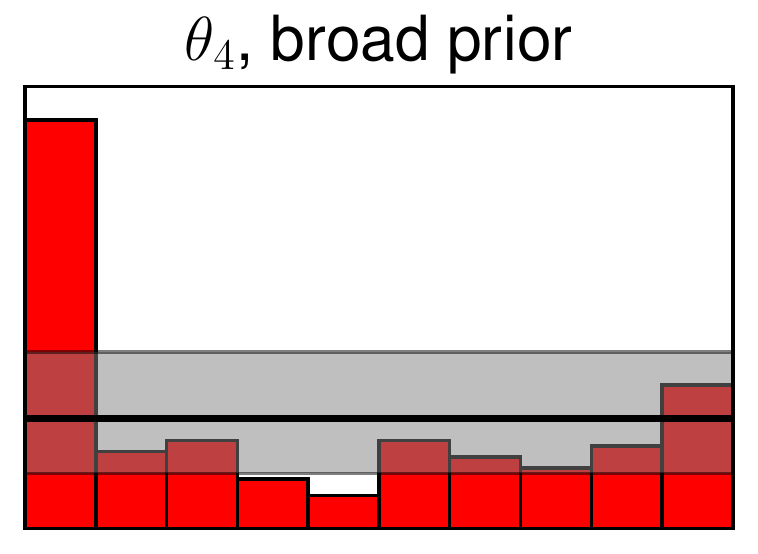}
\end{tabular}
}

\vspace{-0.25em}
\subfloat[Median distance from simulated to observed data.]{
\includegraphics[width=0.44\textwidth]{figs/lv/dist_median.pdf}
}
\subfloat[Likelihood goodness-of-fit vs simulation cost, calculated at true parameters.]{
\includegraphics[width=0.44\textwidth]{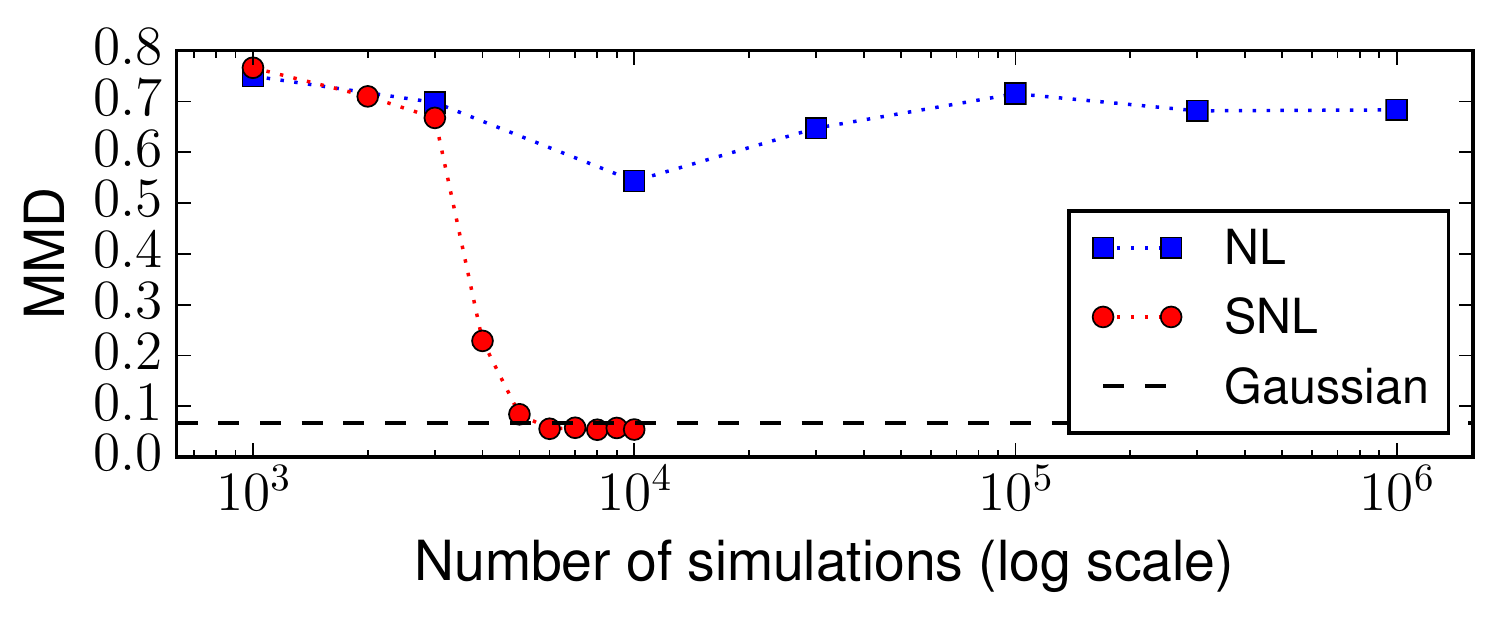}
}\\[0.1em]

\caption{Lotka--Volterra population model.}
\label{fig:lv:all}
\end{figure*}

Figure~\ref{fig:lv:all} shows the results. The SNL posterior is shown in Figure~\ref{fig:lv:all:snl_post}; the extent of each plot corresponds to the broad prior $p_\mathrm{broad}\br{\p}$. We can see that the posterior is tightly concentrated around the true parameters, which suggests that the data $\vect{x}$ is highly informative about the parameters $\p$.

\subsection{Hodgkin--Huxley cortical pyramidal neuron model}

\begin{figure*}[p]
\centering
\captionsetup[subfigure]{justification=centering}

\subfloat[Minus log probability of true parameters vs simulation cost.]{
\includegraphics[width=0.46\textwidth]{figs/hh/loglik.pdf}
}\\[3em]

\subfloat[Calibration test for SNL\@. Histogram outside gray band indicates poor calibration.]{
\begin{tabular}{@{}c@{}c@{}c@{}c@{}}
\includegraphics[width=0.16\textwidth]{figs/hh/calib_likmcmc_original_1.pdf} &
\includegraphics[width=0.16\textwidth]{figs/hh/calib_likmcmc_original_2.pdf} &
\includegraphics[width=0.16\textwidth]{figs/hh/calib_likmcmc_original_3.pdf} &
\includegraphics[width=0.16\textwidth]{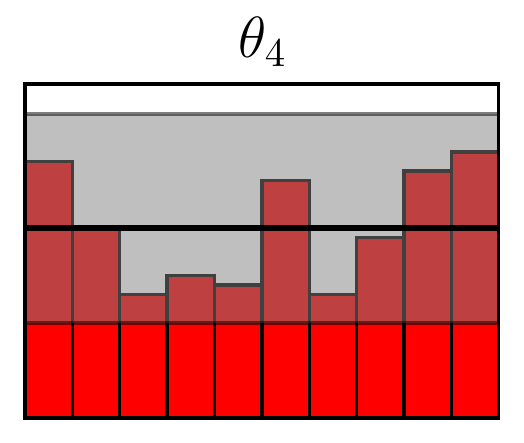} \\
\includegraphics[width=0.16\textwidth]{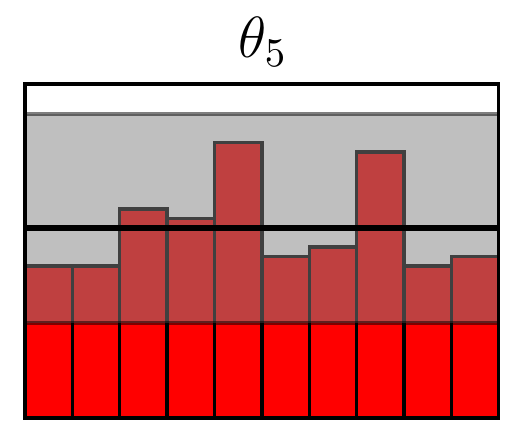} &
\includegraphics[width=0.16\textwidth]{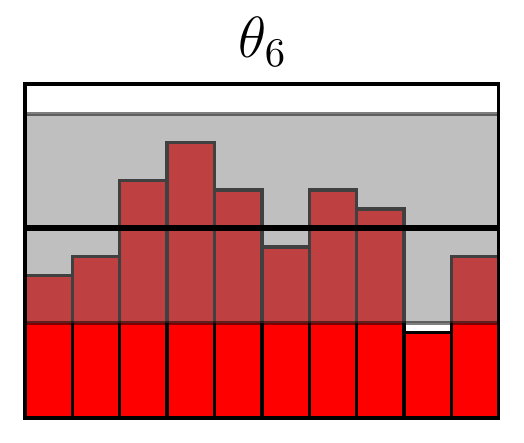} &
\includegraphics[width=0.16\textwidth]{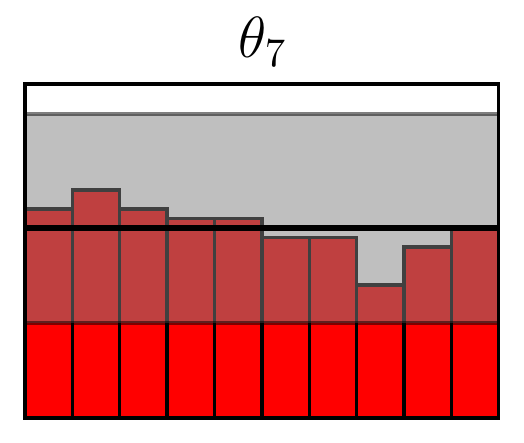} &
\includegraphics[width=0.16\textwidth]{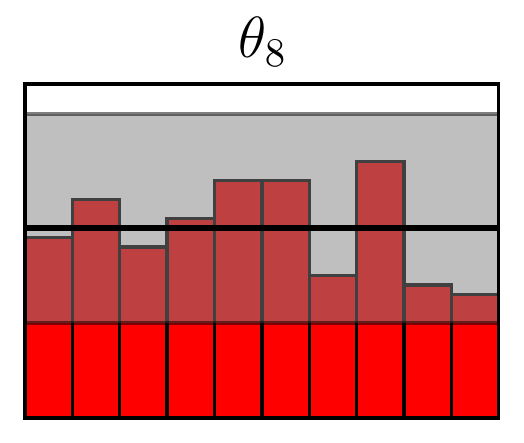} \\
\includegraphics[width=0.16\textwidth]{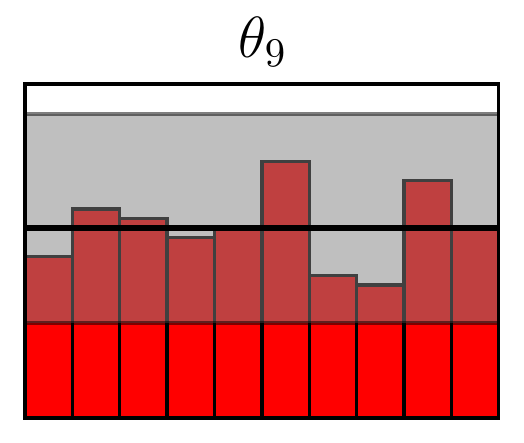} &
\includegraphics[width=0.16\textwidth]{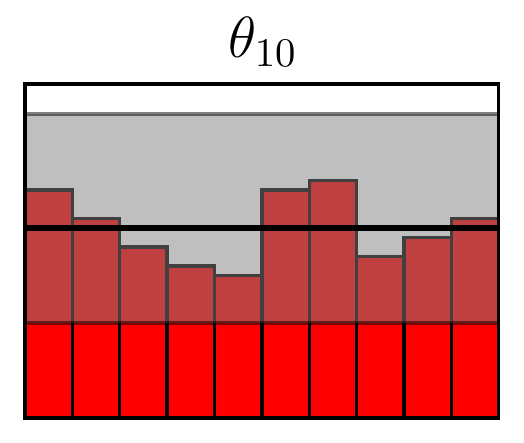} &
\includegraphics[width=0.16\textwidth]{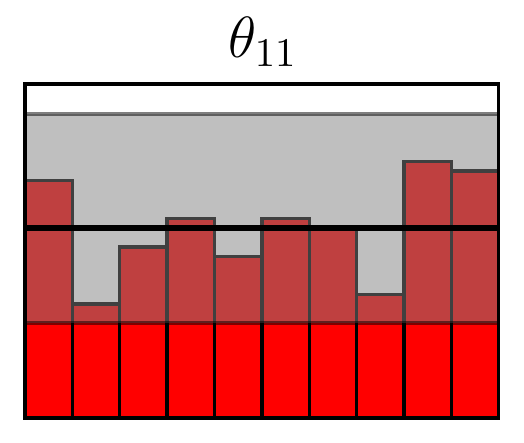} &
\includegraphics[width=0.16\textwidth]{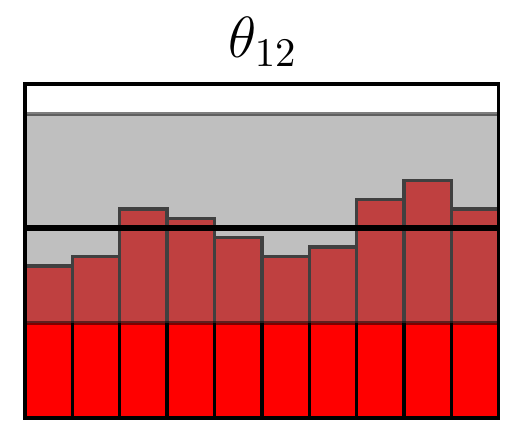} 
\end{tabular}
}\\[2em]

\subfloat[Median distance from simulated to observed data.]{
\includegraphics[width=0.45\textwidth]{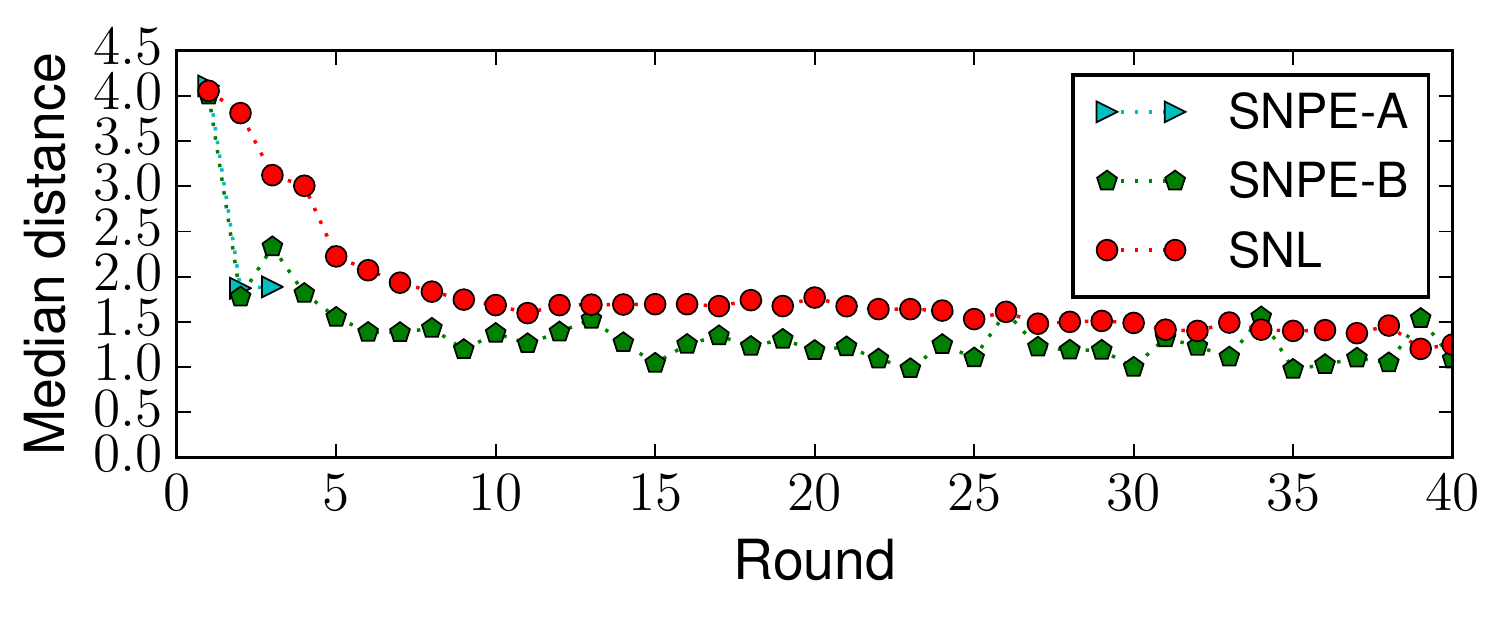}
}
\subfloat[Likelihood goodness-of-fit vs simulation cost, calculated at true parameters.]{
\includegraphics[width=0.45\textwidth]{figs/hh/gof_mmd.pdf}
}\\[2em]

\caption{Hodgkin--Huxley neuron model.}
\label{fig:hh:all}
\end{figure*}

\begin{figure*}[p]
\centering
\def\imwidth{0.08\textwidth}
\def\raiseheight{3.7em}
\renewcommand{\arraystretch}{0}

\begin{tabular}{@{}c@{}c@{}c@{}c@{}c@{}c@{}c@{}c@{}c@{}c@{}c@{}c@{}}
\includegraphics[width=\imwidth]{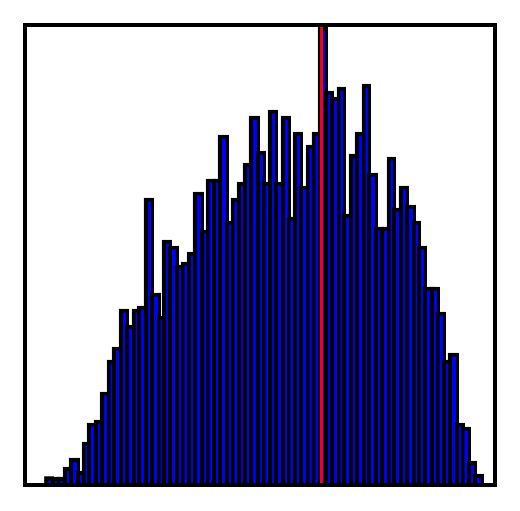} &
\includegraphics[width=\imwidth]{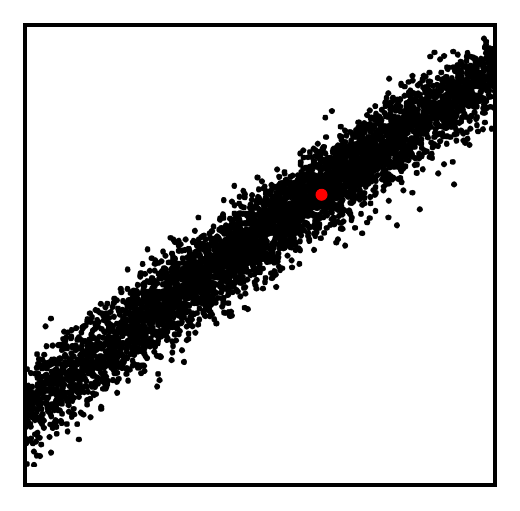} &
\includegraphics[width=\imwidth]{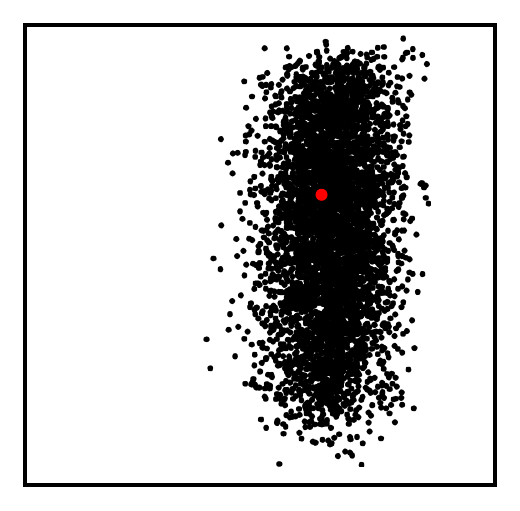} &
\includegraphics[width=\imwidth]{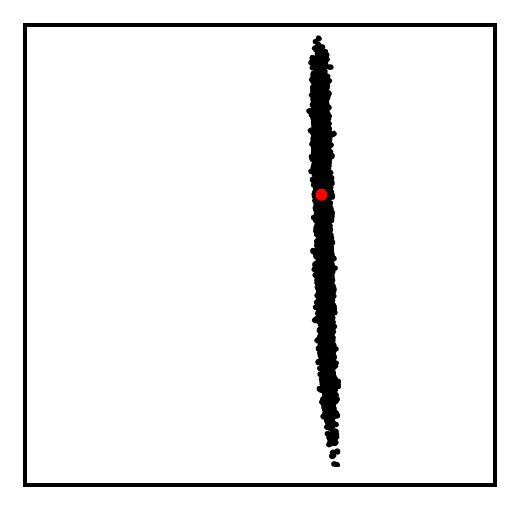} &
\includegraphics[width=\imwidth]{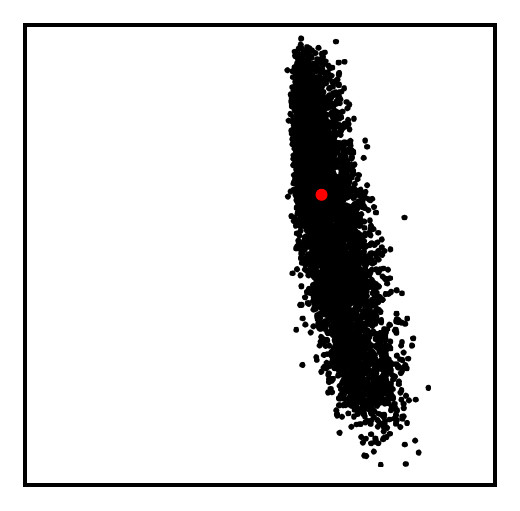} &
\includegraphics[width=\imwidth]{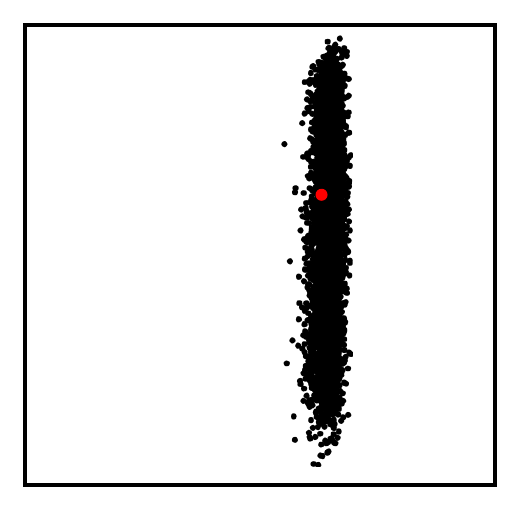} &
\includegraphics[width=\imwidth]{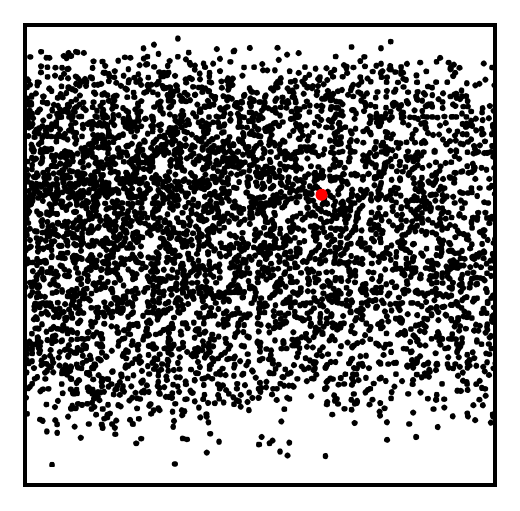} &
\includegraphics[width=\imwidth]{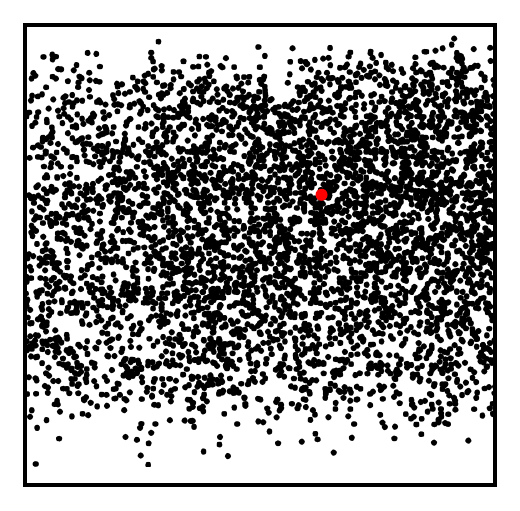} &
\includegraphics[width=\imwidth]{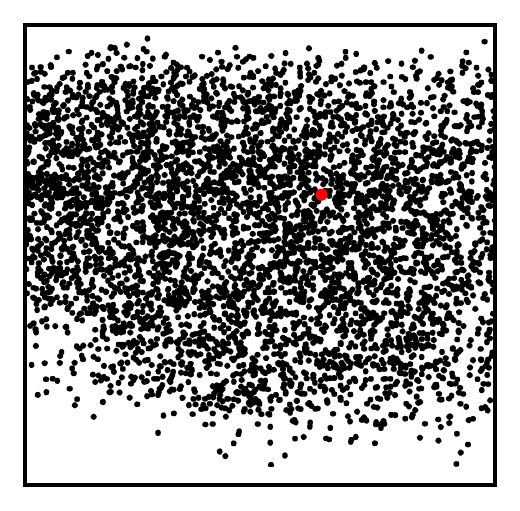} &
\includegraphics[width=\imwidth]{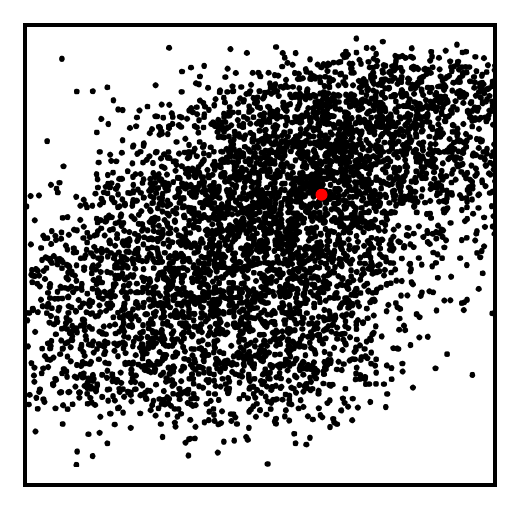} &
\includegraphics[width=\imwidth]{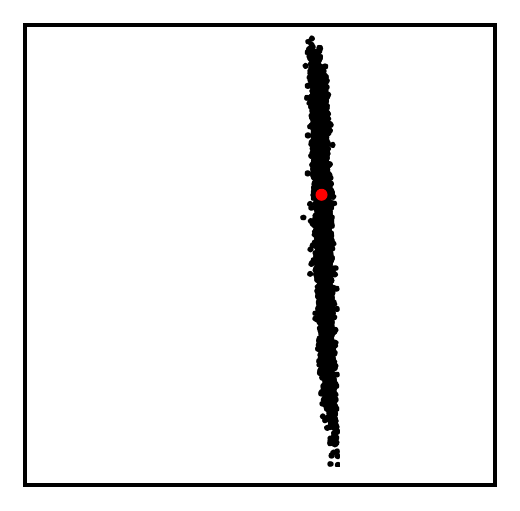} &
\includegraphics[width=\imwidth]{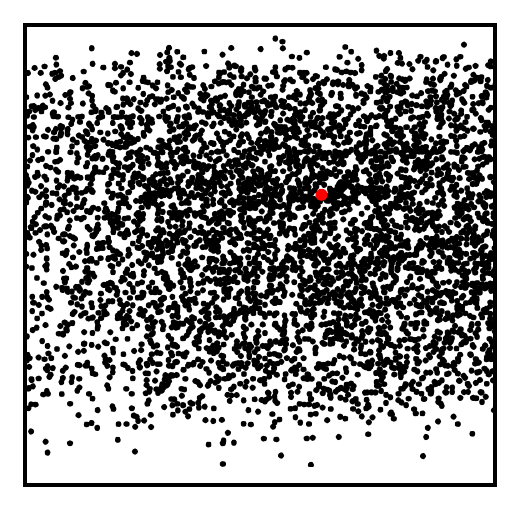} \\
{\scriptsize \raisebox{\raiseheight}{$\log\br{\overline{g}_\mathrm{Na}}$}} &
\includegraphics[width=\imwidth]{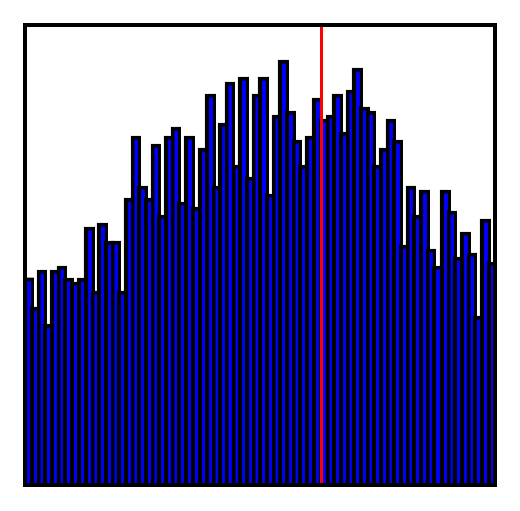} &
\includegraphics[width=\imwidth]{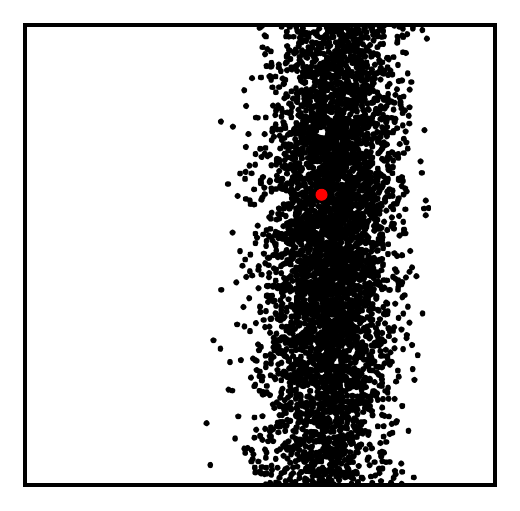} &
\includegraphics[width=\imwidth]{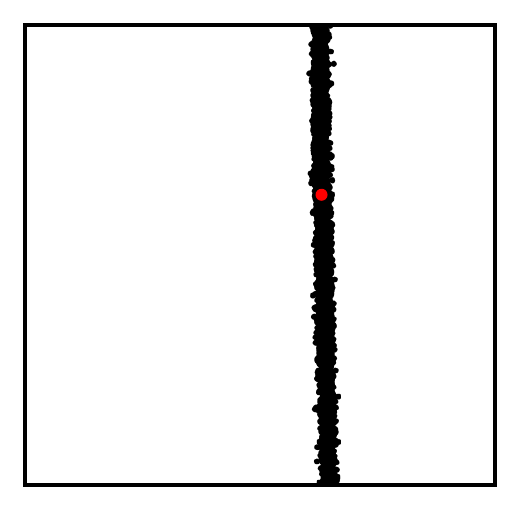} &
\includegraphics[width=\imwidth]{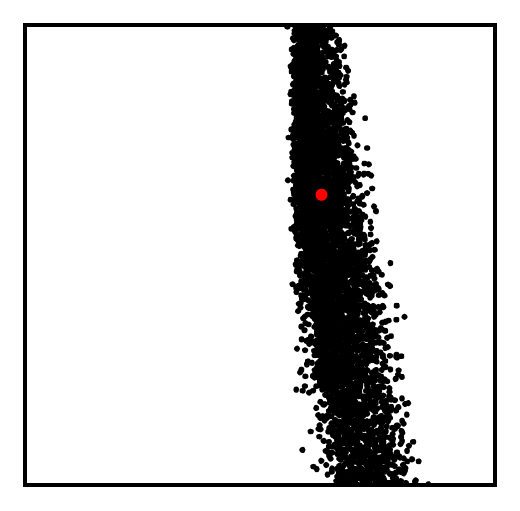} &
\includegraphics[width=\imwidth]{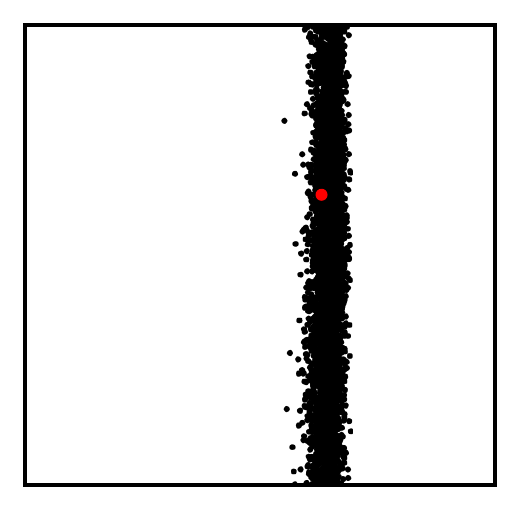} &
\includegraphics[width=\imwidth]{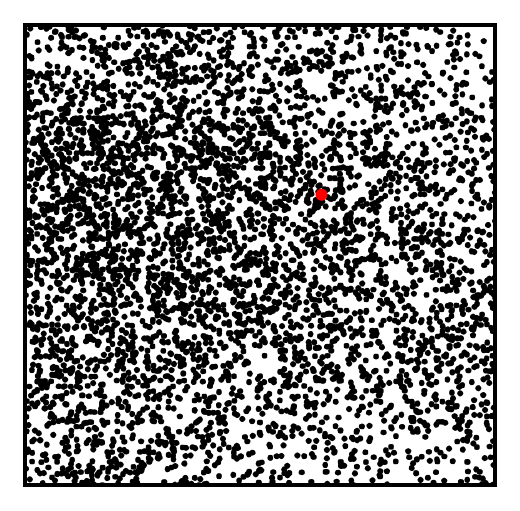} &
\includegraphics[width=\imwidth]{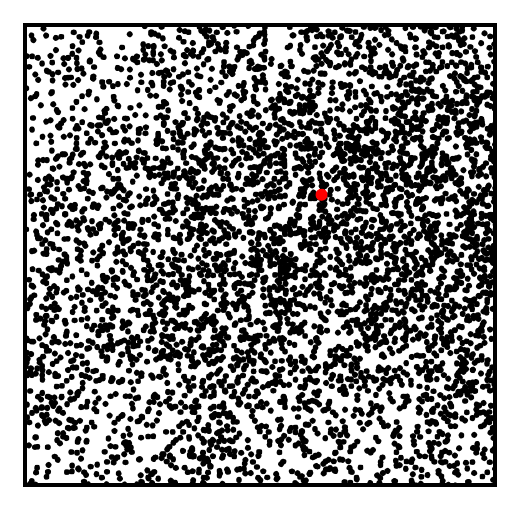} &
\includegraphics[width=\imwidth]{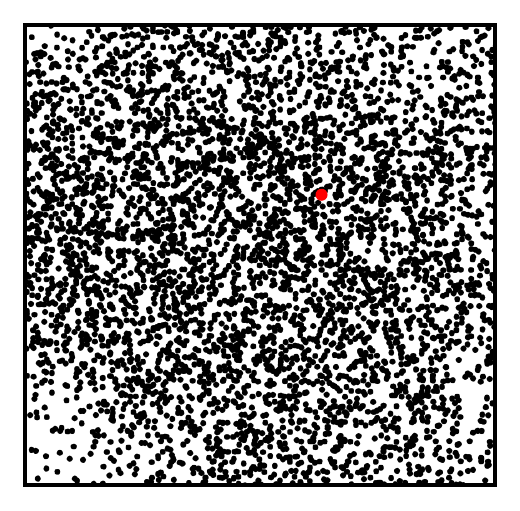} &
\includegraphics[width=\imwidth]{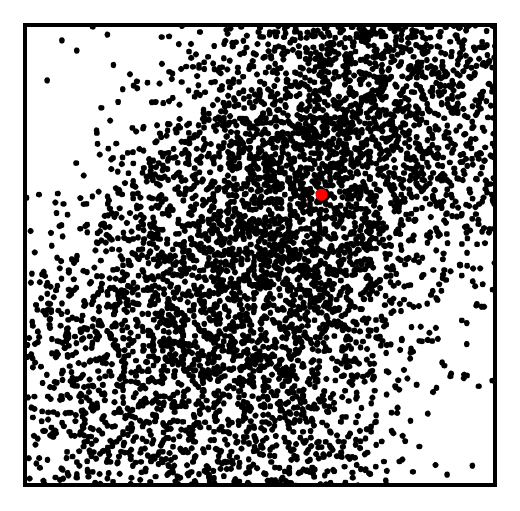} &
\includegraphics[width=\imwidth]{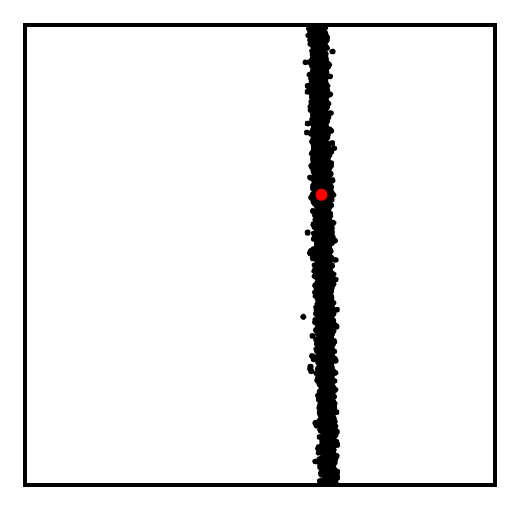} &
\includegraphics[width=\imwidth]{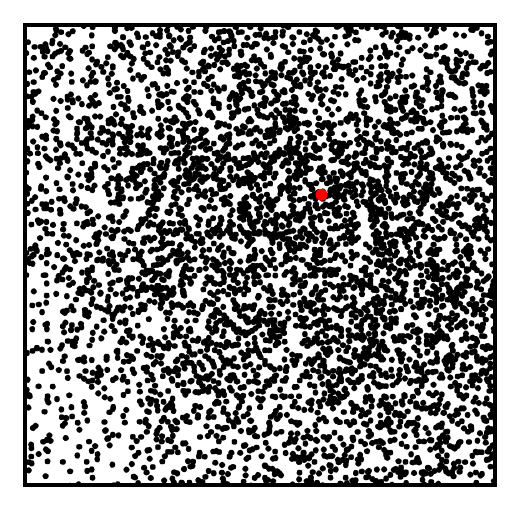} \\
& {\scriptsize \raisebox{\raiseheight}{$\log\br{\overline{g}_\mathrm{K}}$}} &
\includegraphics[width=\imwidth]{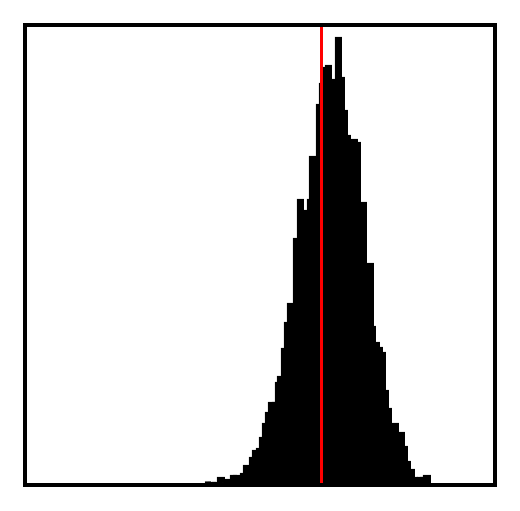} &
\includegraphics[width=\imwidth]{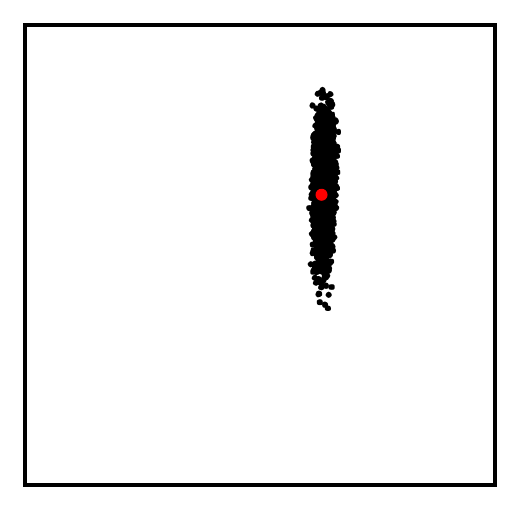} &
\includegraphics[width=\imwidth]{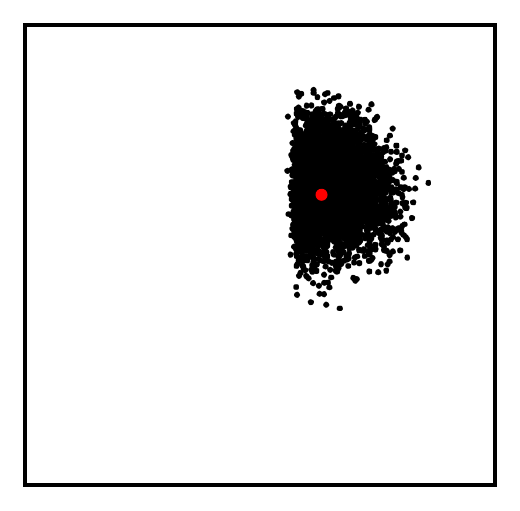} &
\includegraphics[width=\imwidth]{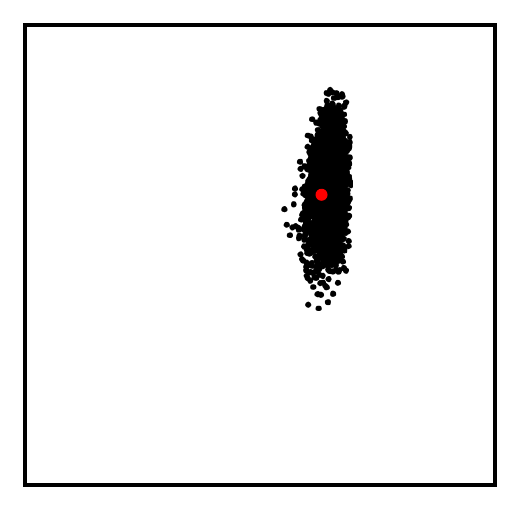} &
\includegraphics[width=\imwidth]{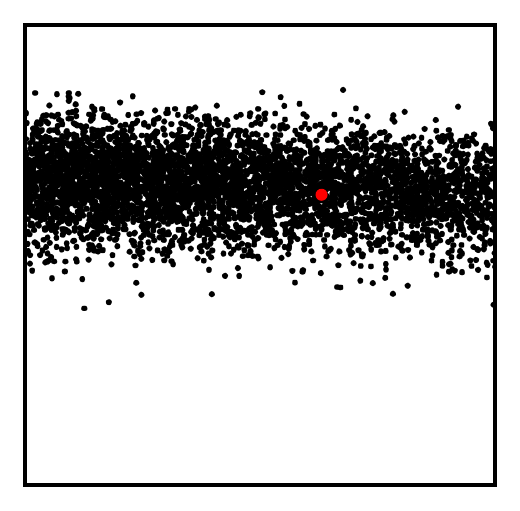} &
\includegraphics[width=\imwidth]{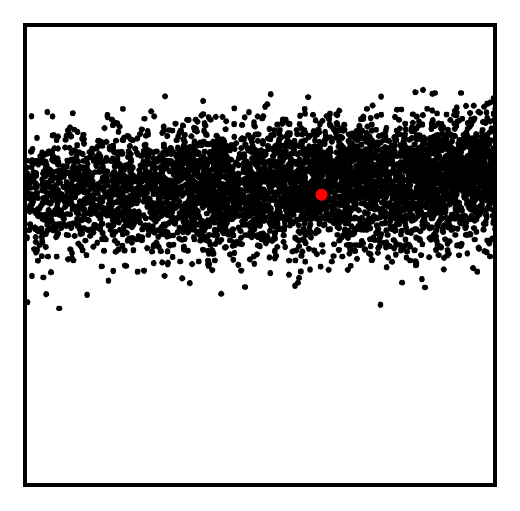} &
\includegraphics[width=\imwidth]{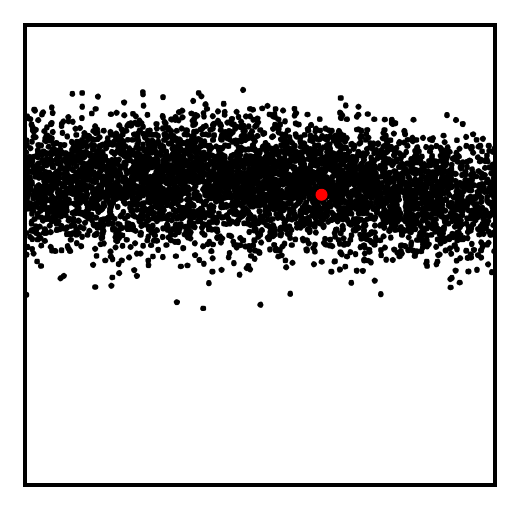} &
\includegraphics[width=\imwidth]{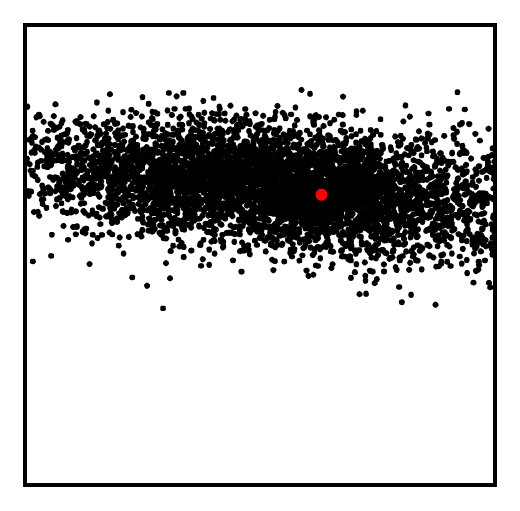} &
\includegraphics[width=\imwidth]{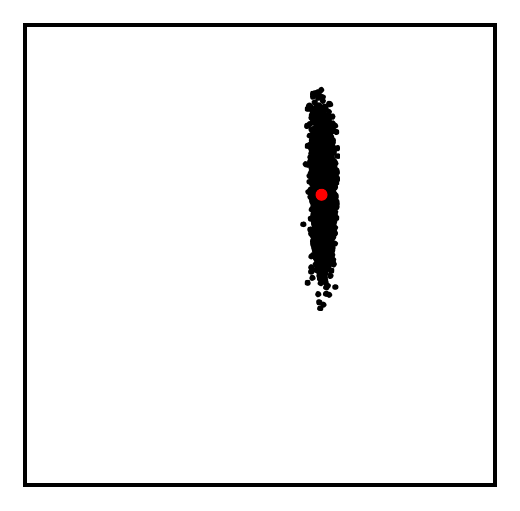} &
\includegraphics[width=\imwidth]{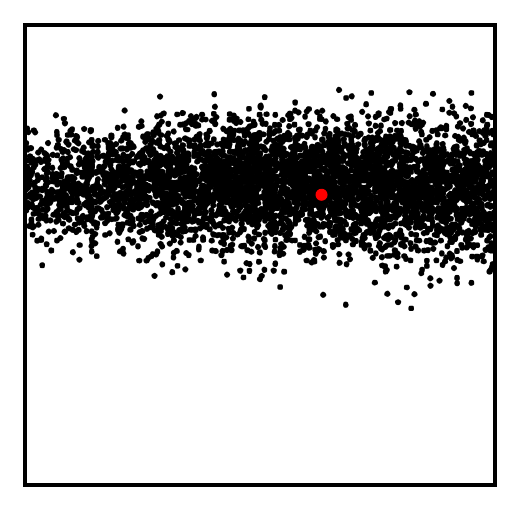} \\
& & {\scriptsize \raisebox{\raiseheight}{$\log\br{\overline{g}_\ell}$}} &
\includegraphics[width=\imwidth]{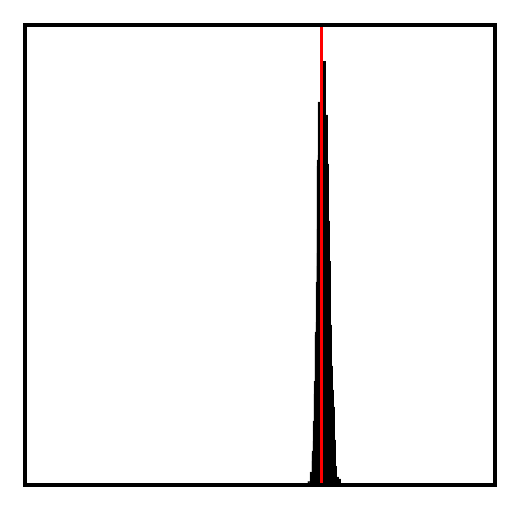} &
\includegraphics[width=\imwidth]{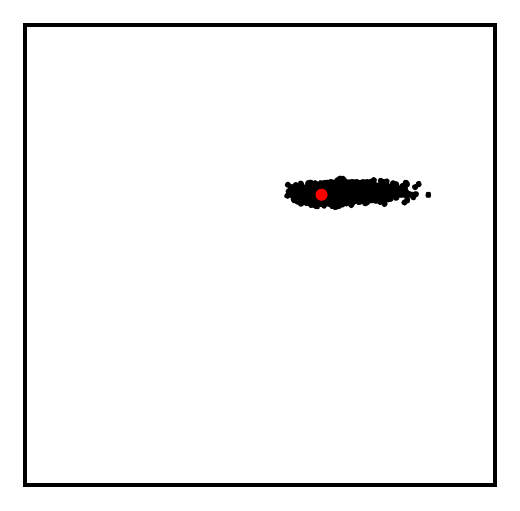} &
\includegraphics[width=\imwidth]{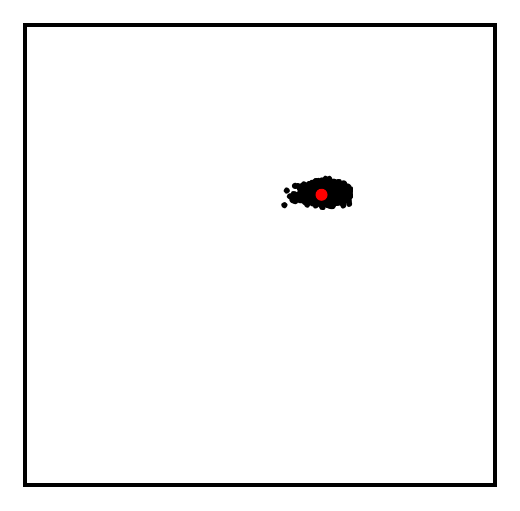} &
\includegraphics[width=\imwidth]{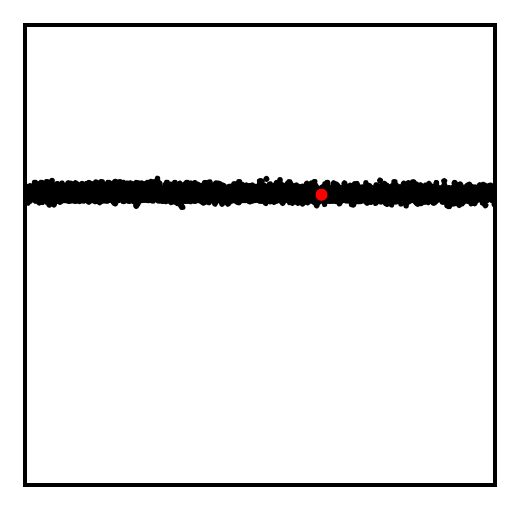} &
\includegraphics[width=\imwidth]{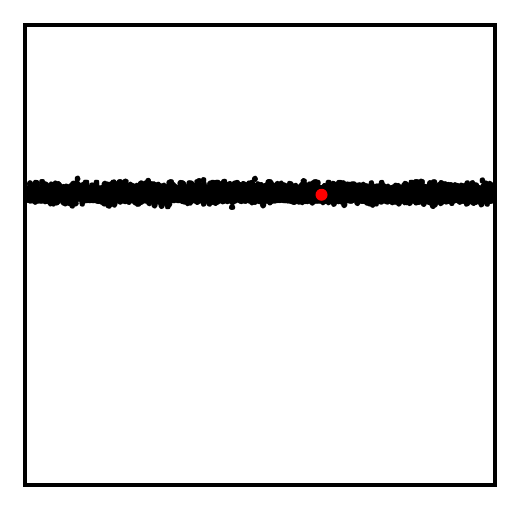} &
\includegraphics[width=\imwidth]{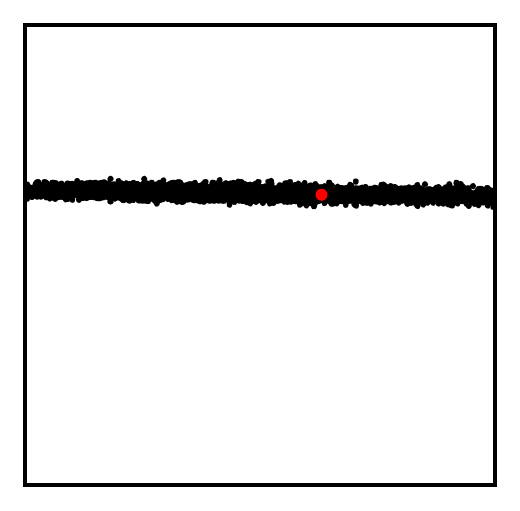} &
\includegraphics[width=\imwidth]{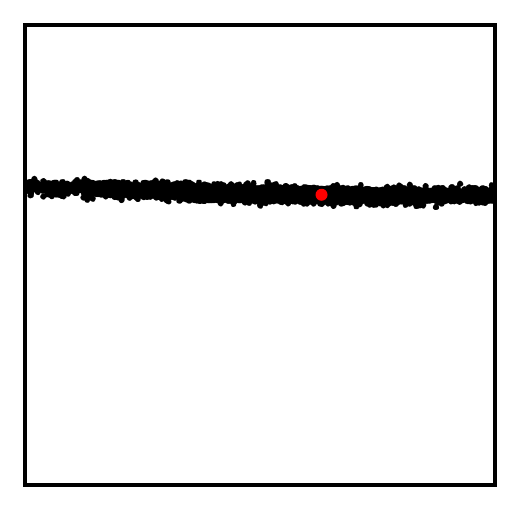} &
\includegraphics[width=\imwidth]{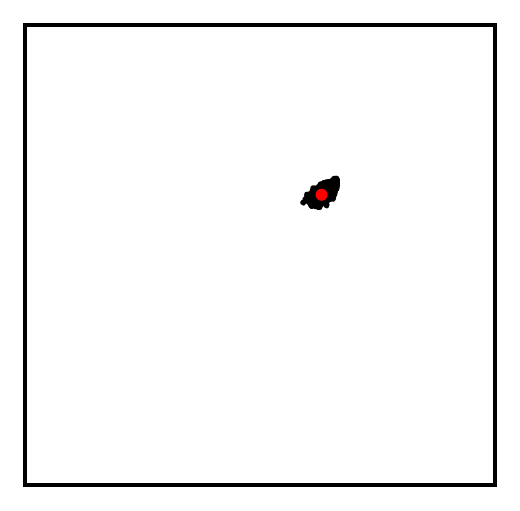} &
\includegraphics[width=\imwidth]{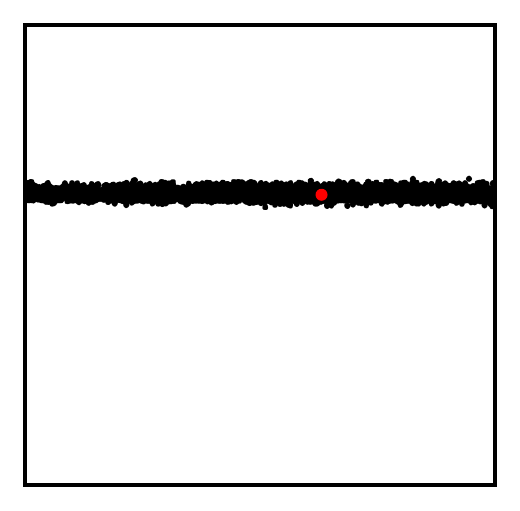} \\
& & & {\scriptsize \raisebox{\raiseheight}{$\log\br{E_\mathrm{Na}}$}} &
\includegraphics[width=\imwidth]{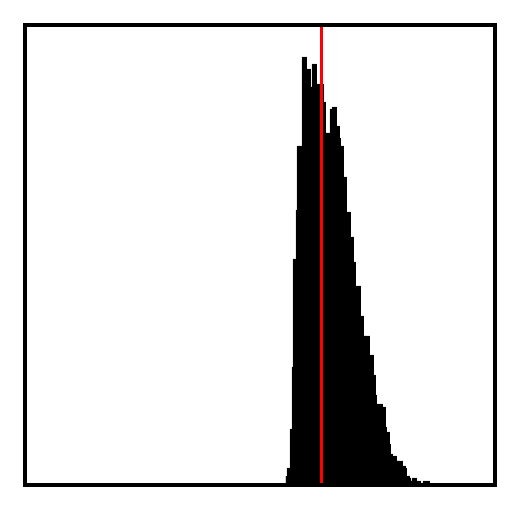} &
\includegraphics[width=\imwidth]{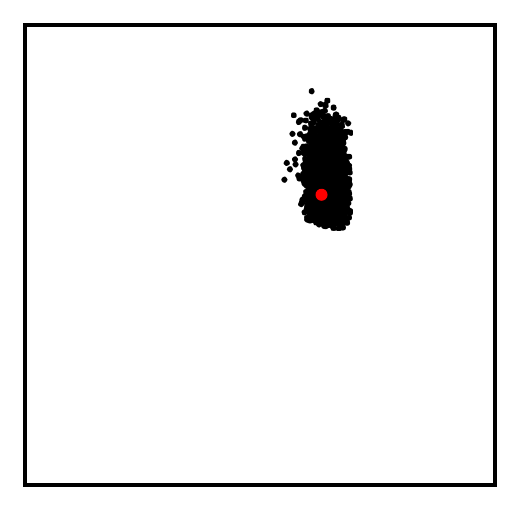} &
\includegraphics[width=\imwidth]{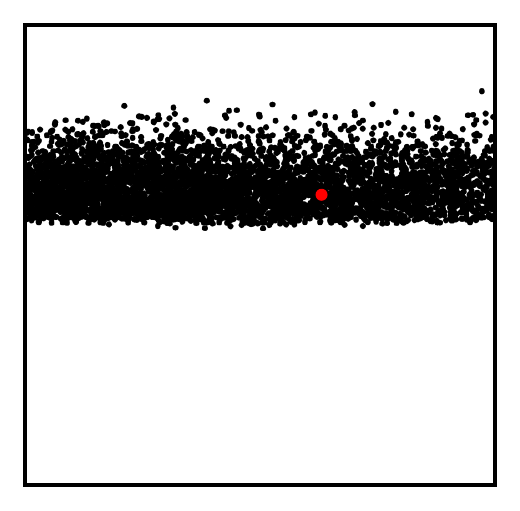} &
\includegraphics[width=\imwidth]{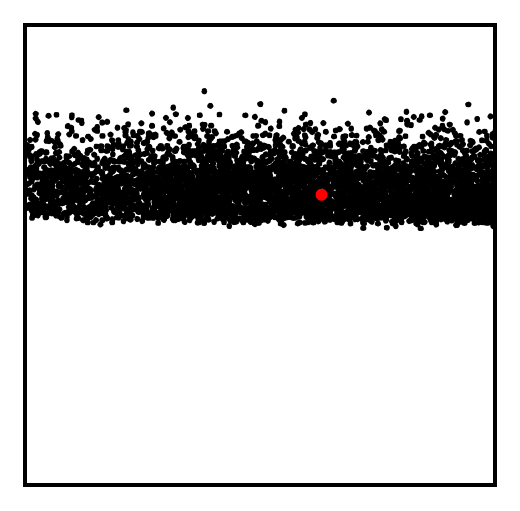} &
\includegraphics[width=\imwidth]{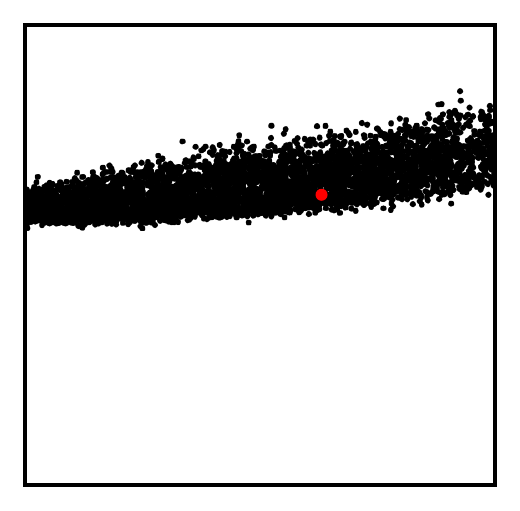} &
\includegraphics[width=\imwidth]{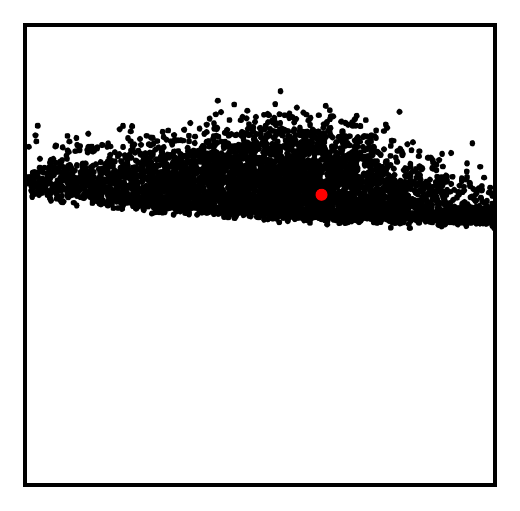} &
\includegraphics[width=\imwidth]{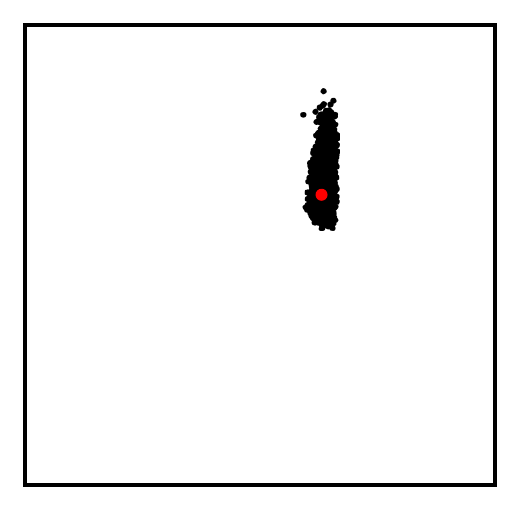} &
\includegraphics[width=\imwidth]{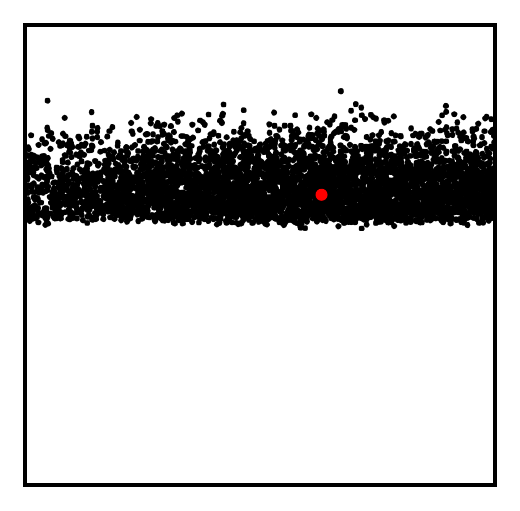} \\
& & & & {\scriptsize \raisebox{\raiseheight}{$\log\br{\hspace{-1pt}-\hspace{-1pt}E_\mathrm{K}}$}} &
\includegraphics[width=\imwidth]{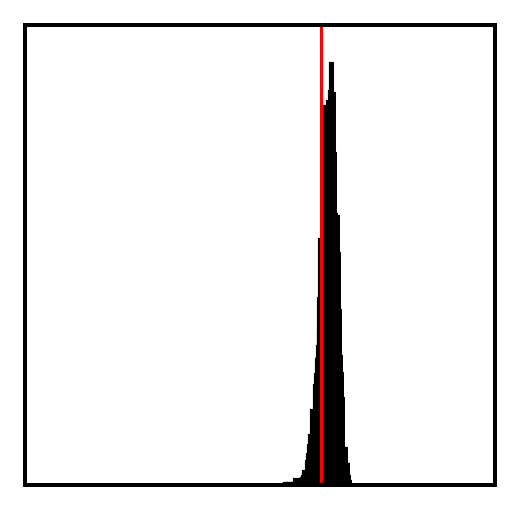} &
\includegraphics[width=\imwidth]{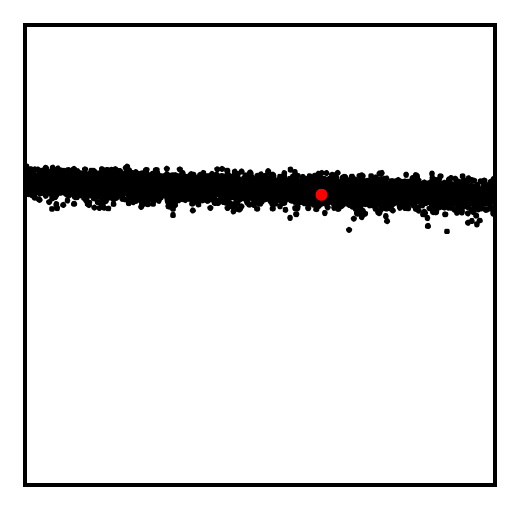} &
\includegraphics[width=\imwidth]{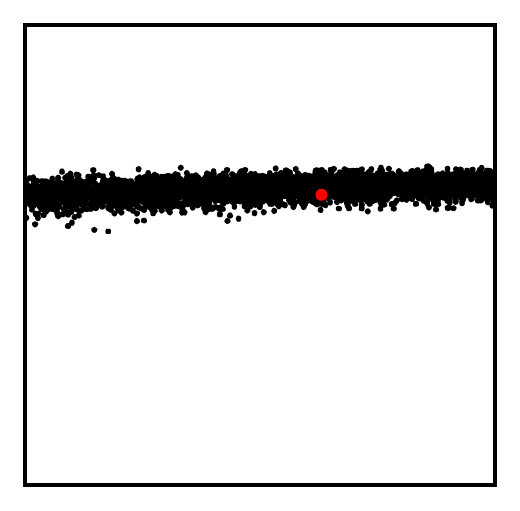} &
\includegraphics[width=\imwidth]{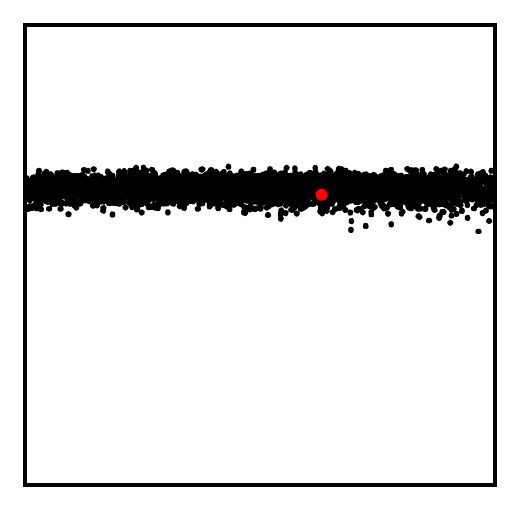} &
\includegraphics[width=\imwidth]{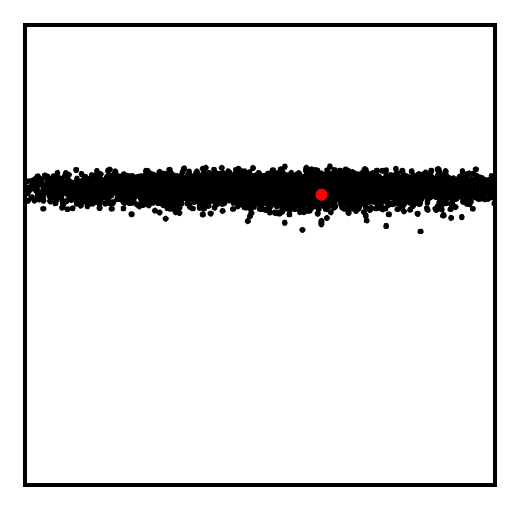} &
\includegraphics[width=\imwidth]{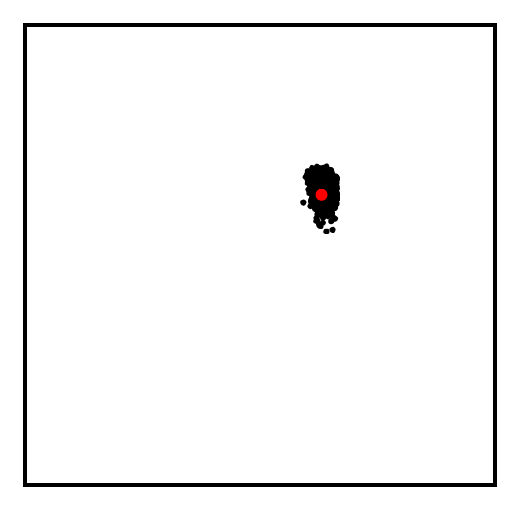} &
\includegraphics[width=\imwidth]{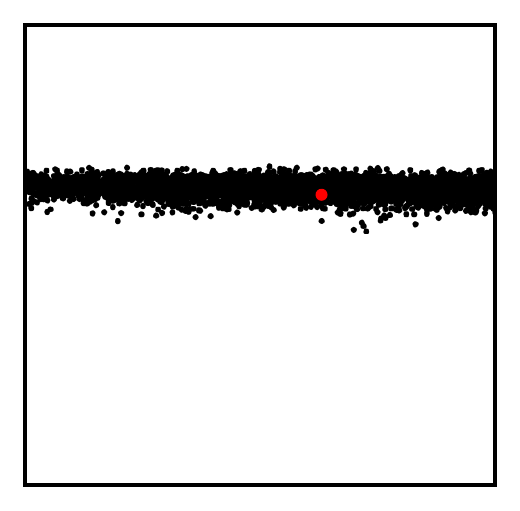} \\
& & & & & {\scriptsize \raisebox{\raiseheight}{$\log\br{-E_\ell}$}} &
\includegraphics[width=\imwidth]{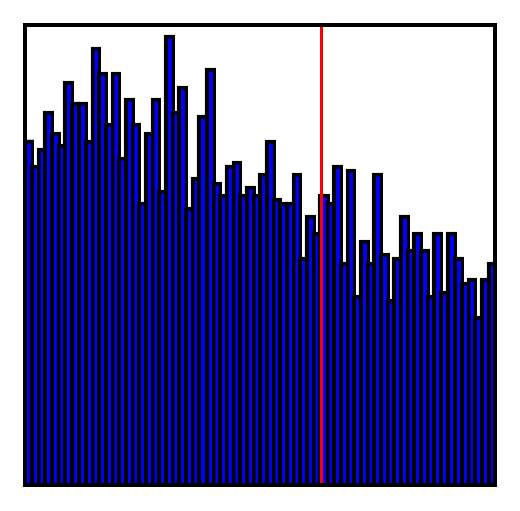} &
\includegraphics[width=\imwidth]{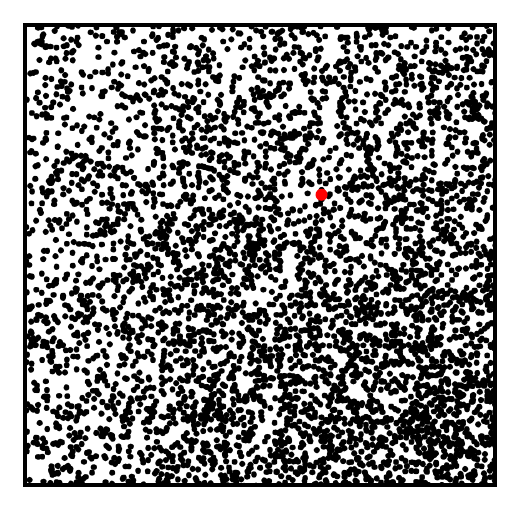} &
\includegraphics[width=\imwidth]{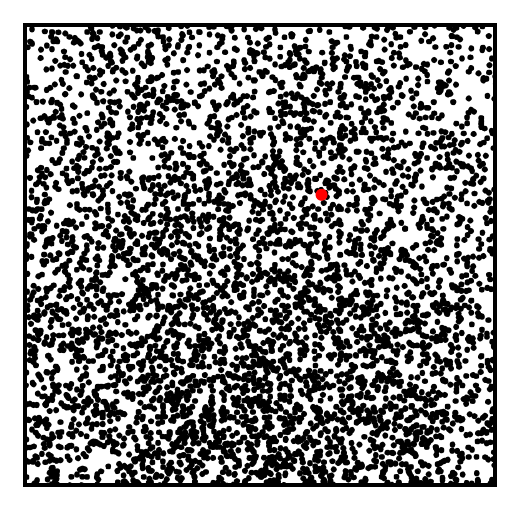} &
\includegraphics[width=\imwidth]{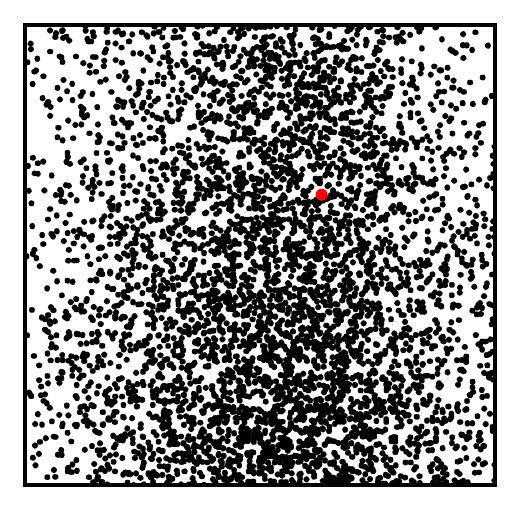} &
\includegraphics[width=\imwidth]{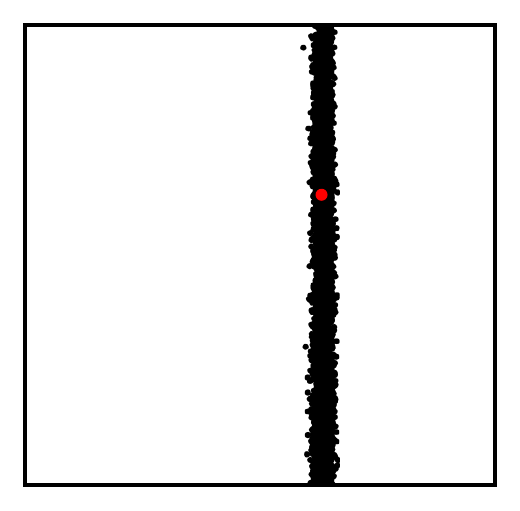} &
\includegraphics[width=\imwidth]{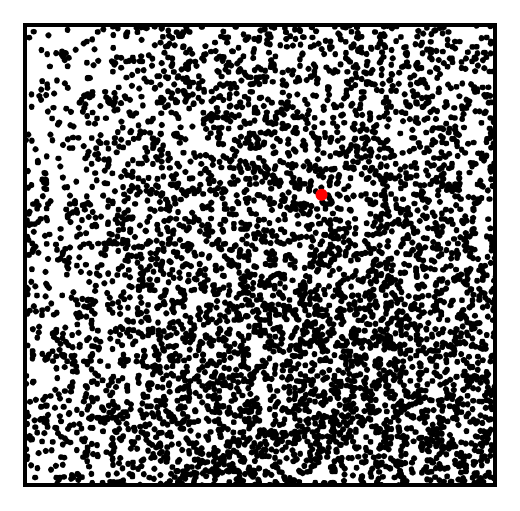} \\
& & & & & & {\scriptsize \raisebox{\raiseheight}{$\log\br{\overline{g}_\mathrm{M}}$}} &
\includegraphics[width=\imwidth]{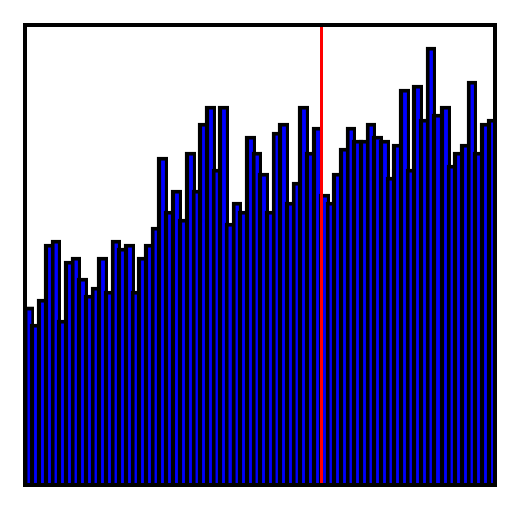} &
\includegraphics[width=\imwidth]{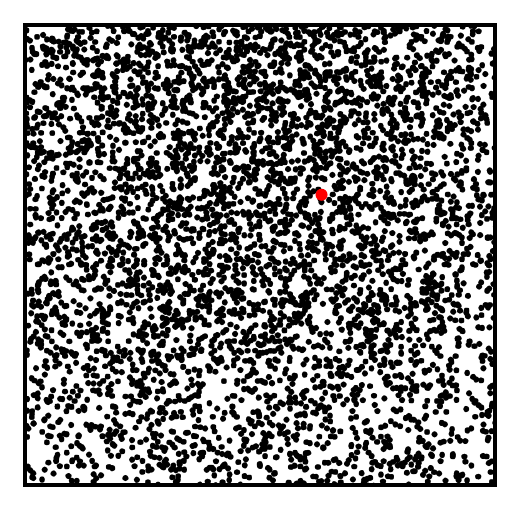} &
\includegraphics[width=\imwidth]{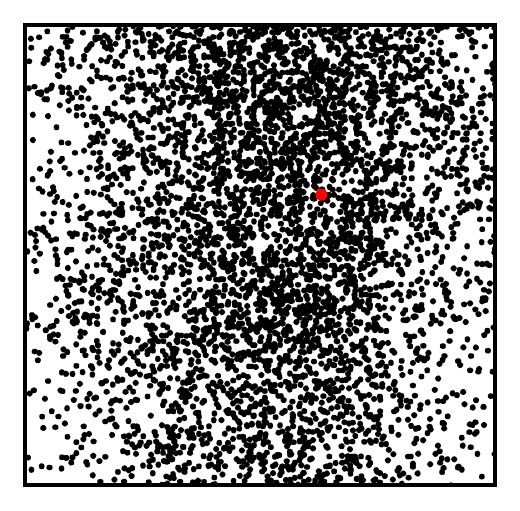} &
\includegraphics[width=\imwidth]{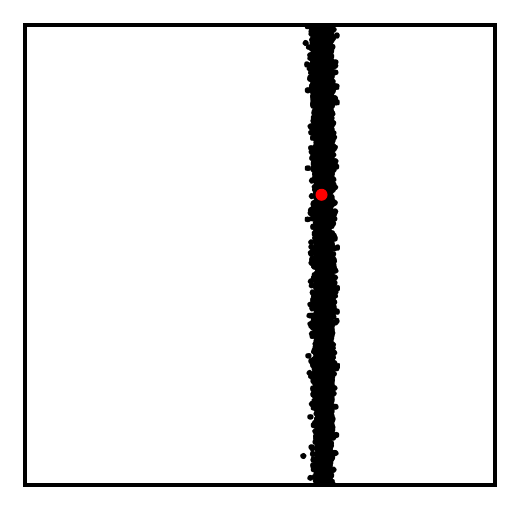} &
\includegraphics[width=\imwidth]{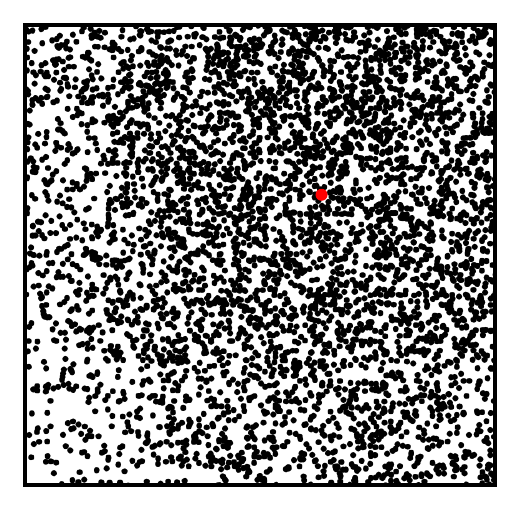} \\
& & & & & & & {\scriptsize \raisebox{\raiseheight}{$\log\br{\tau_\mathrm{max}}$}} &
\includegraphics[width=\imwidth]{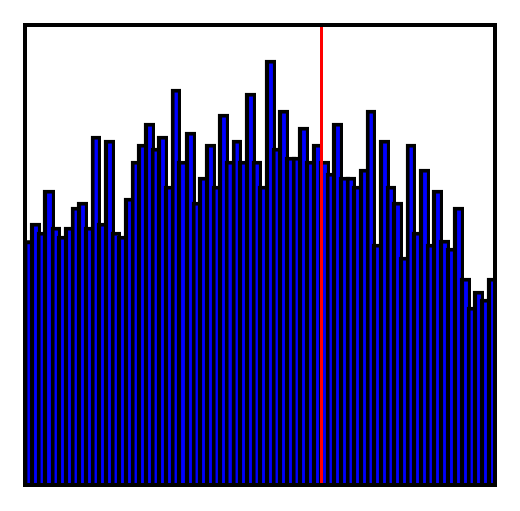} &
\includegraphics[width=\imwidth]{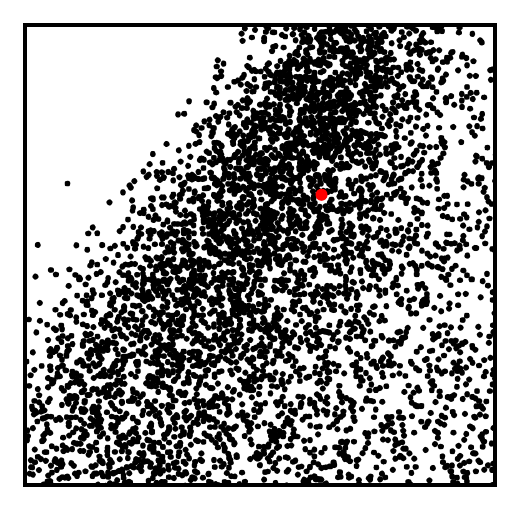} &
\includegraphics[width=\imwidth]{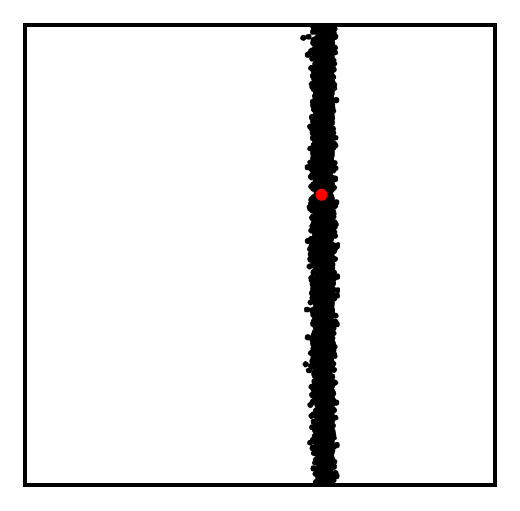} &
\includegraphics[width=\imwidth]{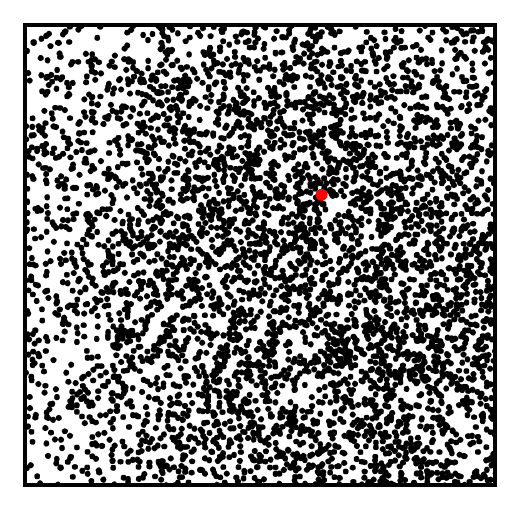} \\
& & & & & & & & {\scriptsize \raisebox{\raiseheight}{$\log\br{k_{\beta\mathrm{n}1}}$}} &
\includegraphics[width=\imwidth]{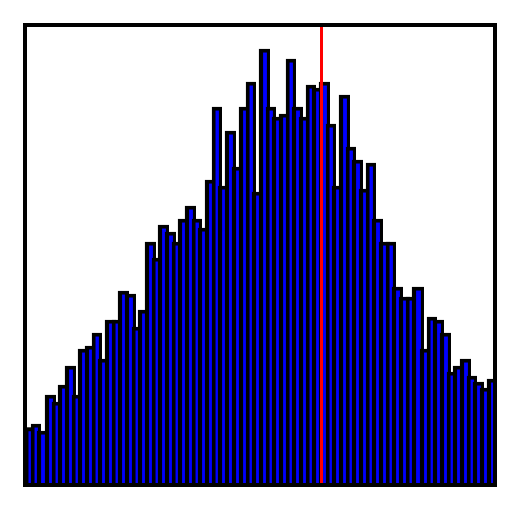} &
\includegraphics[width=\imwidth]{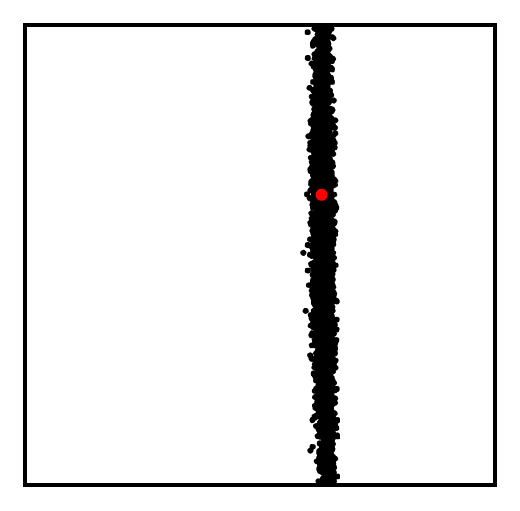} &
\includegraphics[width=\imwidth]{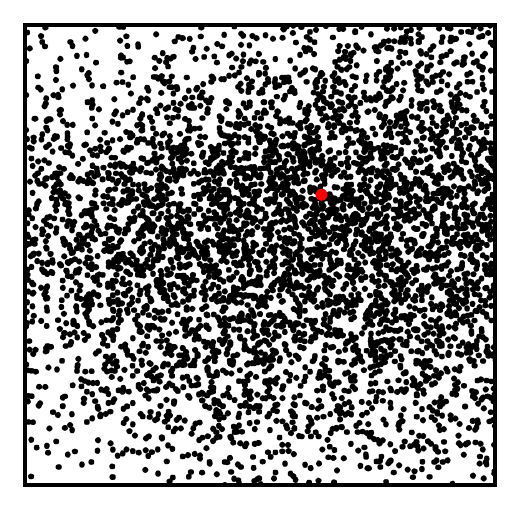} \\
& & & & & & & & & {\scriptsize \raisebox{\raiseheight}{$\log\br{k_{\beta\mathrm{n}2}}$}} &
\includegraphics[width=\imwidth]{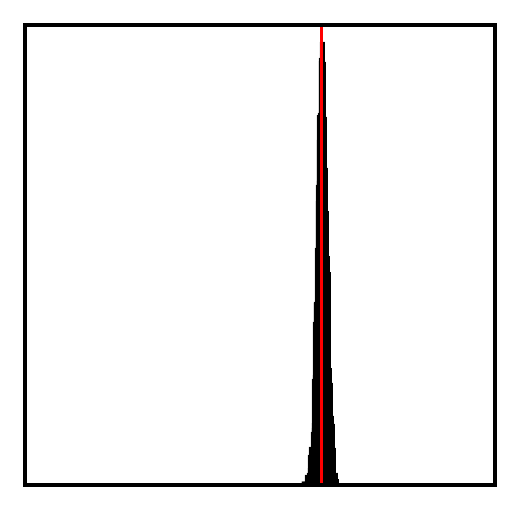} &
\includegraphics[width=\imwidth]{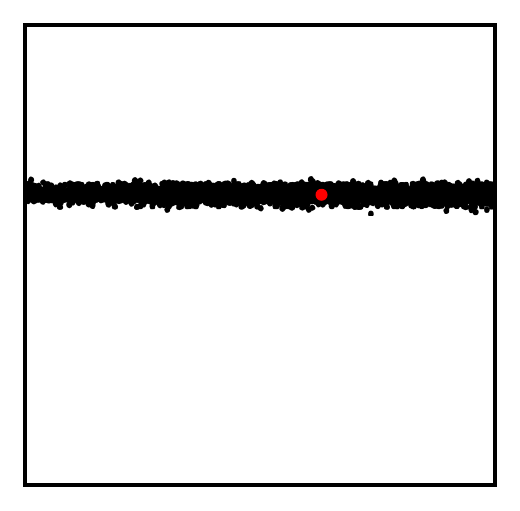} \\
& & & & & & & & & & {\scriptsize \raisebox{\raiseheight}{$\log\br{-V_\mathrm{T}}$}} &
\includegraphics[width=\imwidth]{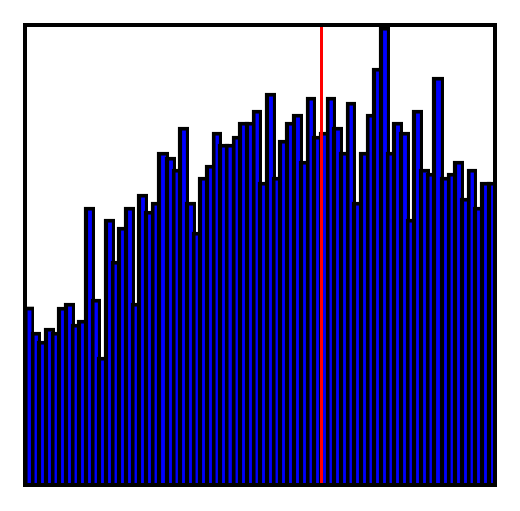} \\
& & & & & & & & & & & {\scriptsize \raisebox{\raiseheight}{$\log\br{\sigma}$}}
\end{tabular}

\caption{Hodgkin--Huxley model: MCMC samples from SNL posterior. True parameters $\p^*$ are indicated in red. The range of each histogram is $\left[\theta_i^*\!-\!\log2,\,\, \theta_i^*\!+\!\log1.5\right]$  for $i=1,\ldots,12$. Compare with Figure~G.2 by \citet{Lueckmann:2017}.}
\label{fig:hh:snl_post}
\end{figure*}

The results are shown in Figure~\ref{fig:hh:all}. The SNL posterior is
shown in Figure~\ref{fig:hh:snl_post}; the extent of each plot
corresponds to the uniform prior. In the SNL posterior, it can be seen
that all parameters are clustered around their true values (red dots
and lines). 

The sodium and equilibrium potentials are relatively tightly
clustered, and the potassium equilibrium potential less so:
\begin{align}
E_\mathrm{Na}&=50\pm2\text{ \si{mV}}\\
E_\mathrm{K}&=-99 \text{ \si{mV} (range [\SI{-121}{mV}, \SI{-90}{mV}])}\\
E_\ell&= -70\pm4\text{ \si{mV}}.
\end{align}
The tight clustering reflects that concentrations, and hence
equilibrium potentials, are maintained within a range by neuronal ion
exchangers and and glial buffering \citep{Somjen:2002}. Furthermore,
when regulation of ion concentration fails, pathological brain states
can arise \citep{Somjen:2002}. The longer tail of the potassium
equilibrium potential posterior might be due to it having relatively
little influence on the mean of $V(t)$, since at lower potentials the
potassium conductance $\overline{g}_\mathrm{K}n^4$ will be relatively
small, so, according to Equation~\eqref{eq:hh:IK}, the potassium
current will also be small. The quantity $V_\mathrm{T}$, which adjusts
the threshold of spike initiation, is also fairly tightly controlled,
which will tend to keep the firing rate around a constant value.
                               
In contrast to the equilibrium potentials, the conductances vary more,
within a factor of $1.8$ for $g_\ell$, and a factor of $3$ for
$\overline{g}_\mathrm{Na}$, $\overline{g}_\mathrm{K}$ and
$\overline{g}_\mathrm{M}$. Moreover, $\overline{g}_\mathrm{Na}$ and
$\overline{g}_\mathrm{K}$ are correlated, which is consistent with
their opposing depolarizing and hyperpolarizing influences on the
membrane potential. A higher sodium conductance could lead to the cell
being hyper-excitable, but this should be counteracted by a greater
potassium conductance. This allows for their wide range, and is
consistent with the biological evidence for diverse but correlated
sets of channel conductances underlying particular activity patterns
\citep{Goldman:2001}. In contrast, $\overline{g}_\mathrm{M}$ appears
to have relatively little influence over the output, and is not
correlated with any other parameters. 

The parameter $\tau_\mathrm{max}$, which also relates to the M-type
potassium channel, also has little effect. Further simulations of
other neuron types could be undertaken to see if these parameters are
generally loosely constrained, which could then lead to experimentally
testable predictions about the density and variability of M-type
conductances.

The parameters $k_\mathrm{\beta n1}$ and $k_\mathrm{\beta n2}$ also
have relatively wide ranges, and there is a weak correlation between
the two. Increasing $k_\mathrm{\beta n1}$ effectively increases the
half-activation voltage of potassium conductances, and increasing
$k_\mathrm{\beta n2}$ makes the slope of transition less pronounced.
The lack of posterior at high $k_\mathrm{\beta n1}$ (high threshold)
and low $k_\mathrm{\beta n2}$ (sharper transition) might cause the
neuron not to repolarize quickly enough, and hence be hyper-excitable.

Finally, we note that the posterior computed by SNL is qualitatively consistent with the posterior reported by \citet{Lueckmann:2017} in Figure G.2 of their article.

\bibliography{references}

\end{document}